%% file: iclr_2027.tex
\documentclass{article}

\usepackage{iclr2027_conference,times}

\usepackage[utf8]{inputenc}
\usepackage[T1]{fontenc}
\usepackage{amsmath}
\usepackage{amssymb}
\usepackage{amsfonts}
\usepackage{booktabs}
\usepackage{longtable}
\usepackage{multirow}
\usepackage{tabularx}
\usepackage{graphicx}
\usepackage{microtype}
\usepackage{xcolor}
\usepackage{url}
\usepackage{algorithm}
\usepackage{algpseudocode}
\usepackage[pdfusetitle,hidelinks]{hyperref}

\newif\ifshowrevisions
\showrevisionsfalse
\definecolor{revisionblue}{RGB}{0,70,140}
\DeclareRobustCommand{\rev}[1]{{\ifshowrevisions\color{revisionblue}\fi#1}}
\newenvironment{revision}{\ifshowrevisions\color{revisionblue}\fi}{}
\algrenewcommand\algorithmicrequire{\textbf{Input:}}
\algrenewcommand\algorithmicensure{\textbf{Output:}}

\graphicspath{{Figures/}}
\newcommand{\sysbiogithub}[1]{\href{https://github.com/hengrao02/SysBio-SR-KineSelect}{#1}}
\newcommand{\sysbiohuggingface}[1]{\href{https://huggingface.co/datasets/HengRao/SysBio-SR}{#1}}

\title{SysBio-SR: Benchmarking Symbolic Regression for Trajectory Fidelity and Mechanism Recovery in Systems Biology}

\author{%
  \textbf{Heng Rao\textsuperscript{1}\quad
          Dan Gao\textsuperscript{2}\quad
          Benjamin J Hoffman\textsuperscript{2}\quad
          Yu Gu\textsuperscript{1}}\\
  \textbf{Zhenghao Liu\textsuperscript{1}\quad
          Ge Yu\textsuperscript{1}\quad
          Yang Cao\textsuperscript{3}\quad
          Minghan Chen\textsuperscript{2}}%
  \thanks{Correspondence to Minghan Chen at
    \href{mailto:chenm@wfu.edu}{\texttt{chenm@wfu.edu}}.}\\[6pt]
  \textsuperscript{1} Northeastern University\qquad
  \textsuperscript{2} Wake Forest University\\
  \textsuperscript{3} Virginia Polytechnic Institute and State University\\
  GitHub: \url{https://github.com/hengrao02/SysBio-SR-KineSelect}\\
  Hugging Face: \url{https://huggingface.co/datasets/HengRao/SysBio-SR}
}
\hypersetup{
  pdfauthor={Heng Rao, Dan Gao, Benjamin J Hoffman, Yu Gu, Zhenghao Liu, Ge Yu, Yang Cao, Minghan Chen},
  pdfdisplaydoctitle=true
}
\iclrfinalcopy

\begin{document}

\maketitle
\lhead{Preprint}

\begin{abstract}
\input{sections/abstract.tex}
\end{abstract}

\input{sections/introduction}
\input{sections/related_work}

\input{sections/dataset}
\input{sections/evaluation}

\input{sections/method.tex}

\input{sections/experiments}
\input{sections/conclusion}

\clearpage
\subsection*{AI use statement}

We used ChatGPT for literature searches, figure preparation using recorded experimental data, and assistance with code review.
We used Gemini for manuscript language editing.
We also used generative AI tools to obtain feedback on experimental design.
We did not use these tools to formulate research questions or hypotheses or to design the methods or mathematical formulations.
We manually reviewed the code and checked the mathematical formulations and method descriptions.
We verified numerical values in the tables, figures, and text against the original experimental records.
We have reviewed the entire manuscript and take responsibility for its final content, including material prepared with AI assistance.

\subsection*{Ethics statement}

This work uses publicly available, literature-derived systems-biology ODE models and trajectories simulated from these models.
It involves no new experiments on human participants or animals and no collection or processing of personally identifiable data.

\subsection*{Reproducibility statement}

Appendix~\ref{app:benchmark_details} documents the benchmark construction, trajectory generation, and evaluation protocols, including reaction matching, score aggregation, and failure handling.
Appendix~\ref{app:kineselect_details} specifies the KineSelect algorithm, including RSS, KFI, and FPMS.
Appendix~\ref{app:supp_experimental_results} reports supplementary analyses and experiments, including selection sensitivity and ablations, noise robustness, and representative cases.
Appendix~\ref{app:reproducibility} describes the experimental pipeline, baseline implementations, method configurations, search budgets, and settings for the selection and noise experiments.
We provide the SysBio-SR benchmark and experiment code on \sysbiogithub{GitHub}.
The benchmark is also available on \sysbiohuggingface{Hugging Face}.

{
\small
\bibliographystyle{iclr2027_conference}
\bibliography{references}
}

\appendix
\input{appendix}

\end{document}

%% file: sections/abstract.tex
Symbolic regression seeks to discover interpretable governing equations for dynamical systems from data. In systems biology, however, accurately reproducing system trajectories does not necessarily imply recovery of the underlying reaction mechanisms.
Existing benchmarks evaluate numerical accuracy, expression complexity, or equation-level similarity, but not agreement between recovered reactions and reference mechanisms in biological systems.
We introduce SysBio-SR, a benchmark of 198 literature-based biological ODE systems with reaction annotations spanning mass action, Michaelis--Menten, Hill, and complex kinetics.
Trajectory fidelity is evaluated across all systems under in-distribution reconstruction and out-of-distribution (OOD) extrapolation, while mechanism recovery is assessed on the subset of 161 systems.
We evaluate trajectory fidelity across 11 symbolic-discovery methods and assess mechanism recovery for three reaction-based methods.
Joint analysis across systems highlights how severely trajectory-based evaluation overestimates true mechanism recovery.
Motivated by these findings, we develop KineSelect, a framework that searches candidate reaction topologies under biologically grounded kinetic laws to select optimal mechanisms that balance trajectory fidelity with structural parsimony.
Compared with the strongest baseline, KineSelect improves aggregate OOD extrapolation \(R^2\) by 19.2\% and overall reaction F1 score by 14.3\%.

%% file: sections/introduction.tex
\section{Introduction}

Biological systems evolve through interacting processes such as production, degradation, and regulation~\citep{kholodenko2006cell, fell1997understanding, aldridge2006physicochemical}, which are classically modeled using ordinary differential equations (ODEs).
Traditionally, constructing these mathematical models relies on labor-intensive manual curation from domain literature and expert intuition. Symbolic regression (SR) offers a promising data-driven alternative to automate this workflow by inferring interpretable governing equations directly from time series measurements~\citep{koza1994genetic, schmidt2009distilling, brunton2016discovering}.
Unlike black-box neural approximations, SR discovers explicit functional forms to expedite model development and uncover novel regulatory rules.

\begin{figure}[t]
    \centering
    \includegraphics[width=\linewidth]{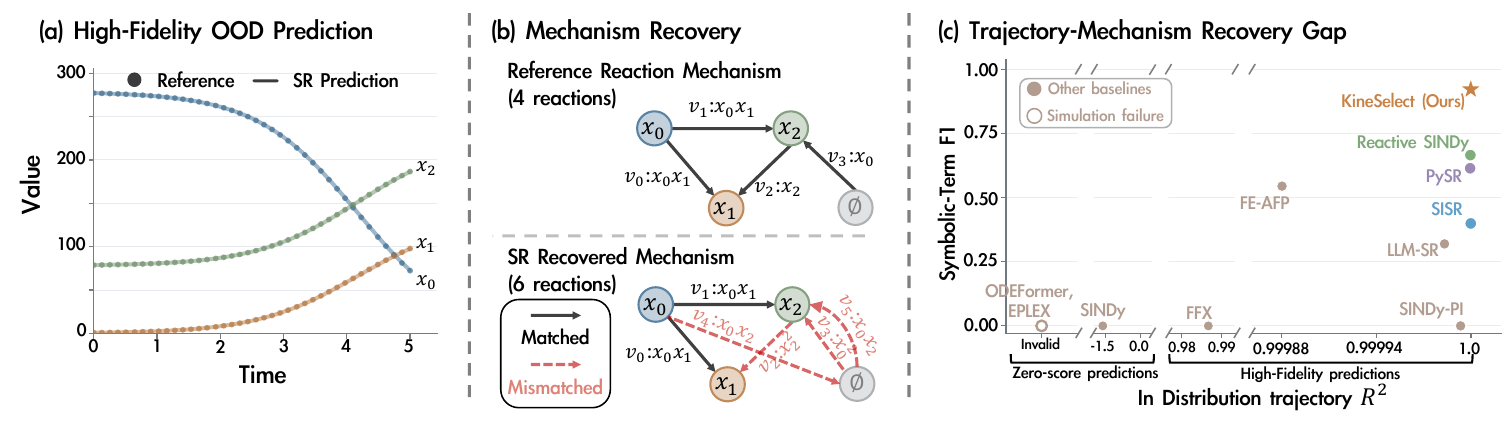}
    \vspace{-0.15in}
    \caption{
        \textbf{Trajectory--mechanism gap.}
        \rev{Single-system example from \citet{munz2009zombies}.
        (a) An SR baseline closely matches the OOD trajectory. (b) Reference and recovered mechanisms. (c) ID \(R^2\) versus symbolic-term \(F_1\).
        We manually match complete signed terms, ignoring parameter values.}
    }
    \label{fig:gap}
\end{figure}
In systems biology, a mechanism comprises both the reaction network topology and its governing kinetic rate laws. However, existing symbolic regression benchmarks evaluate models primarily on numerical trajectory accuracy, expression complexity, or simple equation matching ~\citep{la2021contemporary, matsubara2022rethinking, d2023odeformer, bideh2026mdbench}.
Moreover, these suites focus heavily on physics-inspired scalar expressions~\citep{udrescu2020ai, la2021contemporary} or generic dynamical systems~\citep{d2023odeformer, bideh2026mdbench, liu2025can}, leaving mechanism recovery in coupled biological ODEs unexamined.
This creates a critical evaluation gap: because distinct reaction pathways can produce virtually similar trajectory curves~\citep{szederkenyi2011inference}, high trajectory fidelity does not imply biological correctness.
\rev{On the Munz2009 system~\citep{munz2009zombies}, multiple baselines achieve near-perfect trajectory fidelity $R^2 \approx 1.0$ but symbolic-term F1 of 0--0.67 (Figure~\ref{fig:gap}(c)).
For this example, we compute symbolic-term F1 by manually matching predicted and reference ODE terms. Terms must agree in symbolic form and sign, while parameter values may differ.
LLM-SRBench~\citep{shojaee2025llm} evaluates whole-expression symbolic accuracy using GPT-4o to judge mathematical equivalence between predicted and reference equations after replacing constants with parameter placeholders.}

\begin{figure}[htbp]
\centering
\includegraphics[width=0.85\linewidth]{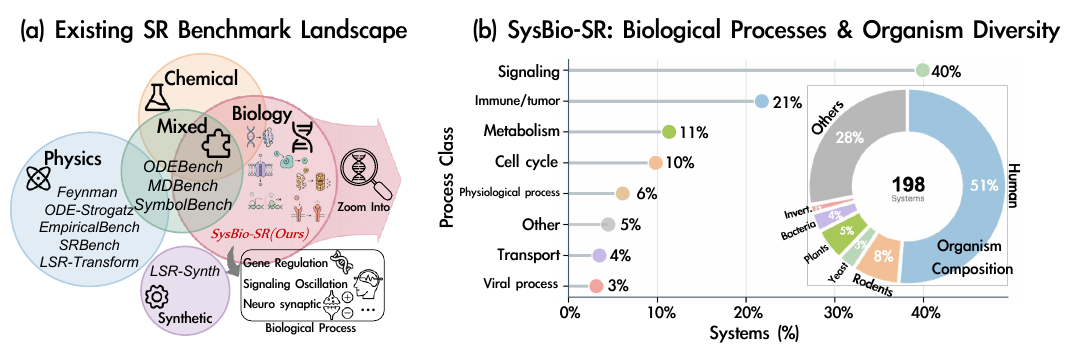}
\caption{\textbf{Positioning and biological coverage of SysBio-SR.}
(a) Landscape of SR benchmarks across scientific domains.
(b) Biological process and organism composition across 198 systems.
}
\label{fig:overview}
\end{figure}

To address this gap, we present \textbf{SysBio-SR}, a benchmark designed for mechanism-aware evaluation of symbolic regression in systems biology. SysBio-SR curates 198 biological ODE models covering signaling, metabolism, cell-cycle dynamics, and other biological processes across diverse organisms (Figure~\ref{fig:overview}).
To enable explicit mechanism verification, we provide reference annotations for all reactions and their kinetic rate laws, including mass action, Michaelis--Menten, and Hill dynamics~\citep{klipp2016systems, johnson2013century, weiss1997hill}.
SysBio-SR highlights two challenges in biological symbolic regression: (1) local kinetic-law selection (identifying the specific rate law for each reaction)~\citep{xu2023sbmlkinetics}, and (2) global mechanism identification (selecting ground-truth reaction network among trajectory-equivalent alternatives)~\citep{wong2023estimating,szederkenyi2011inference}.
\rev{These observations motivate us to explore how reaction priors and model selection can improve agreement with biological mechanisms while fitting the observed trajectories.}

To tackle these challenges, we introduce \textbf{KineSelect}, a biological symbolic regression framework that searches candidate reaction topologies, identifies their kinetic laws, and selects the optimal model by balancing trajectory fidelity with structural parsimony. Our contributions are summarized as follows:

\begin{itemize}
    \item
    We present SysBio-SR, the first reaction-annotated benchmark suite enabling mechanism recovery evaluation for biological ODE discovery, comprising 198 curated biological systems spanning mass action, Michaelis–Menten, Hill, and complex kinetics.

    \item
    We propose KineSelect, a kinetic selection framework that navigates expression spaces to biologically grounded kinetic rate laws, boosting aggregate OOD extrapolation \(R^2\) by 19.2\% and overall reaction F1 score by 14.3\% over the strongest baseline.

    \item
    Through evaluation of 11 SR methods across 198 systems, we expose a fundamental limitation of symbolic regression in systems biology: high trajectory fidelity drastically overestimates real-world utility by failing to recover ground-truth mechanisms.

\end{itemize}

%% file: sections/related_work.tex
\section{Related Work}

\subsection{Benchmarks for Symbolic Regression and ODE Discovery}
\label{subsec:related_benchmarks}

Existing SR benchmarks like
AI Feynman~\citep{udrescu2020ai, matsubara2022rethinking}, SRBench family~\citep{la2021contemporary, de2024srbench++, imai2025call}, and LLM-SRBench~\citep{shojaee2025llm} focus primarily on scalar physics-inspired or synthetic equations, with limited coverage of ODE systems.
Recent ODE benchmarks, including ODEBench~\citep{d2023odeformer}, MDBench~\citep{bideh2026mdbench}, and SymbolBench~\citep{liu2025can}, expand to multi-dimensional coupled systems (up to six variables), but draw from generic scientific domains and lack structural annotations for mechanism assessment.
In Table~\ref{tab:benchmark_comparison}, SysBio-SR fills this gap with 198 systems biology ODE models with reference reaction signatures for mechanism recovery.

\begin{table}[htbp]
\centering
\caption{\textbf{Comparison of symbolic regression benchmarks.}
\# Equations counts equations across all systems.
We count only SR equation-discovery instances.
Per-benchmark composition is detailed in Appendix~\ref{app:compared_benchmark_composition}.
Structural annot.\ indicates whether explicit structural annotations are provided (e.g., physical interaction terms or reaction rate laws).
\checkmark~= yes; $\times$~= no; $\sim$~= partial.}
\label{tab:benchmark_comparison}
\resizebox{\linewidth}{!}{%
\begin{tabular}{lrrrcccc}
\toprule
Benchmark & \# Systems & \# Equations & Max dim. & Domain & ODE & Coupled & Structural annot. \\
\midrule
AI Feynman~\citep{udrescu2020ai}                   & 120 & 120 & 1          & Physics        & $\times$   & $\times$   & $\times$   \\
SRBench~\citep{la2021contemporary}                 & 123 & 130 & 2          & Physics        & $\sim$     & $\times$   & $\times$   \\
EmpiricalBench~\citep{cranmer2023interpretable}    & 9   & 9   & 1          & Physics        & $\times$   & $\times$   & $\times$   \\
ODEBench~\citep{d2023odeformer}                    & 63  & 117 & 4          & Mixed          & \checkmark & \checkmark & $\times$   \\
LLM-SRBench~\citep{shojaee2025llm}                 & 239 & 239 & 1          & Phys.+Synth.   & $\sim$     & $\times$   & $\times$   \\
SymbolBench (SR task)~\citep{liu2025can}           & 156 & 383 & 4          & Mixed          & \checkmark & \checkmark & $\times$   \\
MDBench~\citep{bideh2026mdbench}                   & 77  & 141 & 6          & Mixed          & \checkmark & \checkmark & $\times$   \\
\midrule
\textbf{SysBio-SR} & \textbf{198} & \textbf{661} & \textbf{5} & \textbf{Sys.\ Biology} & \checkmark & \checkmark & \checkmark \\
\bottomrule
\end{tabular}%
}
\end{table}

\subsection{Symbolic Regression Methods for Dynamical Systems}

Symbolic regression methods for dynamical systems span sparse regression, evolutionary search, and neural symbolic approaches.
Sparse regression methods such as SINDy~\citep{brunton2016discovering}, including extensions for rational dynamics~\citep{kaheman2020sindy} and reaction networks~\citep{hoffmann2019reactive}, select active terms from predefined candidate libraries.
Evolutionary methods like genetic programming~\citep{koza1994genetic} and PySR~\citep{cranmer2023interpretable} explore broader expression spaces with complexity penalties, but frequently produce mathematically redundant or biologically uninterpretable expressions.
Stoichiometrically informed symbolic regression (SISR)~\citep{palma2026stoichiometrically} constrains search spaces by evolving candidate reaction mechanisms from time-series data;
however, its scope is limited to case studies without standardized mechanism recovery evaluation and broad kinetic coverage.
Recent neural symbolic approaches leveraging pre-trained transformers~\citep{kamienny2022end, d2023odeformer, vastl2024symformer, chen2025kinformer} and LLMs~\citep{holt2024data, shojaee2024llm, lirobust, xia2025sr} enable rapid equation inference, but remain sensitive to pre-training distributions and underexplored for biological mechanism recovery.

\begin{revision}
\subsection{Structural and Physical Priors in Symbolic Regression}
\label{subsec:structural_physical_priors}

Physical and structural information can restrict expression generation or guide search operations.
AI Feynman~\citep{udrescu2020ai} combines dimensional analysis with symmetry and separability tests on a neural approximation to simplify regression problems.
PhySO~\citep{tenachi2023deep} enforces dimensional consistency during expression generation.
RSSP~\citep{gong2026rssp} guides evolution and pruning using residual and structural sensitivities of expression subtrees, while StruSR~\citep{gong2025strusr} combines surrogate Taylor information with subtree attribution to guide expression evolution.
In their partial differential equation (PDE) settings, RSSP and StruSR use known differential operators to constrain searches for solution functions.

SISR~\citep{palma2026stoichiometrically} couples each reaction's rate law across the affected state equations through stoichiometric constraints.
KineSelect extends this representation with explicit regulator roles.
Within each proposed mechanism, it compares reaction-grouped mass-action, Michaelis--Menten, and Hill candidates through a joint fit to estimated state derivatives.
KineSelect explicitly couples kinetic law selection with reaction topology search, scoring candidate mechanisms on structural parsimony and trajectory fidelity.
SysBio-SR evaluates symbolic regression methods across diverse systems along two axes, trajectory fidelity and mechanism recovery.
\end{revision}

%% file: sections/dataset.tex
\section{The SysBio-SR Benchmark}
\label{sec:benchmark}

SysBio-SR contains 198 literature-derived, coupled biological ODE systems (661 equations) spanning diverse kinetics, with standardized symbolic targets, reference trajectories, and reaction annotations.
All 198 systems support trajectory evaluation, while 161 support mechanism recovery.

\subsection{Problem Formulation}
\label{subsec:task_formulation}

A system with $d$ state variables follows a coupled vector field
\begin{equation}
  \dot{\mathbf{x}}
  = \mathbf{f}(\mathbf{x};\boldsymbol{\theta})
  = \mathbf{S}\mathbf{v}(\mathbf{x};\boldsymbol{\theta}),
  \qquad \mathbf{x}(t)\in\mathbb{R}^{d},
  \label{eq:sysbio_ode}
\end{equation}
where $\boldsymbol{\theta}$ is the parameter vector, $\mathbf{S}\in\mathbb{R}^{d\times r}$ records how $r$ reaction channels change the state variables, and $\mathbf{v}$ contains the corresponding reaction-rate functions.
These rate laws include mass action, Michaelis--Menten, and Hill kinetics commonly used in systems biology models~\citep{johnson2013century,weiss1997hill}.
For system $m$, a symbolic regression method receives a fitting trajectory and returns a symbolic vector field for the coupled system:
\begin{equation}
  \mathcal{D}^{\mathrm{fit}}_m
  = \{(t_i,\mathbf{x}_m(t_i))\}_{i=1}^{N}
  \xrightarrow{\text{SR method}}
  \widehat{\mathbf{f}}_m.
  \label{eq:sysbio_task}
\end{equation}
Here, $N$ is the number of samples in the fitting trajectory.

\subsection{Dataset Construction and Composition}
\label{subsec:dataset_construction}

\begin{figure}[htbp]
  \centering
  \includegraphics[width=\linewidth]{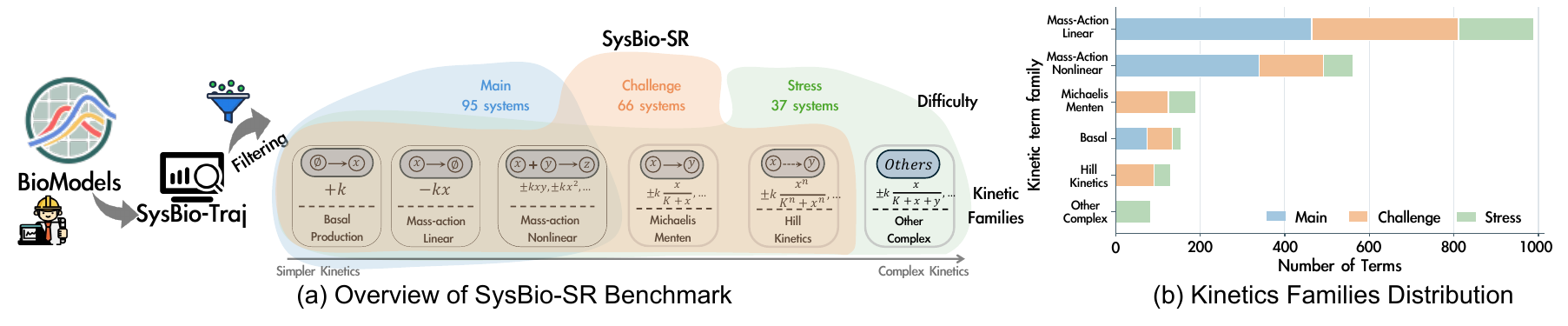}
  \vspace{-0.3in}
  \caption{\textbf{Construction and kinetic composition of SysBio-SR.}
  (a) Curation pipeline partitions systems into three kinetics-based subsets: Main (mass action only; $n=95$), Challenge (with Michaelis--Menten or Hill kinetics; $n=66$), and Stress (with complex non-canonical kinetics; $n=37$).
  (b) Distribution of reaction terms across kinetic families for each subset.
  }
  \label{fig:overview_sysbiosr}
\end{figure}

\textbf{Source models and curation.}
SysBio-Traj~\citep{rao2026regime} collects literature-derived BioModels entries encoded in Systems Biology Markup Language (SBML)~\citep{hucka2003systems,malik2020biomodels,luders2022odebase}.
From this collection, we construct SysBio-SR by retaining deterministic, autonomous ODE systems with at most five state variables, while excluding event-driven or piecewise dynamics, explicit time dependence, and systems that fail numerical validation.
This process yields 198 systems with 661 equations.
Figure~\ref{fig:overview_sysbiosr}(a) summarizes this curation pipeline.

\textbf{Standardization and trajectories.}
For each system, we rename states as $x_0,\ldots,x_{d-1}$ and parameters as $c_0,\ldots,c_{p-1}$, and provide a standardized symbolic ODE representation.
We generate a fitting trajectory with \(N=512\) samples from the default initial condition.
For out-of-distribution (OOD) extrapolation, we generate a second 512-point trajectory from a perturbed initial condition while holding the vector field and parameters fixed.

\textbf{Composition.}
SysBio-SR covers systems with one to five state variables with diverse reaction topologies and kinetic laws.
We partition the 198 systems into three subsets based on maximum kinetic complexity: Main (95 systems, exclusively mass action kinetics), Challenge (66 systems containing Michaelis--Menten or Hill kinetics), and Stress (37 systems, containing complex non-canonical kinetics). Higher complexity subsets may also include reactions from lower kinetic tiers. Figure~\ref{fig:overview_sysbiosr}(b) details the kinetic-family composition across subsets, while Appendix Figure~\ref{fig:sysbiosr_statistics} summarizes distributions for system dimension, reaction count, and species per reaction.

\subsection{Reaction Annotations}
\label{subsec:reaction_annotations}

We provide reaction annotations for all 198 systems.
Starting from the reference ODEs, we decompose each RHS into signed terms, separate rate coefficients from kinetic factors, and label the species-dependent kinetic terms.
The resulting records specify consumption or production, participating species, and kinetic forms such as mass action, Michaelis--Menten, and Hill terms.
We assess annotation consistency through reconstruction checks and manual review.
The annotations provide reference signatures for mechanism-recovery evaluation on the 161 Main and Challenge systems.
The 37 Stress systems retain annotations but include kinetic forms outside the matching protocol.
Appendix~\ref{app:reaction_matching} details annotation construction, representation, and signature matching.

%% file: sections/evaluation.tex
\section{Mechanism-Aware Evaluation Protocol}
\label{sec:evaluation}

\textbf{Evaluation axes.}
We evaluate discovered ODE systems along two axes: trajectory fidelity and mechanism recovery.
Trajectory fidelity is measured on all 198 systems via in-distribution (ID) reconstruction and out-of-distribution (OOD) extrapolation~\citep{d2023odeformer} from perturbed initial conditions.
Mechanism recovery is evaluated on the Main and Challenge subsets by matching predicted terms against reference reaction signatures. To prevent data leakage, reaction annotations are strictly withheld during model fitting and selection, serving exclusively for post-hoc evaluation.

\textbf{Trajectory metrics.}
For each system, we numerically integrate the recovered ODE from both ID and OOD evaluation initial conditions and compare with reference trajectory.
We compute state-averaged coefficients of determination \(R_m^2\) across the evaluation time grid and aggregate them as
\begin{equation}
  R^2\text{-score}
  = \frac{1}{M}\sum_{m=1}^{M}\max\!\left(R_m^2,0\right),
  \label{eq:trajectory_r2_score}
\end{equation}
where \(M\) is the system count in the target set.
System trajectories that diverge, produce non-finite values, or encounter solver integration timeouts receive $R_m^2 = 0$.
We additionally report \(\mathrm{Acc}@.90\), the fraction of systems with \(R_m^2\geq0.90\).
Following ODEFormer~\citep{d2023odeformer}, expression complexity is evaluated as the total symbolic tree node count summed across state equations.
Model selection parsimony is defined in Section~\ref{subsec:fpms}, with full metric specifications in Appendix~\ref{app:metric_contract}.

\textbf{Reaction metrics.}
Equivalent ODE expressions can take different forms~\citep{d2023odeformer}.
We therefore match recovered reactions against reference mechanisms through canonical reaction signatures (Appendix~\ref{app:reaction_matching}).
A predicted signature is counted as a true positive if its reaction topology and kinetic law strictly match an annotated reaction, independent of numerical parameter values.
Unmatched predicted and reference signatures constitute false positives and false negatives, respectively.
We compute precision, recall, and F1 for each system and report their means across systems.

\textbf{Observation noise.}
To evaluate robustness under noisy measurements, we corrupt clean fitting trajectories with independent state-wise Gaussian noise:
\begin{equation}
  \widetilde{x}_i(t)
  =
  x_i(t)+\sigma_{\mathrm{rel}}\,\sigma_i\epsilon_{i,t},
  \qquad
  \epsilon_{i,t}\sim\mathcal{N}(0,1),
  \qquad
  \sigma_{\mathrm{rel}}\in\{0.001,0.01,0.05,0.1\}.
  \label{eq:observation_noise}
\end{equation}
Here, $\sigma_i$ denotes the temporal standard deviation of state $i$ on the clean trajectory to adjust for varying state magnitudes. The relative noise magnitude $\sigma_{\mathrm{rel}}=0$ represents noiseless baseline data. Regardless of the noise level, all discovered systems are strictly evaluated against clean ID reconstruction and OOD extrapolation reference trajectories.
Appendix~\ref{app:id_ood_noise} specifies the full protocol.

\textbf{Evaluated baselines.}
We evaluate 11 symbolic discovery algorithms (12 total configurations) across four methodological families.
(1) Symbolic search and genetic programming: EPLEX (2 configs)~\citep{la2016epsilon}, FE-AFP~\citep{schmidt2010age,la2021contemporary}, FFX~\citep{mcconaghy2011ffx}, and PySR~\citep{cranmer2023interpretable}.
(2) Sparse system identification methods: SINDy~\citep{brunton2016discovering}, SINDy-PI~\citep{kaheman2020sindy}, and Reactive SINDy~\citep{hoffmann2019reactive}.
(3) Neural \& language models: ODEFormer~\citep{d2023odeformer} and LLM-SR~\citep{shojaee2024llm}.
(4) Reaction-structured methods: SISR~\citep{palma2026stoichiometrically} and our KineSelect.
All configurations undergo trajectory evaluation. Mechanism recovery is evaluated exclusively on Reactive SINDy, SISR, and KineSelect by mapping their reaction-structured outputs to reference reaction signatures.
Implementation details and method specific configurations are provided in
Appendix~\ref{app:reproducibility}.

\textbf{KineSelect component studies.}
To evaluate architectural choices in KineSelect, we compare standard $\alpha$-threshold (Appendix~\ref{app:alpha_threshold}) against Fidelity--Parsimony Model Selection (FPMS) on identical candidate archives. Additionally, component ablations isolate the individual contributions of derivative loss and specialized kinetic complexity across both trajectory fidelity and mechanism recovery.

%% file: sections/method.tex
\section{KineSelect}
\label{sec:method}

KineSelect structures biological symbolic regression into three core components: Reaction Structure Search (RSS), Kinetic Form Identification (KFI), and Fidelity–Parsimony Model Selection (FPMS).
In the search loop, RSS proposes regulator-guided reaction topologies, while KFI identifies their kinetic forms and evaluates the resulting mechanisms. The search maintains an archive of non-dominated candidate mechanisms, from which FPMS selects the final mechanism by balancing fidelity and parsimony.

\subsection{Reaction Structure Search}
\label{subsec:method_overview}

\begin{figure}[htbp]
\centering
\includegraphics[width=\linewidth]{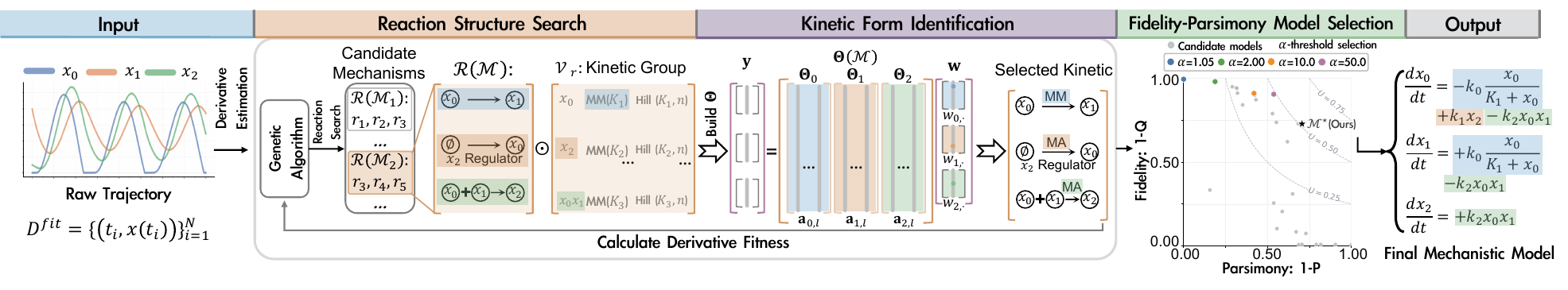}
\vspace{-0.3in}
\caption{\textbf{Overview of the KineSelect framework.}
Reaction Structure Search proposes reaction topologies with reactants, products, and optional regulators. Kinetic Form Identification jointly assigns kinetic expressions across reaction groups and evaluates candidates using derivative loss. Fidelity–Parsimony Model Selection selects the optimal mechanism from the candidate archive by balancing trajectory fidelity and structural parsimony.}
\label{fig:kineselectsr}
\end{figure}

Consistent with Eq.~\eqref{eq:sysbio_ode}, we represent a candidate
mechanistic model $\mathcal{M}$ with reaction set $\mathcal{R}(\mathcal{M})$ as
$\dot{\mathbf{x}}
= \sum_{r\in\mathcal{R}(\mathcal{M})}
\mathbf{s}_r v_r(\mathbf{x};\boldsymbol{\theta}_r)$,
where $\mathbf{s}_r\in\mathbb{Z}^d$ is the stoichiometric vector of reaction $r$, $v_r$ is its kinetic rate law, and $\boldsymbol{\theta}_r$ contains the overall rate constant $k_r$ and any kinetic shape parameters.
Building on the reaction-level evolutionary representation of SISR~\citep{palma2026stoichiometrically}, RSS extends structural search by incorporating regulator roles.
For each proposed reaction topology, RSS assigns reactant, product, and regulator species. Regulators exert zero net stoichiometry ($\mathbf{s}_{r,i} = 0$) but directly modulate the rate law $v_r$ (e.g., regulator \(x_2\) in \(x_0 \xrightarrow{x_2} x_1\) with rate \(v=kx_0x_2\)).
Directly encoding discrete kinetic expressions into each reaction creates a combinatorial search explosion over topological and functional spaces.
To resolve this, RSS restricts its search to reaction topologies and species roles, delegating rate-law selection to KFI during candidate evaluation.
KineSelect maintains separate subpopulations partitioned by model size $c = \vert{}\mathcal{R}(\mathcal{M})\vert{} \in \{1,\dots,c_{\max}\}$, pooling kinetic-resolved candidates into a shared archive $\mathcal{A}$ for selection by FPMS (Figure~\ref{fig:kineselectsr}).

\subsection{Kinetic Form Identification}
\label{subsec:rgks}

Conditioned on the reaction topologies proposed by RSS, KFI resolves the functional form of each rate law within the candidate evaluation loop.
For every proposed reaction $r \in \mathcal{R}(\mathcal{M})$, KFI constructs a candidate library of unit-rate expressions $\mathcal{V}_r = \{\bar{v}_{r,\ell}\}$, where the bar indicates that each candidate excludes rate constants \(k_r\).
Each candidate factorizes into reactant and regulator terms:
\begin{equation}
\bar{v}_{r,\ell}(\mathbf{x}) = \phi_{r,\ell}(\mathbf{x})\,\psi_{r,\ell}(\mathbf{x}).
\label{eq:kinetic_factorization}
\end{equation}
Here, $\phi_{r,\ell}$ models mass action or Michaelis--Menten reactant kinetics, while $\psi_{r,\ell}$ models optional regulatory influences (Explicit functional forms are given in Appendix~\ref{app:kinetic_library}).
Non-linear parameters (e.g., affinity constants or Hill coefficients) are discretized over predefined grid settings, yielding discrete elements in $\mathcal{V}_r$.

To evaluate all kinetic configurations across the system simultaneously,
KFI constructs a reaction-block dictionary from these candidates.
Let $\mathbf{X} \in \mathbb{R}^{N \times d}$ denote the observed trajectory matrix
and $\dot{\mathbf{X}} \in \mathbb{R}^{N \times d}$ its estimated time derivatives.
To prevent state variables with large magnitudes from dominating model selection,
we define a scale-normalization matrix $\Gamma = \operatorname{diag}(\gamma_0,\dots,\gamma_{d-1})$
with $\gamma_j = \max_{1 \le n \le N} |\dot{X}_{nj}|$
for $j=0,\dots,d-1$,
setting the target vector to $\mathbf{y} = \operatorname{vec}(\dot{\mathbf{X}}\Gamma^{-1}) \in \mathbb{R}^{Nd}$.

For each candidate expression $\bar{v}_{r,\ell} \in \mathcal{V}_r$,
define its stoichiometrically coupled dictionary column as:
\begin{equation}
\mathbf{a}_{r,\ell}
=
\frac{\tilde{\mathbf{a}}_{r,\ell}}{\|\tilde{\mathbf{a}}_{r,\ell}\|_2},
\qquad
\tilde{\mathbf{a}}_{r,\ell}
=
\operatorname{vec}\left(\bar{v}_{r,\ell}(\mathbf{X})\,\mathbf{s}_r^{\top}\Gamma^{-1}\right)
\in \mathbb{R}^{Nd}.
\label{eq:kfi_column}
\end{equation}
Grouping columns by reaction yields the reaction-block dictionary
$\boldsymbol{\Theta}_r = [\mathbf{a}_{r,1} \mid \dots \mid \mathbf{a}_{r,|\mathcal{V}_r|}]$,
forming the system-wide dictionary $\boldsymbol{\Theta}(\mathcal{M}) = [\boldsymbol{\Theta}_r]_{r \in \mathcal{R}(\mathcal{M})}$.
Unlike standard SINDy~\citep{brunton2016discovering},
which fits state equations independently,
Eq.~\eqref{eq:kfi_column} enforces stoichiometry vector \(\mathbf{s}_r\) across all affected state derivatives
while allowing zero-net regulators ($\mathbf{s}_{r,i}=0$)
to modulate rates without violating mass conservation.

KFI jointly evaluates competing rate laws
by solving a non-negative regularized regression over the block dictionary:
\begin{equation}
\widehat{\mathbf{w}}
=
\operatorname*{arg\,min}_{\mathbf{w}\geq 0}
\left\| \mathbf{y} - \boldsymbol{\Theta}(\mathcal{M})\mathbf{w} \right\|_2^2
+ \lambda\|\mathbf{w}\|_2^2,
\qquad \lambda \geq 0.
\label{eq:kfi_joint_fit}
\end{equation}
From the fitted weight vector $\widehat{\mathbf{w}}$,
KFI selects the dominant candidate within each reaction group:
\begin{equation}
\ell_r^{\star}
=
\operatorname*{arg\,max}_{\ell \in \{1,\dots,|\mathcal{V}_r|\}} \widehat{w}_{r,\ell}.
\label{eq:kfi_score}
\end{equation}
Finally, KFI refits the rate constants $k_r \geq 0$
for the selected expressions $v_r(\mathbf{x}) = k_r \bar{v}_{r,\ell_r^\star}(\mathbf{x})$
against derivative target $\mathbf{y}$,
returning the resolved model and derivative loss to the evolutionary search loop.

\subsection{Fidelity--Parsimony Model Selection}
\label{subsec:fpms}

Post-search, FPMS identifies the final mechanism from the candidate archive $\mathcal{A}$ by evaluating candidates across four complementary criteria.
Trajectory fidelity is captured by combining derivative loss $\mathcal{L}_i^{\mathrm{der}}$ and numerical trajectory simulation loss $\mathcal{L}_i^{\mathrm{sim}}$.
To compress heavy-tailed loss distributions, both fidelity metrics undergo log-transformation prior to normalization.
Appendix~\ref{app:fpms_contract} specifies the transformation and numerical policies.

Structural parsimony is evaluated using two distinct complexity measures. (1) Expanded tree complexity ($C_i^{\mathrm{tree}}$): The total expression-tree node count obtained after expanding reaction rate laws into full state-level ODEs~\citep{palma2026stoichiometrically}.
(2) Specialized kinetic complexity ($C_i^{\mathrm{spec}}$): The count of non-mass-action kinetic terms (i.e., Michaelis--Menten reactant terms and Hill regulatory expressions), penalizing complex rate expressions independent of stoichiometric fan-out (how many state equations the reaction affects).

All four criteria are independently min--max normalized over $\mathcal{A}$ (denoted by $\widetilde{\cdot} \in [0,1]$).
FPMS aggregates these into a composite lack-of-fidelity score $Q_i$ and a model-complexity score $P_i$:
\begin{equation}
Q_i = \frac{\widetilde{\mathcal{L}}_i^{\mathrm{der}} + \widetilde{\mathcal{L}}_i^{\mathrm{sim}}}{2},
\qquad
P_i = \frac{\widetilde{C}_i^{\mathrm{tree}} + \widetilde{C}_i^{\mathrm{spec}}}{2}.
\label{eq:fpms_criteria}
\end{equation}
Lower \(Q_i\) and \(P_i\) indicate higher fidelity and lower model complexity, respectively.

FPMS selects the optimal model $\mathcal{M}^{\star} \in \mathcal{A}$ by maximizing a multiplicative fidelity--parsimony utility:
\begin{equation}
U_i = (1 - Q_i)(1 - P_i), \qquad
\mathcal{M}^{\star} = \operatorname*{arg\,max}_{\mathcal{M}_i \in \mathcal{A}} U_i.
\label{eq:fpms_utility}
\end{equation}
Unlike additive scoring, this multiplicative objective enforces strict balance: $U_i \to 0$ whenever either fidelity or parsimony is severely compromised ($Q_i \to 1$ or $P_i \to 1$), discouraging the selection of accurate yet overly complex models or simple yet inaccurate mechanisms.

%% file: sections/experiments.tex
\section{Experimental Results}
\label{sec:experiments}

\subsection{Benchmark-Wide Trajectory Recovery}
\label{subsec:trajectory_results}

KineSelect achieves the highest trajectory fidelity under both ID reconstruction and OOD extrapolation (Table~\ref{tab:trajectory_main}).
Relative to the strongest baseline SISR, KineSelect improves overall ID and OOD \(R^2\) by 16.5\% (0.770 vs.\ 0.661) and 19.2\% (0.508 vs.\ 0.426), respectively, while increasing OOD \(\mathrm{Acc}@.90\) by 14.0\% (0.374 vs.\ 0.328).
The concordant gains in average $R^2$ and high-fidelity coverage ($\text{Acc}@.90$) confirm that these improvements reflect robust, system-wide recovery rather than outlier-driven averages.
OOD extrapolation serves as a crucial differentiator among methods with similar ID fitting performance. For instance, while FFX and SISR display similar ID $R^2$, SISR has substantially higher OOD $R^2$, revealing greater dynamical generalization (Table~\ref{tab:trajectory_main}). KineSelect maintains the OOD advantage across all three subsets: relative to the strongest comparator in each regime, it improves OOD $R^2$ by 19.2\% on Main (0.601 vs. 0.504), 20.1\% on Challenge (0.437 vs. 0.364), and 8.2\% on Stress (0.395 vs. 0.365). Stress is the most challenging subset, combining the lowest absolute OOD $R^2$ with the smallest relative gain. A systematic decline from ID to OOD performance is observed across all three subsets for all methods.

\rev{Across multiple search seeds, KineSelect retains higher mean OOD \(R^2\) and reaction F1 than SISR, including under shared FPMS (Appendix~\ref{app:clean_multirun_results}).}

This higher trajectory fidelity is accompanied by increased expression complexity relative to SISR (62.13 vs. 37.66 nodes). This reflects KineSelect's capacity to express the non-mass-action kinetic rate laws (e.g., Hill equation) necessary for faithful dynamics.

\input{Table/trajectory_main}

\subsection{Trajectory Fidelity and Mechanism Recovery}
\label{subsec:trajectory_reaction}

\begin{figure}[htbp]
  \centering
  \includegraphics[width=\linewidth]{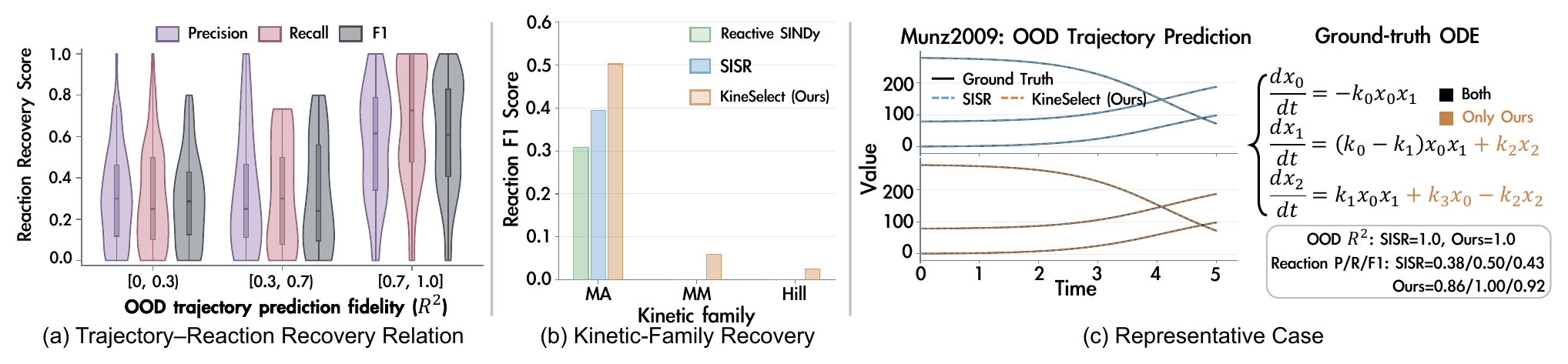}
  \vspace{-0.3in}
  \caption{\textbf{Trajectory fidelity vs. mechanism recovery.}
  (a) KineSelect reaction precision, recall, and F1 distributions by OOD \(R^2\) intervals.
  (b) Reaction F1 by kinetic family.
  (c) Predicted trajectories for the Munz2009 system under SISR (top) and KineSelect (bottom) against ground truth. Black and orange equation terms denote recovery by both methods and exclusively by KineSelect, respectively. }
  \label{fig:trajectory_reaction}
\end{figure}

High trajectory fidelity does not guarantee true mechanism recovery.
In Figure~\ref{fig:trajectory_reaction}(a), reaction precision, recall, and $F_1$ score distributions remain broadly spread even within the highest OOD $R^2$ regime, confirming that accurate curve-fitting can coexist with incorrect or incomplete reaction recovery.
See Appendix~\ref{app:high_fidelity_mechanism} for detailed F1 at OOD \(R^2\geq0.90\).
Furthermore, trajectory metrics mask severe performance gaps across kinetic families (Figure~\ref{fig:trajectory_reaction}(b)). While trajectory fits remain high, reaction $F_1$ scores drop sharply for Michaelis--Menten and Hill kinetics compared to mass action due to increasing functional complexity.
The Munz2009 system~\citep{munz2009zombies} illustrates this divergence (Figure~\ref{fig:trajectory_reaction}(c)):
while methods achieve a perfect OOD \(R^2\) of 1.00,
KineSelect attains a reaction F1 of 0.92, compared with 0.43 for SISR.
See Appendix~\ref{app:representative_cases} for additional case studies.

\subsection{Mechanism Recovery Across Subsets and Kinetic Families}
\label{subsec:kinetic_recovery}

KineSelect ranks first in precision, recall, and F1 across the Overall, Main, and Challenge evaluations (Table~\ref{tab:reaction_main}).
Relative to SISR, it improves Overall reaction F1 by 14.3\% (0.447 vs.\ 0.391), with the gain persisting on Challenge (0.285 vs.\ 0.225).
The large gain in recall compared to precision (\(+0.066\), versus \(+0.029\)) indicates that KineSelect captures a larger proportion of true reactions per system while still reducing false positives.
All three methods suffer performance drops on the Challenge subset, highlighting the inherent difficulty of mechanism recovery.

Mechanism recovery varies significantly across kinetic families (Figure~\ref{fig:trajectory_reaction}(b)). On mass action dynamics, KineSelect raises reaction $F_1$ to $0.503$ (vs. $0.395$ for SISR). For non-mass-action kinetics, baseline zero scores (Reactive SINDy and SISR) reflect expressivity limits rather than search failure, as neither framework includes rational functional terms in its candidate basis~\citep{hoffmann2019reactive,palma2026stoichiometrically}. In contrast, only KineSelect achieves non-zero recovery ($F_1 = 0.059$ for Michaelis--Menten and $0.025$ for Hill). While these low scores highlight the difficulty of identifying complex rate laws from trajectory data, they demonstrate that basis expressivity is a strict prerequisite for non-mass-action discovery. Kinetic-family evaluations provide a crucial diagnostic tool alongside trajectory metrics (see Appendix~\ref{app:kinetic_family_recovery} for detailed precision--recall breakdowns).

\input{Table/reaction_main}

\subsection{Component Ablations and Noise Robustness}
\label{subsec:selection_robustness}

\begin{figure}[htbp]
  \centering
  \includegraphics[width=\linewidth]{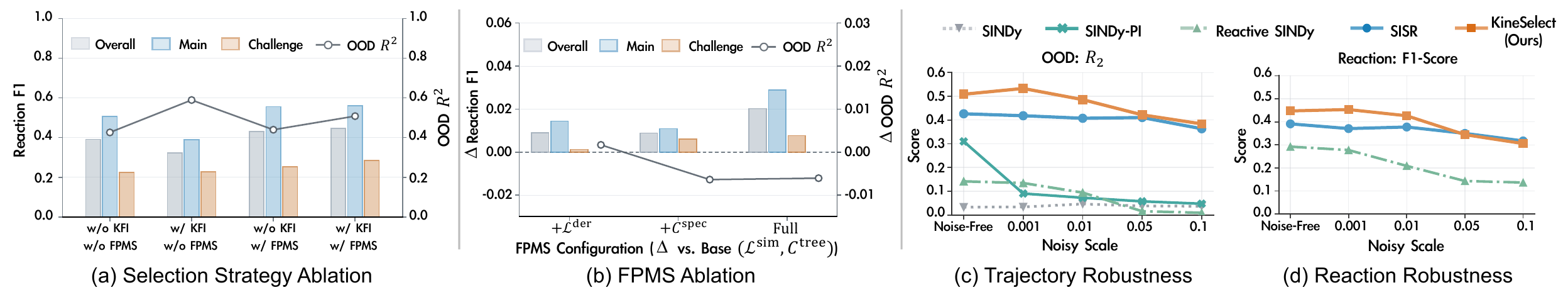}
  \vspace{-0.3in}
  \caption{\textbf{KineSelect component ablations and noise robustness.}
  (a) KFI and FPMS ablations of KineSelect, with \(\alpha\)-threshold selection used when FPMS is disabled.
  (b) Performance changes \(\Delta\) when adding derivative loss and specialized kinetic complexity criteria individually or jointly to the base FPMS criteria (simulation loss and expression-tree complexity).
  (c,d) OOD extrapolation \(R^2\) and reaction F1 under increasing relative noise levels.}
  \label{fig:selection_noise}
  \label{fig:noisy_robustness}
\end{figure}

Figure~\ref{fig:selection_noise}(a) shows that the reaction-recovery benefit of KFI depends critically on the final model selection strategy.
Under a standard \(\alpha\)-threshold rule, enabling KFI improves OOD \(R^2\) (0.598 vs.\ 0.426) but reduces reaction F1 (0.323 vs.\ 0.391).
Applying FPMS elevates KFI reaction F1 from 0.323 to 0.447, yielding a 3.7\% gain in F1 and a 15.5\% increase in OOD \(R^2\) for the main subset.
Sweeping \(\alpha\) further reveals divergent optima for trajectory fit versus mechanism recovery (Appendix~\ref{app:selection_sensitivity}).

Ablating individual FPMS criteria demonstrates their additive value (Figure~\ref{fig:selection_noise}(b)).
Starting from the base criteria using simulation loss and expression-tree complexity, introducing either derivative loss or specialized kinetic complexity improves overall reaction F1. Combining all four criteria achieves peak F1 across the Overall, Main, and Challenge benchmarks while preserving OOD trajectory fit. Crucially, reaction F1 provides fine-grained discrimination between criteria variants where the OOD trajectory scores remain indistinguishable. KineSelect adopts this complete four-criterion configuration (see Appendix~\ref{app:selection_sensitivity} for complete sensitivity and ablation results).

Finally, evaluating robustness across relative noise levels shows that KineSelect substantially outperforms SISR in both trajectory fit and reaction recovery at low-to-moderate noise ($\sigma_{\mathrm{rel}} \le 0.01$; Figure~\ref{fig:selection_noise}(c,d)). As noise increases ($\sigma_{\mathrm{rel}} \ge 0.05$), the performance gap narrows, yet KineSelect retains superior OOD trajectory extrapolation ($R^2 = 0.421$ vs.\ $0.410$ at $\sigma_{\mathrm{rel}}=0.05$; $0.383$ vs.\ $0.363$ at $\sigma_{\mathrm{rel}}=0.10$).
\rev{Results across different noise seeds support the same low-noise advantage in mean OOD \(R^2\) and reaction F1 (Appendix~\ref{app:noisy_multirun_results}).
Under shared FPMS, KineSelect retains higher mean OOD \(R^2\) at all four noise levels, while reaction F1 means become close at 5\% and 10\% noise.
Appendix~\ref{app:ood_noise_distributions} shows OOD score distributions across systems at each noise level.}

%% file: Table/trajectory_main.tex
\begin{table}[htbp]
  \centering
  \caption{\textbf{Trajectory fidelity on SysBio-SR.}
  Expression complexity, \(R^2\), and \(\mathrm{Acc}@.90\)
  under ID reconstruction and OOD extrapolation for the full benchmark and its subsets. \(\mathrm{Acc}@.90\) denotes the fraction of systems with \(R^2\geq0.90\). Bold = best; underlined = second-best.
  }
  \label{tab:trajectory_main}
  \begingroup
  \footnotesize
  \setlength{\tabcolsep}{1.5pt}
  \renewcommand{\arraystretch}{1}
  \newcolumntype{Y}{>{\centering\arraybackslash}X}
  \begin{tabularx}{\linewidth}{@{}l c *{10}{Y}@{}}
    \toprule
    \multirow{3}{*}{\textbf{Method}}
      & \multicolumn{5}{c}{\textbf{Overall}}
      & \multicolumn{2}{c}{\textbf{Main}}
      & \multicolumn{2}{c}{\textbf{Challenge}}
      & \multicolumn{2}{c}{\textbf{Stress}} \\
    \cmidrule(lr){2-6}\cmidrule(lr){7-8}\cmidrule(lr){9-10}\cmidrule(lr){11-12}
      & \multirow{2}{*}{\textbf{Comp.}\(\downarrow\)}
      & \multicolumn{2}{c}{\(R^2\)} & \multicolumn{2}{c}{Acc.}
      & \multicolumn{2}{c}{\(R^2\)}
      & \multicolumn{2}{c}{\(R^2\)}
      & \multicolumn{2}{c}{\(R^2\)} \\
      & & ID & OOD & ID & OOD
      & ID & OOD & ID & OOD & ID & OOD \\
    \midrule
    EPLEX (default) & 138.03 & 0.014 & 0.014 & 0.005 & 0.010 & 0.011 & 0.008 & 0.024 & 0.030 & 0.000 & 0.000 \\
    EPLEX (poly) & 84.48 & 0.086 & 0.050 & 0.056 & 0.030 & 0.105 & 0.058 & 0.082 & 0.056 & 0.042 & 0.017 \\
    FE-AFP & 51.82 & 0.123 & 0.094 & 0.106 & 0.071 & 0.161 & 0.121 & 0.103 & 0.081 & 0.061 & 0.049 \\
    FFX & 45.92 & 0.658 & 0.367 & 0.525 & 0.222 & \underline{0.755} & 0.402 & 0.592 & \underline{0.364} & 0.529 & 0.280 \\
    PySR & \textbf{27.27} & 0.342 & 0.233 & 0.303 & 0.177 & 0.488 & 0.332 & 0.234 & 0.174 & 0.161 & 0.086 \\
    \midrule
    SINDy & 83.62 & 0.038 & 0.033 & 0.035 & 0.030 & 0.035 & 0.018 & 0.063 & 0.075 & 0.000 & 0.000 \\
    SINDy-PI & 159.86 & 0.391 & 0.310 & 0.379 & 0.273 & 0.491 & 0.393 & 0.333 & 0.261 & 0.241 & 0.182 \\
    Reactive SINDy & 46.95 & 0.166 & 0.142 & 0.101 & 0.091 & 0.246 & 0.220 & 0.094 & 0.068 & 0.089 & 0.074 \\
    \midrule
    ODEFormer & \underline{30.23} & 0.215 & 0.136 & 0.111 & 0.061 & 0.270 & 0.158 & 0.167 & 0.118 & 0.161 & 0.112 \\
    LLM-SR & 85.26 & 0.313 & 0.222 & 0.288 & 0.182 & 0.403 & 0.299 & 0.224 & 0.129 & 0.242 & 0.191 \\
    \midrule
    SISR & 37.66 & \underline{0.661} & \underline{0.426} & \underline{0.561} & \underline{0.328} & 0.737 & \underline{0.504} & \underline{0.594} & 0.347 & \underline{0.587} & \underline{0.365} \\
    KineSelect & 62.13 & \textbf{0.770} & \textbf{0.508} & \textbf{0.662} & \textbf{0.374} & \textbf{0.860} & \textbf{0.601} & \textbf{0.709} & \textbf{0.437} & \textbf{0.647} & \textbf{0.395} \\
    \bottomrule
  \end{tabularx}
  \endgroup
\end{table}

%% file: Table/reaction_main.tex
\begin{table}[htbp]
  \centering
  \caption{\textbf{Reaction recovery on SysBio-SR.}
  Mean precision, recall, and F1 scores across systems for three methods that output reaction structures. Bold = best; underlined = second-best.}
  \label{tab:reaction_main}
  \begingroup
  \footnotesize
  \setlength{\tabcolsep}{1.5pt}
  \renewcommand{\arraystretch}{1}
  \newcolumntype{Y}{>{\centering\arraybackslash}X}
  \begin{tabularx}{\linewidth}{@{}l *{9}{Y}@{}}
    \toprule
    \multirow{2}{*}{\textbf{Method}}
      & \multicolumn{3}{c}{\textbf{Overall}}
      & \multicolumn{3}{c}{\textbf{Main}}
      & \multicolumn{3}{c}{\textbf{Challenge}} \\
    \cmidrule(lr){2-4}\cmidrule(lr){5-7}\cmidrule(lr){8-10}
      & Precision & Recall & F1 & Precision & Recall & F1 & Precision & Recall & F1 \\
    \midrule
    Reactive SINDy & 0.320 & 0.370 & 0.292 & 0.383 & 0.477 & 0.363 & 0.228 & 0.215 & 0.191 \\
    SISR & \underline{0.409} & \underline{0.416} & \underline{0.391} & \underline{0.528} & \underline{0.543} & \underline{0.507} & \underline{0.239} & \underline{0.234} & \underline{0.225} \\
    KineSelect & \textbf{0.438} & \textbf{0.482} & \textbf{0.447} & \textbf{0.544} & \textbf{0.606} & \textbf{0.560} & \textbf{0.285} & \textbf{0.305} & \textbf{0.285} \\
    \bottomrule
  \end{tabularx}
  \endgroup
\end{table}

%% file: sections/conclusion.tex
\section{Conclusion and Limitations}

This work introduced SysBio-SR, a benchmark of 198 literature-based biological ODE systems designed to evaluate symbolic regression on both trajectory fidelity and mechanism recovery.
Crucially, our benchmark reveals that the two objectives are decoupled: high trajectory accuracy does not guarantee true mechanistic discovery.
KineSelect achieves superior OOD extrapolation \(R^2\) and overall reaction F1 compared to strong baselines like SISR, while producing more complex expressions.

The SysBio-SR benchmark focuses on low-dimensional, deterministic ODE systems evaluated on simulated trajectories from curated reference models.
\rev{For each system, SysBio-SR provides a single training trajectory, which may cover too little of the state space to distinguish among competing mechanisms. Future work could examine how fitting a model jointly to trajectories from multiple initial conditions affects mechanism recovery.}
\rev{Our mechanism-recovery evaluation currently covers three reaction-based methods. Extending this evaluation to generic SR methods requires consistent expression normalization and mapping to reaction signatures.}
Future work will investigate reliable recovery of reactions involving Michaelis--Menten or Hill kinetics, as well as extend the benchmark to higher-dimensional systems and more complex kinetic mechanisms.
We recommend assessing biological symbolic regression through both trajectory fidelity and explicit mechanistic agreement.

%% file: appendix.tex
\input{supplementary/benchmark_details}
\input{supplementary/method_details}
\input{supplementary/experiment_details}

\input{supplementary/reproducibility_details}

%% file: supplementary/benchmark_details.tex
\clearpage
\section{SysBio-SR Benchmark and Evaluation Details}
\label{app:benchmark_details}

\subsection{Benchmark Composition}
\label{app:benchmark_composition}

\begin{figure}[htbp]
  \centering
  \includegraphics[width=\linewidth]{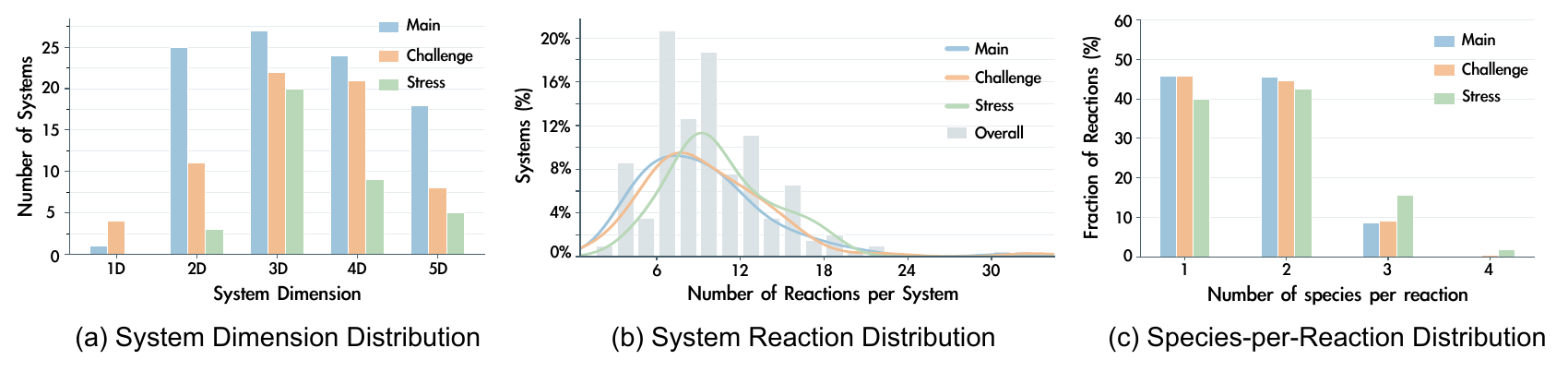}
  \caption{\textbf{Structural composition of SysBio-SR.}
  (a) Number of systems by state dimension and subset.
  (b) Distribution of reactions per system; gray bars show the overall distribution, and colored curves show the distributions within each subset.
  (c) Distribution of the number of species involved per reaction, by subset.}
  \label{fig:sysbiosr_statistics}
\end{figure}

Figure~\ref{fig:sysbiosr_statistics} summarizes system size and reaction structure across all 198 systems.
State dimension and reaction count span overlapping ranges across the three subsets (Figure~\ref{fig:sysbiosr_statistics}(a,b)).
The subsets distinguish kinetic form while retaining variation in system size.
In Figure~\ref{fig:sysbiosr_statistics}(c), each species counts once across reactants, products, and the reaction-rate expression, including regulators.
Percentages are normalized over all reactions within each subset.

The kinetic-family term counts appear in Figure~\ref{fig:overview_sysbiosr}(b).
Basal denotes zero-order terms \(k\), Mass-Action Linear denotes first-order terms \(kx_i\), and Mass-Action Nonlinear denotes higher-order monomials \(k\prod_i x_i^{a_i}\) of total degree at least two.
The remaining families are Michaelis--Menten, Hill, and other nonlinear terms.

\subsection{Benchmark Organization and Evaluation Scope}
\label{app:benchmark_organization}

To pin down what enters each evaluation, we organize the release around the three kinetics-based subsets of Section~\ref{subsec:dataset_construction} and a per-system file schema.
The three subsets contain 95 Main, 66 Challenge, and 37 Stress systems, totaling 198 systems.
All 198 systems have reaction annotations and enter trajectory evaluation.
The reaction-evaluation subset comprises the 161 Main and Challenge systems.
The 37 Stress systems are excluded from reaction scoring.

The per-system release format is defined as follows:

\begin{verbatim}
benchmark/
|-- benchmark.yaml
|-- main/                    (95 systems)
|   |-- BIOMD0000000006/     (example: Tyson1991)
|   |   |-- Tyson1991.xml
|   |   |-- equation.json
|   |   |-- model.py
|   |   |-- reaction_annotation.json
|   |   `-- traj/
|   |       |-- traj_default.csv
|   |       |-- traj_perturbed.csv
|   |       `-- trajectory_metadata.json
|   `-- ...
|-- challenge/               (66 systems)
`-- stress/                  (37 systems)
\end{verbatim}

Each system includes its source SBML file~\citep{hucka2003systems, xu2023sbmlkinetics}.
\texttt{equation.json} provides the standardized ODEs, and \texttt{reaction\_annotation.json} stores their reaction-level reference annotations.
Appendix~\ref{app:reaction_matching} describes annotation construction, kinetic components, and signature matching.
\texttt{model.py} provides an executable Python implementation of the reference ODEs.
The \texttt{traj/} directory holds the default and perturbed trajectories used for ID reconstruction and OOD extrapolation.
The expanded branch shows BIOMD0000000006~\citep{tyson1991modeling}, a two-state system released with trajectories sampled at 512 time points.

Stress annotations include complex rational rate laws and other kinetic forms outside the current reaction-matching protocol.
These annotations remain available even though Stress systems do not enter reaction-metric aggregates.

\subsection{ID Reconstruction, OOD Extrapolation, and Observation Noise}
\label{app:id_ood_noise}

To separate what varies between fitting and evaluation, we fix the fitting trajectory, the evaluation trajectory, and the corruption applied in the noise study.

\paragraph{Trajectory generation.}
Each system \(m\) uses a system-specific time horizon \([0,T_m]\) with 512 uniformly spaced observations. The default and perturbed trajectories share the same time grid and differ only in their initial conditions. System-specific horizons are provided with the released benchmark, together with a standalone reference simulator using LSODA~\citep{petzold1983automatic}.

\paragraph{ID reconstruction.}
The ID condition evaluates the recovered system against the clean default trajectory of the same system.
In the noise-free setting this is the trajectory used for fitting.
In the noise study, the fitting trajectory is corrupted first, and evaluation remains on the clean default trajectory.

\paragraph{OOD extrapolation.}
The OOD condition shifts the initial condition while retaining the same vector field, parameter setting, and evaluation time span.
The recovered system is evaluated against the trajectory generated from the perturbed initial condition.
The two conditions therefore differ only in the evaluation initial condition.

\paragraph{Observation noise.}
To model imperfect measurement, we corrupt only the fitting trajectory with state-scaled Gaussian noise at relative levels \(\sigma_{\mathrm{rel}}\in\{0.001,0.01,0.05,0.1\}\) (Equation~\ref{eq:observation_noise}), where the per-state scaling uses the temporal standard deviation of that state.
We leave the initial observation unchanged so that noise does not alter the fitting initial condition.
After perturbation, we set negative concentrations to zero.
We use noise seed 2026 and score all recovered systems against the clean default and clean perturbed trajectories.
We label the clean setting \textbf{Noise-free} and the corrupted settings by their \textbf{relative noise level} \(\sigma_{\mathrm{rel}}\).

\rev{The original results use one run per noise level. Additional runs with different noise seeds appear in Appendix~\ref{app:noisy_multirun_results}.}

\subsection{Metrics, System-Level Averaging, and Failures}
\label{app:metric_contract}

To define the reported aggregates, we first compute each metric at the system level.

\paragraph{System-level averaging.}
We compute each metric separately for each system and report the unweighted mean across the fixed target set of the corresponding evaluation (198 systems for trajectory metrics, 161 for reaction metrics).

\paragraph{Trajectory metrics.}
For system \(m\), let \(x_{m,j}(t_n)\) and \(\widehat{x}_{m,j}(t_n)\) denote the reference and recovered values of state \(j\) at the \(N_m\) evaluation times.
We compute the state-wise scores and their equal-weight system average as
\begin{equation}
  R_{m,j}^2
  = 1-
  \frac{\sum_{n=1}^{N_m}
    \left[x_{m,j}(t_n)-\widehat{x}_{m,j}(t_n)\right]^2}
  {\sum_{n=1}^{N_m}
    \left[x_{m,j}(t_n)-\bar{x}_{m,j}\right]^2},
  \qquad
  R_m^2=\frac{1}{d_m}\sum_{j=1}^{d_m}R_{m,j}^2,
  \label{eq:statewise_system_r2}
\end{equation}
where \(d_m\) is the number of states and \(\bar{x}_{m,j}=N_m^{-1}\sum_{n=1}^{N_m}x_{m,j}(t_n)\).
The state-wise fraction applies to nonconstant reference states.

For a constant reference state, an exactly matching prediction receives \(R^2=1\); otherwise, we mark the state-wise \(R^2\) as undefined.
We assess trajectory fidelity with the coverage-aware \(R^2\)-score of Equation~\ref{eq:trajectory_r2_score}: negative per-system \(R^2\) values and failed or non-finite simulations contribute zero, and \(\mathrm{Acc}@.90\) is the fraction of target systems with \(R^2\geq0.90\).
We report expression complexity as a cross-method evaluation metric.
Section~\ref{subsec:fpms} defines the parsimony criteria used for FPMS model selection.
Recovered equations that cannot be parsed are reported separately from the complexity metric.

\paragraph{Reaction metrics.}
We operationalize mechanism recovery through exact reaction-signature matching (Section~\ref{app:reaction_matching}).
The three reaction-eligible methods listed in Section~\ref{sec:evaluation} contribute reaction metrics.
For system \(m\), let \(\mathrm{TP}_m\) count matched signatures, \(\mathrm{FP}_m\) unmatched predicted signatures, and \(\mathrm{FN}_m\) unmatched reference signatures.
Per-system precision, recall, and F1 follow as
\begin{equation}
P_m
=
\frac{\mathrm{TP}_m}{\mathrm{TP}_m+\mathrm{FP}_m},
\qquad
R_m
=
\frac{\mathrm{TP}_m}{\mathrm{TP}_m+\mathrm{FN}_m},
\qquad
F_{1,m}
=
\frac{2P_mR_m}{P_m+R_m},
\label{eq:reaction_p_r_f1}
\end{equation}
and we report their unweighted means over the \(M=161\) reaction-scored systems:
\begin{equation}
\bar P
=
\frac{1}{M}\sum_{m=1}^{M}P_m,
\qquad
\bar R
=
\frac{1}{M}\sum_{m=1}^{M}R_m,
\qquad
\bar F_1
=
\frac{1}{M}\sum_{m=1}^{M}F_{1,m}.
\label{eq:reaction_system_means}
\end{equation}
An eligible method with an empty predicted set scores \(P_m=R_m=F_{1,m}=0\) (Section~\ref{app:reaction_matching}).

\paragraph{Kinetic-family metrics.}
Each predicted and reference record receives one diagnostic family label.
A record containing a Hill component is labeled Hill.
Otherwise a record containing a Michaelis--Menten component is labeled Michaelis--Menten.
All remaining records are labeled mass action.
True positives, false positives, and false negatives are pooled per label across the 161 systems, and per-label precision, recall, and F1 follow from the pooled counts.
Family metrics therefore use label-level pooling rather than the system-level averaging of the reaction metrics above.
This unit of counting differs from the reference-equation terms of Figure~\ref{fig:overview_sysbiosr}(b).

\subsection{Reaction Annotations and Matching}
\label{app:reaction_matching}

To compare recovered and reference reactions, we canonicalize both sides into term-projected, unit-stoichiometric reaction signatures and match them exactly.
Exact matching means equality over the signature fields defined below.
This subsection defines the construction and matching rules summarized in Section~\ref{sec:evaluation}.

\paragraph{Annotation construction.}
We construct reference annotations from the standardized ODE expressions.
For each state equation, we decompose the RHS into nonzero additive terms and separate each term's sign, coefficient, and kinetic factor.
A negative term defines a consuming record for the target species, whereas a positive term defines a producing record.
We identify the species-dependent factors and assign their kinetic-law classes and associated parameters.
Each equation-level record retains its originating equation index.
The annotations also provide a reaction-level representation that consolidates equation-level contributions with shared kinetics, as illustrated by the nonlinear example below.

\paragraph{Annotation representation.}
Each reaction is stored as a structured record containing its stoichiometry,
rate coefficient, kinetic factor, and species-dependent kinetic components.
For example, \(x_0 \rightarrow \varnothing\) with
\(v = 0.25\,x_0x_2/(0.02+x_0)\)
is represented by the following abridged record:
\begin{verbatim}
{
  "reactants": [1, 0, 0, 0, 0],
  "products":  [0, 0, 0, 0, 0],
  "coefficient": 0.25,
  "kinetic_factor": "x_0*x_2/(x_0 + 0.02)",
  "substrates": [
    {"species_idx": 0, "law": "mm", "order": 1, "K": 0.02}
  ],
  "modifiers": [
    {"species_idx": 2, "law": "linear", "order": 1}
  ]
}
\end{verbatim}

Here, \texttt{substrates} denote reactant-dependent kinetic components,
whereas \texttt{modifiers} denote regulatory dependencies.

For the 161 reaction-scored systems, the component vocabulary includes mass action (\(x^a\)), monomial dependence (\(x^q\)), Michaelis--Menten saturation (\(x/(K+x)\)), and Hill-type activation and inhibition.
In the examples, monomial dependence is written as \(\mathrm{Mono}(q)\), corresponding to the label \texttt{linear}.

\paragraph{Annotation checks.}
We reconstruct the RHS expressions from the annotated contributions and compare them with the reference ODEs using symbolic and numerical checks.
We also verify that the annotated kinetic components reproduce the corresponding kinetic factors.
All annotations are subsequently manually reviewed for consistency with the reference ODEs and kinetic forms.

\paragraph{A projection example.}
Consider the single reaction \(2A+B\rightarrow C\) with mass-action rate \(v=k\,x_A^2 x_B\).
The reaction expands into the coupled ODE terms
\[
2A+B\xrightarrow{v} C,
\qquad
v=k\,x_A^2 x_B,
\qquad
\begin{bmatrix}\dot{x}_A\\ \dot{x}_B\\ \dot{x}_C\end{bmatrix}
=
\begin{bmatrix}-2\\ -1\\ 1\end{bmatrix}
k\,x_A^2 x_B.
\]
All three terms share the same rate \(v\) and originate from this single reaction.
Term projection produces one signature per affected state equation.
The arrows below encode unit consumption or production of the target species; the remaining columns record the corresponding kinetic components.
\(\mathrm{MA}(q)\) denotes a mass-action reactant-dependent term of order \(q\).
\(\mathrm{Mono}(q)\) denotes a monomial kinetic dependency \(x^q\).
Kinetic roles are assigned relative to each projected signature.

\begin{center}
\setlength{\tabcolsep}{4pt}
\begin{tabular}{@{}>{\raggedright\arraybackslash}p{0.22\linewidth}>{\raggedright\arraybackslash}p{0.16\linewidth}>{\raggedright\arraybackslash}p{0.20\linewidth}>{\raggedright\arraybackslash}p{0.34\linewidth}@{}}
\toprule
Equation term & Unit-projected stoichiometry & Reactant-dependent term & Regulatory term(s) \\
\midrule
\(-2k x_A^2 x_B\) in \(\dot{x}_A\) & \(A\rightarrow\varnothing\) & A: MA(2) & B: Mono(1) \\
\(-k x_A^2 x_B\) in \(\dot{x}_B\) & \(B\rightarrow\varnothing\) & B: MA(1) & A: Mono(2) \\
\(+k x_A^2 x_B\) in \(\dot{x}_C\) & \(\varnothing\rightarrow C\) & none & A: Mono(2), B: Mono(1) \\
\bottomrule
\end{tabular}
\end{center}

The example shows three properties of the protocol.
First, the stoichiometric coefficient 2 on \(A\) yields exactly one consuming signature for \(\dot{x}_A\).
Unit projection normalizes stoichiometric multiplicities to one, while the order 2 remains part of the mass-action component.
Second, the kinetic roles depend on the target equation.
In each consuming signature, dependence on the consumed species is reactant-dependent and the remaining dependence is regulatory.
The producing signature represents both dependencies as regulatory terms, although \(A\) and \(B\) are reactants of the original reaction.
The \(\mathrm{Mono}(2)\) entries preserve the kinetic order after reassignment.
Third, a recovered term \(-k'x_A^2 x_B\) in \(\dot{x}_A\) matches the first signature even when its coefficient differs from \(k\).
A recovered term \(-k'x_A\) does not match.

\newpage
\paragraph{A nonlinear kinetic example.}
The released annotation of BIOMD0000000004~\citep{goldbeter1991minimal}, a five-state model over C, M, X, MI, and XI, contains 11 equation-level records, which are consolidated into seven reaction-level records.
The table shows the reaction-level representation.
\(\mathrm{MM}(K)\) denotes a Michaelis--Menten reactant-dependent term with affinity \(K\), and \(\mathrm{Hill}^+(K,n)\) a Hill-activation regulatory term.

\begin{center}
\setlength{\tabcolsep}{4pt}
\begin{tabular}{@{}>{\raggedright\arraybackslash}p{0.14\linewidth}>{\raggedright\arraybackslash}p{0.22\linewidth}>{\raggedright\arraybackslash}p{0.28\linewidth}>{\raggedright\arraybackslash}p{0.28\linewidth}@{}}
\toprule
Reaction & Reactant-dependent term(s) & Regulatory term(s) & Kinetic form \\
\midrule
\(\mathrm C\rightarrow\varnothing\) & C: MA(1) & none & \(v=0.01\,x_{\mathrm C}\) \\
\(\varnothing\rightarrow\mathrm C\) & none & none & \(v=0.025\) \\
\(\mathrm C\rightarrow\varnothing\) & C: MM(0.02) & X: Mono(1) & \(v=0.25\,\tfrac{x_{\mathrm C}}{0.02+x_{\mathrm C}}\,x_{\mathrm X}\) \\
\(\mathrm X\rightarrow\mathrm{XI}\) & X: MM(0.005) & none & \(v=0.5\,\tfrac{x_{\mathrm X}}{0.005+x_{\mathrm X}}\) \\
\(\mathrm{XI}\rightarrow\mathrm X\) & XI: MM(0.005) & M: Mono(1) & \(v=1.0\,\tfrac{x_{\mathrm{XI}}}{0.005+x_{\mathrm{XI}}}\,x_{\mathrm M}\) \\
\(\mathrm M\rightarrow\mathrm{MI}\) & M: MM(0.005) & none & \(v=1.5\,\tfrac{x_{\mathrm M}}{0.005+x_{\mathrm M}}\) \\
\(\mathrm{MI}\rightarrow\mathrm M\) & MI: MM(0.005) & C: \(\mathrm{Hill}^+\)(0.5, 1) & \(v=3.0\,\tfrac{x_{\mathrm{MI}}}{0.005+x_{\mathrm{MI}}}\,\tfrac{x_{\mathrm C}}{0.5+x_{\mathrm C}}\) \\
\bottomrule
\end{tabular}
\end{center}

The 11 equation-level records and 7 reaction-level records yield the same signature multiset after unit projection.
The final reaction-level record consolidates two equation-level records with the same rate coefficient and kinetic factor.
The consuming record \(\mathrm{MI}\rightarrow\varnothing\) retains the Michaelis--Menten dependence on \(\mathrm{MI}\) as a reactant-dependent term and the Hill-activation dependence on \(\mathrm C\) as a regulatory term.
The producing record \(\varnothing\rightarrow\mathrm M\) represents both dependencies as regulatory terms and represents the Michaelis--Menten factor as Hill activation at \(n=1\).
At \(n=1\), Michaelis--Menten saturation and Hill activation share the form \(x/(K+x)\).

\paragraph{Reference reactions.}
For each of the 161 reaction-scored systems, the annotation records the reference mechanism as a list of reactions.
Each reaction carries reactants and products as per-species integer counts, a rate coefficient, and a list of kinetic components.
Each kinetic-term entry records its species index, kinetic law, order, and, where applicable, an affinity constant \(K\) and a Hill exponent \(n\).
The component law vocabulary is: mass action (\(x^a\)), monomial dependence (\(x^q\)), Michaelis--Menten saturation (\(x/(K+x)\)), Hill activation (\(x^n/(K^n+x^n)\)), Hill inhibition (\(K^n/(K^n+x^n)\)), and Hill saturation (a Hill factor in a reactant-dependent role).

\paragraph{Signature construction.}
We canonicalize each reaction into one equation-level signature for every state with a nonzero net stoichiometric coefficient.
Let \( \mathbf{s}_r = \mathbf{s}_r^{+} - \mathbf{s}_r^{-} \) denote the net stoichiometric vector of reaction \(r\).
For each state \(x_j\) with \(s_{r,j} \neq 0\), we project \(s_{r,j}\) to its sign, retaining whether the reaction consumes or produces \(x_j\) while discarding the magnitude of the stoichiometric coefficient.
Kinetic dependence on a consumed target species is represented by a reactant-dependent term.
All remaining kinetic dependencies are represented as regulatory terms.
For a production signature, all kinetic dependencies are regulatory.
We canonicalize regulatory forms while preserving kinetic orders: Michaelis--Menten saturation becomes Hill activation with \(n=1\), and Hill saturation reassigned from a reactant-dependent role becomes Hill activation.

\paragraph{Matching.}
Each signature contains unit reactant and product vectors and two sorted kinetic-term lists, one reactant-dependent and one regulatory.
Each entry records its species index, kinetic law, and law-dependent details.
Two signatures match exactly when all four parts agree.
We discard predicted terms with fitted coefficient magnitude below \(10^{-12}\) before signature construction.
We use reference annotations as recorded.
We pool signatures across source reactions and compute multiset agreement as \(\mathrm{TP}=\sum_{\sigma}\min(\mathrm{true}_\sigma,\mathrm{pred}_\sigma)\).
Unmatched predicted and reference signatures count as false positives and false negatives.
We compute and average per-system precision, recall, and F1 as defined in Section~\ref{app:metric_contract}.

We compare every component's kinetic-law class but exclude numerical rate coefficients, affinity constants \(K\), and Hill exponents \(n\).
The score measures agreement between projected signatures.
If an eligible method returns no model, or we cannot map its output to the reaction representation, we assign it an empty predicted multiset.
Because all 161 reference multisets are nonempty, such cases receive zero precision, recall, and F1.

\newpage
\subsection{Composition of Benchmarks Compared in Related Work}
\label{app:compared_benchmark_composition}

Table~\ref{tab:benchmark_comparison} compares SysBio-SR with existing symbolic regression benchmarks.
This subsection documents how the counts reported for each external benchmark are composed.
Only instances posed as symbolic-regression tasks with ground-truth equation targets are counted.

\paragraph{AI Feynman.}
The AI Feynman database~\citep{udrescu2020ai} contains 120 equations: 100 regression mysteries drawn from the Feynman Lectures on Physics plus a bonus set of 20 additional equations.

\paragraph{SRBench.}
The SRBench ground-truth suite~\citep{la2021contemporary} comprises 130 datasets with known model forms, drawn from the Feynman Symbolic Regression Database and the ODE-Strogatz repository.
Its Feynman portion contains 116 scalar equations.
The underlying database contains 120 equations, but one is absent from the suite and three are excluded for containing arcsin or arccos terms.
Its Strogatz portion poses one dataset per state variable of seven two-state ODE systems, giving 14 single-state datasets in total.
Counting full systems rather than per-state datasets yields 123 systems (116 scalar and seven two-state) with 130 right-hand-side equations, as reported in Table~\ref{tab:benchmark_comparison}.
The remaining 122 black-box regression problems of SRBench have no ground-truth equation target and are excluded from the comparison.

\paragraph{EmpiricalBench.}
EmpiricalBench~\citep{cranmer2023interpretable} contains nine historical empirical relations from the physical sciences, including Hubble's law, Kepler's third law, and the ideal gas law.
Each instance poses one scalar target equation, so the reported system and equation counts both equal nine.

\paragraph{ODEBench.}
ODEBench~\citep{d2023odeformer} contains 63 ODE systems spanning state dimensions 1--4.
The reported 117 right-hand-side equations are the implied sum over the dimension distribution of the suite.

\paragraph{LLM-SRBench.}
The LLM-SRBench paper~\citep{shojaee2025llm} reports 239 equation-discovery tasks, each with one scalar target.
The set comprises 111 LSR-Transform tasks that reformulate established physical models and 128 LSR-Synth tasks spanning chemistry, biology, physics, and materials science, so the reported system and equation counts both equal 239.

\paragraph{SymbolBench.}
SymbolBench~\citep{liu2025can} evaluates symbolic regression over coupled ODE systems.
Its regression task extends ODEBench's 63 systems with additional systems, yielding 156 instances that span state dimensions 1--4.
The reported 383 right-hand-side equations are the implied sum over this dimension distribution.
Its Boolean-network and causal-discovery tasks are not equation-recovery tasks and are excluded from Table~\ref{tab:benchmark_comparison}.

\paragraph{MDBench.}
MDBench~\citep{bideh2026mdbench} combines 63 ODE systems (dimensions 1--4, contributing 117 right-hand-side equations) with 14 PDE systems (up to six state variables, contributing 24 right-hand-side equations).
All 77 systems carry ground-truth dynamics, giving 141 equations in total.

%% file: supplementary/method_details.tex
\clearpage
\section{KineSelect: Algorithmic Details}
\label{app:kineselect_details}

RSS proposes reaction structures, and KFI selects one kinetic realization per reaction during candidate evaluation.
FPMS selects the final model when the search produces a finite simulation-based archive.
The selected model is an executable reaction mechanism with fitted rate constants.
Ground-truth reaction annotations and reaction-matching scores remain evaluation-only.

\subsection{Reaction Structure Search}
\label{app:reaction_search}

KineSelect represents each proposed reaction by its reactant stoichiometry, product stoichiometry, and an optional regulator role.
For reaction \(r\), we write
\begin{equation}
g_r=\left(\mathbf{s}_r^-,\mathbf{s}_r^+,m_r\right),
\qquad
\mathbf{s}_r^-,\mathbf{s}_r^+\in\mathbb{Z}_{\geq0}^{d},
\label{eq:app_reaction_gene}
\end{equation}
where \(\mathbf{s}_r^-\) and \(\mathbf{s}_r^+\) encode reactants and products, and \(m_r\in\{\varnothing,0,\ldots,d-1\}\) identifies an optional regulator.
We restrict each reaction gene to at most one proposed regulator to avoid combinatorial enumeration of regulator subsets during structural and kinetic search.
When \(m_r\neq\varnothing\), we require \([\mathbf{s}_r^-]_{m_r}=[\mathbf{s}_r^+]_{m_r}=0\), so the regulator can affect the rate without entering the reactant or product stoichiometry.
The net stoichiometric vector is \(\mathbf{s}_r=\mathbf{s}_r^+-\mathbf{s}_r^-\).
For example, \(x_0\rightarrow x_1\) with rate \(v(\mathbf{x})=kx_0x_2\) assigns \(x_2\) as a regulator.
Its gene uses \(\mathbf{s}_r^-= (1,0,0)^\top\), \(\mathbf{s}_r^+=(0,1,0)^\top\), and \(m_r=2\).

RSS follows SISR's reaction-set evolutionary framework~\citep{palma2026stoichiometrically} and extends each reaction gene with the optional regulator role defined above.
It maintains separate subpopulations for models with \(\lvert\mathcal{R}(\mathcal{M})\rvert=c\), where \(c\in\{1,\ldots,c_{\max}\}\).
Each individual in subpopulation \(c\) contains exactly \(c\) reaction genes.
Crossover pools the genes of two parents, removes duplicates, and randomly draws \(c\) genes to form an offspring.
Mutation replaces one randomly selected gene with a gene from the pre-enumerated admissible reaction library.
For example, with \(c=3\), parent mechanisms
\[
\mathcal{R}(\mathcal{M}_1)=\{g_1,g_2,g_3\},
\qquad
\mathcal{R}(\mathcal{M}_2)=\{g_2,g_4,g_5\}
\]
form the deduplicated pool \(\{g_1,g_2,g_3,g_4,g_5\}\), from which one possible offspring is \(\{g_1,g_4,g_5\}\).
A mutation may then replace \(g_4\) with another admissible gene \(g_6\).
KFI assigns one kinetic realization to each reaction in every proposal and returns the derivative loss used to rank that proposal within its subpopulation.
After pre-screening candidates by derivative loss, each reaction-count subpopulation retains the locally non-dominated candidates under three archive objectives: simulation loss, expanded ODE expression-tree complexity, and active-reaction count.
The search merges these local sets, deduplicates candidates by reaction structure, and retains the globally non-dominated candidates under these archive objectives as the fixed archive \(\mathcal{A}\).
FPMS then re-scores \(\mathcal{A}\) using the four fidelity and parsimony criteria defined in Section~\ref{app:fpms_contract} without modifying its candidates.
If the search cannot form a finite simulation-based archive, it uses a predefined derivative-based fallback to return a model.

\subsection{Kinetic Library and Kinetic Form Identification}
\label{app:kinetic_library}

KFI separates kinetic-candidate selection from overall rate-constant estimation.
For each proposed reaction, it constructs unit-rate kinetic candidates by pairing a reactant-dependent kinetic term with a regulatory term:
\begin{equation}
\bar v_{r,(a,b)}(\mathbf{x})
=
\phi_{r,a}\left(\mathbf{x};\mathbf{s}_r^-\right)
\psi_{r,b}\left(\mathbf{x};m_r\right).
\label{eq:app_unit_rate_basis}
\end{equation}
Here \((a,b)\) indexes an admissible term pair, and a single index \(\ell\) enumerates the resulting set \(\mathcal{V}_r=\{\bar v_{r,\ell}\}\).
For a fixed proposed mechanism \(\mathcal{M}\), directly encoding one kinetic candidate in every reaction gene would admit \(\prod_{r\in\mathcal{R}(\mathcal{M})}|\mathcal{V}_r|\) joint kinetic assignments.
KFI avoids their explicit evolutionary enumeration by representing the alternatives with \(\sum_{r\in\mathcal{R}(\mathcal{M})}|\mathcal{V}_r|\) dictionary columns, fitting them jointly, and selecting one candidate per reaction group.
\rev{
During discovery, we generate each species' \(K\) candidates from its fitting observations and use the shared set \(n\in\{1,2,4\}\).
For species \(x_q\), the set \(\mathcal{K}_q\) uses the 0.05 and 0.50 quantiles of its positive fitting observations.}
When no positive observations are available, we use the fallback grid recorded in the full configuration.
\rev{Each admissible component setting defines a distinct kinetic choice.
Appendix~\ref{app:kinetic_parameter_coverage} reports how these fixed rules relate to the reference parameter distribution.}
KFI then refits only the overall rate constant:
\begin{equation}
v_r(\mathbf{x})=k_r\bar v_{r,\ell_r^\star}(\mathbf{x}),
\qquad
\dot{\mathbf{x}}=
\sum_{r\in\mathcal{R}(\mathcal{M})}
\mathbf{s}_r k_r\bar v_{r,\ell_r^\star}(\mathbf{x}).
\label{eq:app_physical_rate_law}
\end{equation}

Table~\ref{tab:kinetic_library} lists the admissible reactant-dependent and regulatory terms used to construct these candidates, together with their availability conditions.

\begin{table}[!htp]
\centering
\setlength{\tabcolsep}{3pt}
\renewcommand{\arraystretch}{1.14}
\caption{\textbf{Kinetic terms used to construct KFI candidates.}
KFI pairs a reactant-dependent kinetic term with a regulatory term.
Here \(a_{r,q}=[\mathbf{s}_r^-]_q\), \(m=m_r\), and \(\mathcal{K}_q\) denotes the finite species-specific candidate set for \(x_q\).}
\label{tab:kinetic_library}
\begin{tabular}{@{}>{\raggedright\arraybackslash}p{0.13\textwidth}>{\raggedright\arraybackslash}p{0.20\textwidth}>{\raggedright\arraybackslash}p{0.42\textwidth}>{\raggedright\arraybackslash}p{0.20\textwidth}@{}}
\toprule
Term type & Law & Form & Availability \\
\midrule
Reactant-dependent & Mass action & \(\displaystyle \phi_r^{\mathrm{MA}}(\mathbf{x})=\prod_{q=0}^{d-1}x_q^{a_{r,q}}\) & Available for every reaction; it equals \(1\) for a source reaction. \\
\addlinespace
Reactant-dependent & \mbox{Michaelis--Menten} & \(\displaystyle \phi_{r,j,K}^{\mathrm{MM}}(\mathbf{x})=x_j^{a_{r,j}-1}\frac{x_j}{K+x_j}\prod_{q\neq j}x_q^{a_{r,q}}\) & Available when \(a_{r,j}>0\) and \(K\in\mathcal{K}_j\). Unavailable for source reactions. \\
\addlinespace
Regulatory & None & \(\displaystyle \psi^{\mathrm{none}}(\mathbf{x})=1\) & Available for every reaction. \\
\addlinespace
Regulatory & Linear & \(\displaystyle \psi^{\mathrm{lin}}(\mathbf{x};m)=x_m\) & Requires \(m\). \\
\addlinespace
Regulatory & Hill activation & \(\displaystyle \psi_{K,n}^{\mathrm{act}}(\mathbf{x};m)=\frac{x_m^n}{K^n+x_m^n}\) & Requires \(m\), \(K\in\mathcal{K}_m\), and \(n\in\{1,2,4\}\). \\
\addlinespace
Regulatory & Hill inhibition & \(\displaystyle \psi_{K,n}^{\mathrm{inh}}(\mathbf{x};m)=\frac{K^n}{K^n+x_m^n}\) & Requires \(m\), \(K\in\mathcal{K}_m\), and \(n\in\{1,2,4\}\). \\
\bottomrule
\end{tabular}
\end{table}

KFI builds the scale-normalized, stoichiometrically coupled dictionary in Eq.~\eqref{eq:kfi_column} and solves the joint regularized fit in Eq.~\eqref{eq:kfi_joint_fit}.
This system-level fit lets kinetic alternatives compete in the context of the complete proposed mechanism.
A regulator can change the temporal amplitude through \(\bar v_{r,\ell}(\mathbf{X})\), while its zero stoichiometric entry prevents a direct derivative contribution to the regulator state.

Before normalization, we replace near-zero derivative scales with a safe value and set near-zero dictionary columns to zero.
Before the joint fit, we prune nearly collinear candidates within each reaction group, retaining simpler representatives.
The full configuration records the cosine thresholds and numerical tolerances.

KFI computes each retained candidate's kinetic contribution score from its fitted weight \(\widehat w_{r,\ell}\) obtained in Eq.~\eqref{eq:kfi_joint_fit}.
For nondegenerate unit-normalized columns, non-negative fitted weights, and \(\|\mathbf{y}\|_2>0\),
\begin{equation}
\eta_{r,\ell}
=
\frac{\|\mathbf{a}_{r,\ell}\widehat w_{r,\ell}\|_2}{\|\mathbf{y}\|_2}
=
\frac{\widehat w_{r,\ell}}{\|\mathbf{y}\|_2},
\qquad
\operatorname*{arg\,max}_{\ell}\eta_{r,\ell}
=
\operatorname*{arg\,max}_{\ell}\widehat w_{r,\ell}.
\label{eq:app_rgks_score}
\end{equation}
Thus, \(\eta_{r,\ell}\) is a target-normalized fitted amplitude that yields the same maximizing-index set as \(\widehat w_{r,\ell}\).
KFI selects the highest-scoring kinetic candidate in each reaction group and refits the overall rate constants \(\widehat{k}_r\) using the selected kinetic forms.
Algorithm~\ref{alg:rgks} summarizes the procedure.

\newpage
\begin{algorithm}[!ht]
\caption{Kinetic Form Identification (KFI)}
\label{alg:rgks}
\begin{algorithmic}[1]
\Require Proposed reaction model \(\mathcal{M}\), fitting trajectory \(\mathbf{X}\), derivative estimate \(\dot{\mathbf{X}}\), kinetic-library configuration, and regularization coefficient \(\lambda\)
\Ensure Kinetic-resolved model \(\widehat{\mathcal{M}}\) and derivative loss \(\mathcal{L}^{\mathrm{der}}\)
\State Construct \(\Gamma\) and \(\mathbf{y}=\operatorname{vec}(\dot{\mathbf{X}}\Gamma^{-1})\)
\ForAll{\(r\in\mathcal{R}(\mathcal{M})\)}
    \State Enumerate the admissible unit-rate candidates \(\mathcal{V}_r\)
    \State Construct the normalized columns \(\{\mathbf{a}_{r,\ell}\}_{\ell}\)
\EndFor
\State Prune nearly collinear columns within each group and update \(\{\mathcal{V}_r\}_r\)
\State Assemble \(\boldsymbol{\Theta}(\mathcal{M})\) and solve the joint fit in Eq.~\eqref{eq:kfi_joint_fit}
\ForAll{\(r\in\mathcal{R}(\mathcal{M})\)}
    \State Compute \(\eta_{r,\ell}\) for all \(\ell\in\{1,\ldots,|\mathcal{V}_r|\}\)
    \State Select \(\ell_r^\star\) as the first maximizer in Eq.~\eqref{eq:app_rgks_score} under the fixed kinetic-library order
\EndFor
\State Jointly refit the selected unnormalized unit-rate columns to obtain \(\{\widehat{k}_r\}_{r\in\mathcal{R}(\mathcal{M})}\)
\State Form \(\widehat{\mathcal{M}}\) and compute \(\mathcal{L}^{\mathrm{der}}\)
\State \Return \(\widehat{\mathcal{M}}\) and \(\mathcal{L}^{\mathrm{der}}\)
\end{algorithmic}
\end{algorithm}

The fitted weights \(\widehat{\mathbf{w}}\) guide kinetic-candidate selection, while the refitted rate constants \(\{\widehat{k}_r\}_r\) define the executable ODE.

\subsection{Fidelity--Parsimony Model Selection}
\label{app:fpms_contract}

\paragraph{\texorpdfstring{\(\alpha\)-threshold selection}{Alpha-threshold selection}.}
\label{app:alpha_threshold}
We use \(\alpha\)-threshold selection as a deterministic baseline for archive-level model selection.
It operationalizes the loss--complexity principle used by SISR~\citep{palma2026stoichiometrically}.
This study defines the threshold rule for the controlled selector comparison.

For a fixed candidate archive \(\mathcal{A}\), let \(\mathcal{L}^{\mathrm{sim}}_{\min} =\min_{\mathcal{M}_i\in\mathcal{A}}\mathcal{L}^{\mathrm{sim}}_i\).
The rule retains
\begin{equation}
\mathcal{A}_{\alpha}
=
\left\{
\mathcal{M}_i\in\mathcal{A}:
\mathcal{L}^{\mathrm{sim}}_i
\leq
\alpha\mathcal{L}^{\mathrm{sim}}_{\min}
\right\}.
\label{eq:alpha_threshold}
\end{equation}
Here, \(\mathcal{L}^{\mathrm{sim}}_i\) compares the simulated fitting trajectory with the observed trajectory.
We set \(\alpha=1.05\), which allows up to a \(5\%\) increase above the minimum simulation loss.
We chose this default after inspecting SISR's reaction F1 and OOD \(R^2\) on this benchmark.
The implementation adds \(\epsilon=10^{-12}\) to the right-hand side for numerical comparison.
Within this set, the selector chooses the model with the lowest expanded ODE expression-tree complexity \(C_i^{\mathrm{tree}}\), which FPMS also uses as its tree-parsimony criterion.

\paragraph{FPMS.}

FPMS scores complete mechanisms in the shared archive \(\mathcal{A}\).
For each candidate \(\mathcal{M}_i\in\mathcal{A}\), it uses derivative loss \(\mathcal{L}_i^{\mathrm{der}}\) and simulation loss \(\mathcal{L}_i^{\mathrm{sim}}\).
Simulation loss integrates each fitted candidate from the first observed state over the fitting time grid and compares the solution with the same observed trajectory \(\mathbf{X}\).
Failed or non-finite simulations receive infinite loss.

A reaction is active when \(\lvert\widehat{k}_{i,r}\rvert>10^{-14}\).
Expanded ODE expression-tree complexity \(C_i^{\mathrm{tree}}\) follows the SISR-inspired node count after expanding each active reaction into every affected state equation~\citep{palma2026stoichiometrically}.
It includes sign, stoichiometric, rate-constant, kinetic-expression, and within-equation addition nodes.
Specialized kinetic-component count \(C_i^{\mathrm{spec}}\) separately counts Michaelis--Menten reactant-dependent terms and Hill regulatory terms.

For example, an active mass-action reaction \(x_0\rightarrow x_1\) with rate \(v(\mathbf{x})=kx_0\) expands to
\[
\dot{x}_0=-kx_0,
\qquad
\dot{x}_1=+kx_0.
\]
Each term contributes four nodes: one sign node, one unit-stoichiometry node, one rate-constant node, and one kinetic node.
Because each equation contains only one active term, the count includes no addition node, giving \(C_i^{\mathrm{tree}}=8\) and \(C_i^{\mathrm{spec}}=0\).
An active reaction combining a Michaelis--Menten reactant-dependent term with a Hill regulatory term contributes \(2\) to \(C_i^{\mathrm{spec}}\).

The criteria overlap because Michaelis--Menten and Hill components also contribute to \(C_i^{\mathrm{tree}}\).
The tree count covers the expanded state equations, whereas \(C_i^{\mathrm{spec}}\) counts each active specialized kinetic component once, irrespective of the number of affected state equations.

FPMS maps negative or non-finite losses to \(+\infty\) before applying the log transform:
\begin{equation}
\mathcal{L}_i^{\mathrm{der}}
\leftarrow
\log\left(\mathcal{L}_i^{\mathrm{der}}+\varepsilon\right),
\qquad
\mathcal{L}_i^{\mathrm{sim}}
\leftarrow
\log\left(\mathcal{L}_i^{\mathrm{sim}}+\varepsilon\right),
\qquad
\varepsilon=10^{-12}.
\label{eq:app_fpms_log_losses}
\end{equation}
FPMS independently min--max normalizes the transformed losses and the two complexity criteria over \(\mathcal{A}\).
If the finite values of a criterion have zero range, they map to zero and FPMS marks the criterion as degenerate.
Non-finite entries map to one whenever the criterion has a finite value.
If no finite value exists, FPMS also marks the criterion as degenerate and maps all entries to zero.

Tildes on \(\mathcal{L}_i^{\mathrm{der}}\) and \(\mathcal{L}_i^{\mathrm{sim}}\) denote the normalized values of the log-transformed losses.
FPMS forms a lack-of-fidelity score \(Q_i\) and a model-complexity score \(P_i\):
\begin{equation}
Q_i=
\frac{\widetilde{\mathcal{L}}_i^{\mathrm{der}}+
\widetilde{\mathcal{L}}_i^{\mathrm{sim}}}{2},
\qquad
P_i=
\frac{\widetilde{C}_i^{\mathrm{tree}}+
\widetilde{C}_i^{\mathrm{spec}}}{2},
\label{eq:app_fpms_criteria}
\end{equation}
where \(Q_i,P_i\in[0,1]\).
Lower \(Q_i\) and \(P_i\) indicate higher fidelity and lower model complexity, respectively.

FPMS selects the final model by maximizing the multiplicative utility:
\begin{equation}
U_i=(1-Q_i)(1-P_i),
\qquad
\mathcal{M}^\star=
\operatorname*{arg\,max}_{\mathcal{M}_i\in\mathcal{A}}U_i.
\label{eq:app_fpms_utility}
\end{equation}
FPMS resolves ties in \(U_i\) by minimizing \(Q_i\) and then selecting the lexicographically smallest canonical candidate identifier.

%% file: supplementary/experiment_details.tex
\clearpage
\section{Supplementary Experimental Results}
\label{app:supp_experimental_results}

This appendix extends the joint trajectory and mechanism-recovery analysis in Section~\ref{sec:experiments}.
We examine mechanism recovery at high OOD trajectory fidelity, kinetic-family recovery, final-model selection, and OOD performance distributions under observation noise.
We then analyze representative recovery and failure cases.

\subsection{Mechanism Recovery among High-Fidelity Predictions}
\label{app:high_fidelity_mechanism}

OOD \(R^2\geq0.90\) does not guarantee high reaction F1 (Table~\ref{tab:supp_high_fidelity_recovery}).
Among the 65 KineSelect predictions with OOD \(R^2\geq0.90\), 18 systems (27.7\%) have reaction F1 below 0.50.
For SISR and Reactive SINDy predictions with OOD \(R^2\geq0.90\), the corresponding counts are 17 of 56 systems (30.4\%) and 7 of 16 systems (43.8\%), respectively.

KineSelect reaches OOD \(R^2\geq0.90\) on more reaction-eligible systems than SISR: 65 of 161 (40.4\%) versus 56 of 161 (34.8\%).
Among predictions with OOD \(R^2\geq0.90\), mean reaction F1 is similar for KineSelect and SISR, at 0.658 and 0.653, respectively; their median F1 values are 0.714 and 0.727.

\input{Table/supp_high_fidelity_recovery}

\subsection{Kinetic-Family Recovery and Reference Support}
\label{app:kinetic_family_recovery}

Mass-action records dominate the kinetic-family evaluation (Table~\ref{tab:supp_kinetic_family_recovery}).
Mass action appears in 158 systems with 1,309 reference records, compared with 32 systems and 47 records for Michaelis--Menten kinetics.
Hill kinetics account for 163 records across 61 systems.

KineSelect has the highest mass-action precision, recall, and F1, but its nonlinear-family scores expose different recovery difficulties.
For Michaelis--Menten kinetics, precision is 0.034 while recall is 0.213, indicating many unmatched predictions in addition to missed reference signatures.
Hill recovery has low precision and recall, at 0.021 and 0.031.
In the evaluated configurations, Reactive SINDy and SISR lack these nonlinear components (Appendix~\ref{app:method_configurations}).
KineSelect represents both component types, yet their F1 scores remain 0.059 and 0.025, compared with 0.503 for mass action.
Improving nonlinear kinetic recovery requires addressing the low precision of Michaelis--Menten predictions and the low precision and recall of Hill predictions.

\input{Table/supp_kinetic_family_recovery}

\input{supplementary/kinetic_parameter_coverage}

\subsection{Sensitivity and Ablations of Final-Model Selection}
\label{app:selection_sensitivity}

\begin{figure}[htbp]
  \centering
  \includegraphics[width=\linewidth]{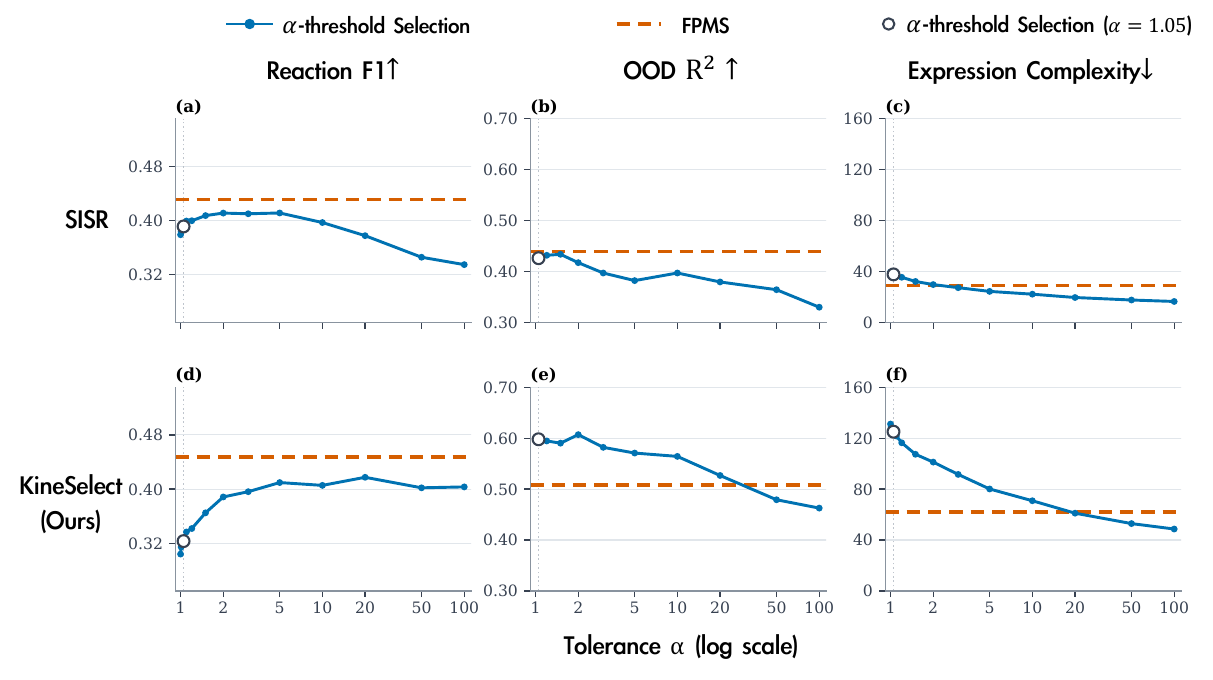}
  \caption{\textbf{Sensitivity of final-model selection to \(\alpha\).}
  The top and bottom rows use candidates from SISR and the KineSelect search, respectively.
  Columns report reaction F1 (a,d), OOD \(R^2\) (b,e), and benchmark expression complexity (c,f).
  Solid curves show 13 \(\alpha\) settings from 1 to 100 on a log axis, dashed lines show FPMS, and outlined circles mark the default \(\alpha=1.05\).
  Trajectory and complexity means use 198 systems, and reaction F1 uses 161.}
  \label{fig:supp_alpha_sensitivity}
\end{figure}

\paragraph{Sensitivity to $\alpha$.}
The preferred threshold depends on both the candidate archive and the evaluation metric (Figure~\ref{fig:supp_alpha_sensitivity}).
We compare \(\alpha\)-threshold selection and FPMS on a fixed candidate archive from each search method.
Appendix~\ref{app:alpha_threshold} defines the threshold rule and the default \(\alpha=1.05\).
For SISR, F1 rises to about 0.411 at \(\alpha\approx2\text{--}5\), then generally declines as the selected expressions become simpler.
OOD \(R^2\) peaks at 0.434 with \(\alpha=1.5\).
FPMS exceeds both tested maxima, reaching F1 of 0.431 and OOD \(R^2\) of 0.440.

KineSelect candidates gain more reaction F1 from relaxing \(\alpha\), with the highest tested F1 occurring at \(\alpha=20\).
From the default \(\alpha=1.05\) to \(\alpha=20\), F1 rises from 0.323 to 0.417, while OOD \(R^2\) decreases from 0.598 to 0.527.
Mean expression complexity falls from 125.16 to 61.10, and mean specialized kinetic-component count falls from 5.64 to 2.45.
Larger thresholds continue to simplify the models but do not improve on the F1 at \(\alpha=20\).
Thus, relaxing the fitting-loss threshold favors simpler models, but the thresholds that maximize reaction F1 differ across candidate archives and from those that maximize OOD \(R^2\).

For KineSelect candidates, FPMS and \(\alpha=20\) yield mean expression complexities of 62.13 and 61.10, respectively, but F1 differs: 0.447 versus 0.417.
FPMS yields lower OOD \(R^2\), at 0.508 versus 0.527.
FPMS avoids choosing a simulation-loss tolerance while achieving higher reaction F1 than every tested \(\alpha\) setting on both candidate sets.

\newpage
\paragraph{FPMS component ablations.}
Full FPMS achieves the highest reaction F1 on both Main and Challenge (Table~\ref{tab:supp_fpms_ablation}).
Adding derivative loss or specialized kinetic-component count to Base raises Overall F1 from 0.427 to about 0.436; using both reaches 0.447.
The individual additions have different effects across subsets: derivative loss yields higher F1 on Main, whereas specialized kinetic-component count yields higher F1 on Challenge.

The similar Overall OOD \(R^2\) values mask opposite changes across Main and Challenge.
Relative to Base, Full FPMS lowers Main OOD \(R^2\) from 0.624 to 0.601 and raises Challenge OOD \(R^2\) from 0.404 to 0.437.
Thus, combining the criteria improves reaction F1 on both subsets, while its trajectory advantage is specific to Challenge.
\rev{Appendix~\ref{app:fpms_ablation_across_seeds} compares the same settings across search seeds.}

\input{Table/supp_fpms_ablation}

\subsection{OOD Performance Distributions under Observation Noise}
\label{app:ood_noise_distributions}

\begin{figure}[!htbp]
  \centering
  \includegraphics[width=\linewidth]{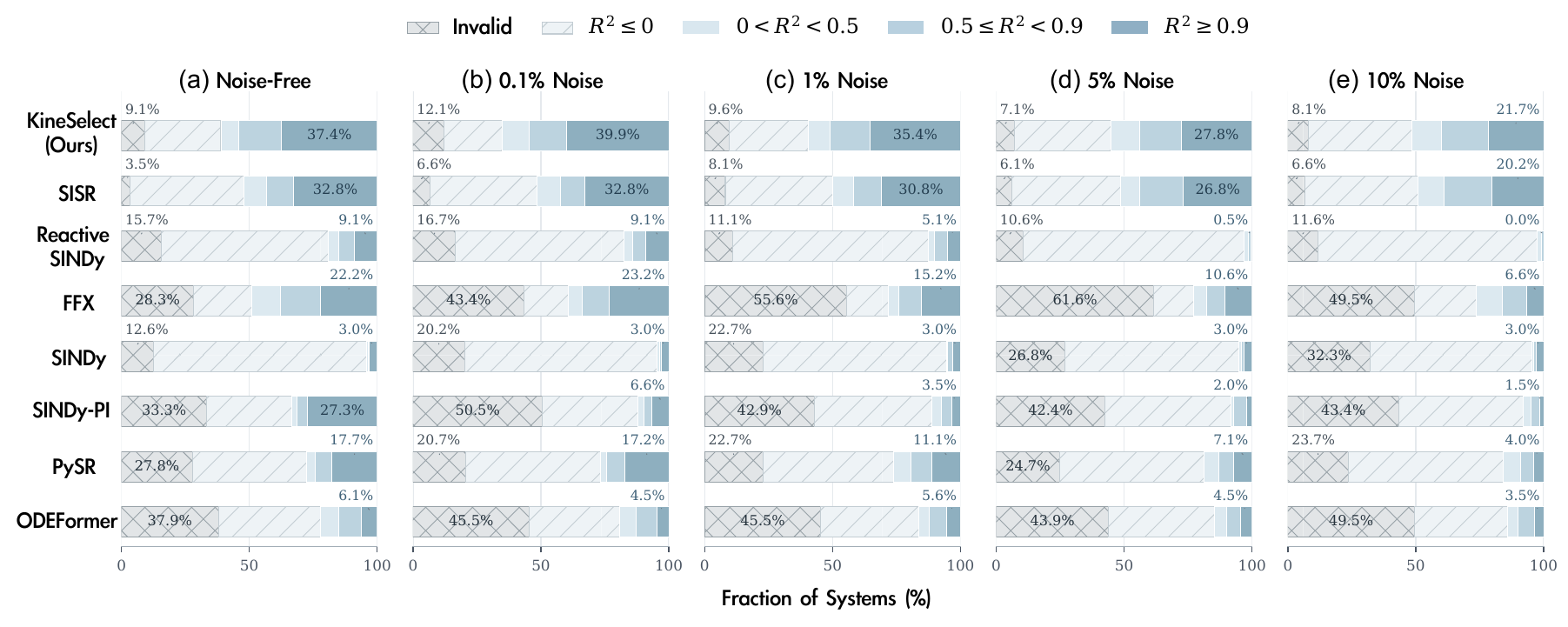}
  \caption{\textbf{OOD performance distributions under observation noise.}
  (a--e) Results for noise-free observations and relative noise levels of 0.1\%, 1\%, 5\%, and 10\%, respectively.
  Each bar represents all 198 systems.
  Cross-hatched segments denote invalid or undefined OOD scores, and diagonally hatched segments denote valid scores with raw OOD \(R^2\leq0\).
  Valid predictions occupy four bins based on raw OOD \(R^2\): \(R^2\leq0\), \(0<R^2<0.5\), \(0.5\leq R^2<0.90\), and \(R^2\geq0.90\).
  Labels give percentages of invalid/undefined scores and predictions with OOD \(R^2\geq0.90\) to one decimal place.}
  \label{fig:supp_ood_noise_distribution}
\end{figure}

KineSelect achieves the highest OOD \(R^2\) and \(\mathrm{Acc}@.90\) across all five tested noise conditions (Table~\ref{tab:supp_noisy_trajectory}).
Figure~\ref{fig:supp_ood_noise_distribution} shows how the OOD score distributions change with noise.
KineSelect reaches OOD \(R^2\geq0.90\) on 74 systems without noise and 43 at 10\% noise, compared with 65 and 40 for SISR.
The difference in the number of systems with OOD \(R^2\geq0.90\) therefore narrows at high noise.

\newpage
The trajectory advantage does not extend to Overall reaction F1 at the higher noise levels (Table~\ref{tab:supp_noisy_reaction}).
At 5\% noise, KineSelect has higher Challenge F1 than SISR (0.277 versus 0.232), but lower Main F1 (0.392 versus 0.432), yielding lower Overall F1.
At 10\% noise, SISR has higher F1 on both subsets, whereas KineSelect retains higher OOD \(R^2\).
The richer kinetic representation may admit alternative fits to noisy observations that preserve trajectory accuracy while matching the reference reactions less well.

The baseline distributions distinguish missing or invalid OOD scores from valid but inaccurate dynamics.
FFX, SINDy-PI, and ODEFormer have large fractions of systems without valid OOD scores, reaching 61.6\% for FFX at 5\% noise.
For Reactive SINDy, valid but non-positive OOD \(R^2\) dominates at high noise: inaccurate trajectories remain a problem even when a score is available.
SINDy reaches OOD \(R^2\geq0.90\) on only 6 of 198 systems (3.0\%) in each of the five conditions.

\begin{table}[!htbp]
  \input{Table/supp_noisy_trajectory}
  \par\bigskip
  \input{Table/supp_noisy_reaction}
\end{table}

\clearpage
\input{supplementary/noisy_multirun_results}

\clearpage
\subsection{Representative Recovery and Failure Cases}
\label{app:representative_cases}

\begin{figure}[htbp]
  \centering
  \includegraphics[width=\linewidth]{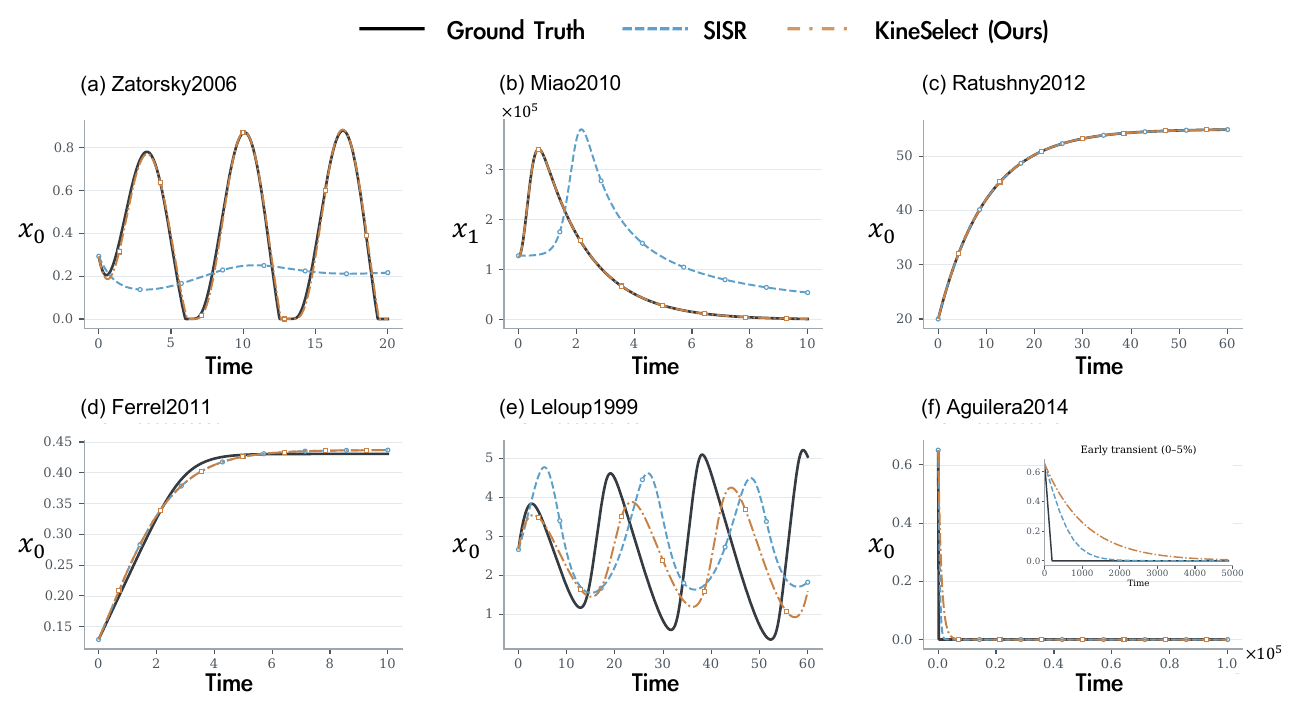}
  \caption{\textbf{Representative OOD trajectories after noise-free fitting.}
  (a-f) OOD trajectories of six representative systems, comparing Ground Truth, SISR, and KineSelect.
  Table~\ref{tab:supp_representative_cases} reports the corresponding equations and complete system-level OOD and reaction metrics.
  Each plot displays one state; the inset in (f) shows the first 5\% of the time interval.}
  \label{fig:supp_representative_trajectories}
\end{figure}

We use six representative systems to illustrate how trajectory fidelity and mechanism recovery align in some cases and separate in others.
Figure~\ref{fig:supp_representative_trajectories} shows OOD trajectories after noise-free fitting, and Table~\ref{tab:supp_representative_cases} compares the reference and recovered equations alongside the full-system metrics.

KineSelect recovers more reference signatures and produces more accurate OOD trajectories than SISR in Zatorsky2006 and Miao2010 (Figure~\ref{fig:supp_representative_trajectories}(a,b)).
In Zatorsky2006~\citep{geva2006oscillations}, KineSelect matches the Michaelis--Menten consumption signature in \(\dot{x}_0\) together with four mass-action signatures.
It closely reproduces the reference oscillations, reaching OOD \(R^2=0.995\) and reaction F1 of 0.714.
SISR matches none of the reference signatures and produces a nearly constant trajectory, with OOD \(R^2=-0.236\) and F1 of 0.
In Miao2010~\citep{miao2010quantifying}, both methods recover the \(x_0x_2\) interaction signatures, but KineSelect also matches the linear consumption and production signatures that SISR misses.
KineSelect closely follows the reference transient (OOD \(R^2=1.000\), F1 \(=0.909\)), whereas SISR extrapolates poorly (OOD \(R^2=-1.507\), F1 \(=0.333\)).
Even with this near-perfect OOD trajectory, KineSelect misses one reference linear-production signature.

For Ratushny2012~\citep{ratushny2012asymmetric}, both methods recover the two reference signatures for constant production and linear consumption, attaining OOD \(R^2=1.000\) and reaction F1 of 1.000 (Figure~\ref{fig:supp_representative_trajectories}(c)).
Ferrel2011~\citep{ferrell2011modeling} gives the contrasting case: both methods again reach OOD \(R^2\approx0.993\), but their reaction F1 is only 0.400 (Figure~\ref{fig:supp_representative_trajectories}(d)).
Both methods match the constant production signature while missing the reference Hill-regulated consumption.
Near-perfect OOD trajectories can therefore correspond to either complete or partial mechanism recovery.

Leloup1999~\citep{leloup1999limit} shows the reverse mismatch: KineSelect matches more reference signatures, yet both methods extrapolate poorly (Figure~\ref{fig:supp_representative_trajectories}(e)).
KineSelect matches inhibitory-Hill and Michaelis--Menten signatures and reaches reaction F1 of 0.588, compared with 0.375 for SISR.
Both models deviate from the reference oscillations, with OOD \(R^2=-0.407\) for KineSelect and \(-1.502\) for SISR.
The matched Hill term uses \(n=2\) instead of the reference \(n=4\).
Numerical Hill exponents are not part of the reaction-signature matching criterion (Appendix~\ref{app:reaction_matching}).

SISR performs better on both trajectory fidelity and mechanism recovery in Aguilera2014~\citep{aguilera2014studying} (Figure~\ref{fig:supp_representative_trajectories}(f)).
SISR matches three reference mass-action signatures, whereas the selected KineSelect model retains only one reference match and includes unmatched Michaelis--Menten and Hill terms.
SISR attains OOD \(R^2=0.495\) and reaction F1 of 0.545, compared with \(-0.152\) and 0.182 for KineSelect, respectively.

KineSelect recovers Michaelis--Menten and Hill signatures beyond SISR's mass-action representation in these examples.
Reliable recovery of nonlinear mechanisms together with accurate OOD dynamics remains an open challenge.
Across these cases, trajectory fidelity and mechanism recovery can improve together or diverge in either direction, motivating their joint evaluation.

\input{Table/supp_representative_cases}

\clearpage
\input{supplementary/clean_multirun_results}

\clearpage
\input{supplementary/multi_ood_results}

\clearpage
\input{supplementary/fpms_ablation_across_seeds}

%% file: Table/supp_high_fidelity_recovery.tex
\begin{table}[htbp]
  \centering
  \caption{\textbf{Mechanism recovery for predictions with OOD \(R^2\geq0.90\).}
  Coverage gives the number and percentage of the 161 reaction-eligible
  systems with OOD \(R^2\geq0.90\). For each method, we report mean and
  median reaction F1, together with the count and percentage of systems
  with F1 \(<0.50\), among systems with OOD \(R^2\geq0.90\).}
  \label{tab:supp_high_fidelity_recovery}
  \begingroup
  \footnotesize
  \setlength{\tabcolsep}{4pt}
  \renewcommand{\arraystretch}{1.15}
  \newcolumntype{Y}{>{\centering\arraybackslash}X}
  \begin{tabularx}{\linewidth}{@{}l*{4}{Y}@{}}
    \toprule
    \textbf{Method} & \textbf{Coverage} & \textbf{Mean F1} & \textbf{Median F1} & \textbf{F1 \(<0.50\)} \\
    \midrule
    Reactive SINDy & 16 (9.9\%) & 0.552 & 0.519 & 7 (43.8\%) \\
    SISR & 56 (34.8\%) & 0.653 & 0.727 & 17 (30.4\%) \\
    KineSelect & 65 (40.4\%) & 0.658 & 0.714 & 18 (27.7\%) \\
    \bottomrule
  \end{tabularx}
  \endgroup
\end{table}

%% file: Table/supp_kinetic_family_recovery.tex
\begin{table}[htbp]
  \centering
  \caption{\textbf{Kinetic-family recovery and reference support.}
  The \emph{Systems} column counts systems containing reference records of the family;
  the \emph{Records} column counts projected
  unit-stoichiometry reference records with multiplicity. System supports
  can overlap. Precision, recall, and F1 use TP, FP, and FN counts pooled
  by kinetic-family label across the 161 reaction-eligible systems
  (Appendix~\ref{app:metric_contract}). Bold and underlined values denote
  the best and second-best results within each family, including ties.}
  \label{tab:supp_kinetic_family_recovery}
  \begingroup
  \footnotesize
  \setlength{\tabcolsep}{3pt}
  \renewcommand{\arraystretch}{1.15}
  \newcolumntype{Y}{>{\centering\arraybackslash}X}
  \begin{tabularx}{\linewidth}{@{}>{\raggedright\arraybackslash}p{0.20\linewidth} YY >{\raggedright\arraybackslash}p{0.21\linewidth} YYY@{}}
    \toprule
    \textbf{Family} & \textbf{Systems} & \textbf{Records} & \textbf{Method} & \textbf{Precision}\(\uparrow\) & \textbf{Recall}\(\uparrow\) & \textbf{F1}\(\uparrow\) \\
    \midrule
    \multirow{3}{*}{Mass action} & \multirow{3}{*}{158} & \multirow{3}{*}{1309} & Reactive SINDy & 0.260 & 0.380 & 0.309 \\
     &  &  & SISR & \underline{0.379} & \underline{0.412} & \underline{0.395} \\
     &  &  & KineSelect & \textbf{0.552} & \textbf{0.461} & \textbf{0.503} \\
    \addlinespace
    \multirow{3}{*}{Michaelis--Menten} & \multirow{3}{*}{32} & \multirow{3}{*}{47} & Reactive SINDy & \underline{0.000} & \underline{0.000} & \underline{0.000} \\
     &  &  & SISR & \underline{0.000} & \underline{0.000} & \underline{0.000} \\
     &  &  & KineSelect & \textbf{0.034} & \textbf{0.213} & \textbf{0.059} \\
    \addlinespace
    \multirow{3}{*}{Hill} & \multirow{3}{*}{61} & \multirow{3}{*}{163} & Reactive SINDy & \underline{0.000} & \underline{0.000} & \underline{0.000} \\
     &  &  & SISR & \underline{0.000} & \underline{0.000} & \underline{0.000} \\
     &  &  & KineSelect & \textbf{0.021} & \textbf{0.031} & \textbf{0.025} \\
    \bottomrule
  \end{tabularx}
  \endgroup
\end{table}

%% file: supplementary/kinetic_parameter_coverage.tex
\begin{revision}
\subsubsection{Reference Kinetic Parameters and Candidate Coverage}
\label{app:kinetic_parameter_coverage}

\paragraph{Reference scope.}
We examine MM and Hill shape parameters in the standardized reference annotations for the 161 Main + Challenge systems.
We count each distinct kinetic factor once per system, distinguishing its species, kinetic law, order, \(K\), and \(n\).
This yields 189 factors with \(K\) across 66 systems and 138 Hill factors across 61 systems.

\paragraph{Hill exponents and reference scales.}
Figure~\ref{fig:supp_kinetic_parameter_coverage}(a) shows the distribution of reference Hill exponents \(n\).
Figure~\ref{fig:supp_kinetic_parameter_coverage}(b) shows the reference \(K\) values normalized by \(\operatorname{median}(X_j^+)\).
Here, \(x_j\) denotes the state variable for species \(j\), and \(X_j^+\) contains its effective positive observations along the default fitting trajectory.

\paragraph{Distance to fixed candidates.}
For each factor, we compare its reference \(K\) with the corresponding species-specific candidate set \(\mathcal{K}_j\) from Appendix~\ref{app:kinetic_library}, retaining the configured positivity threshold, floor, and deduplication rules.
Figure~\ref{fig:supp_kinetic_parameter_coverage}(c) reports the empirical cumulative distribution of
\[
d_K=\min_{K_c\in\mathcal{K}_j}\left|\log_{10}\frac{K_{\mathrm{ref}}}{K_c}\right|.
\]
Of the 189 reference values, 53 (28.0\%) lie within a factor of two of the nearest candidate and 118 (62.4\%) within a factor of ten.
Figure~\ref{fig:supp_kinetic_parameter_coverage}(c) shows that the finite \(K\) grid can remain far from reference values.
Even when the candidate set contains the reference exponent, we must distinguish combinations of \(K\), \(n\), and kinetic form within each proposed reaction structure.
Jointly identifying reaction structure and kinetic laws remains challenging when candidate laws produce similar responses over the observed state range.

\begin{figure}[!htbp]
  \centering
  \includegraphics[width=\linewidth]{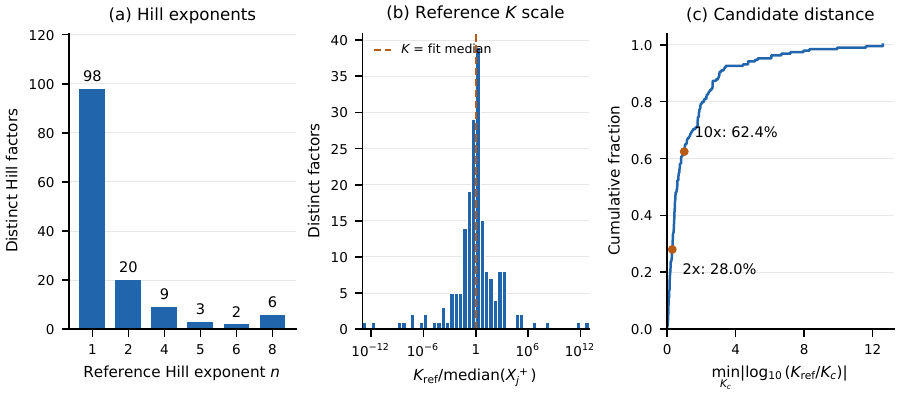}
  \caption{\rev{\textbf{Reference \(n\) and \(K\) distributions and distances to \(K\) candidates.}
  Counts use distinct kinetic factors within each Main or Challenge system.
  (a) Reference exponents of 138 Hill factors.
  (b) Reference \(K\) divided by the corresponding median of \(X_j^+\) for 189 factors.
  The dashed line marks a ratio of one.
  (c) Cumulative fraction of factors versus the log-distance to the nearest fixed \(K\) candidate.
  Orange markers indicate distances within factors of two and ten.}}
  \label{fig:supp_kinetic_parameter_coverage}
\end{figure}
\end{revision}

%% file: Table/supp_fpms_ablation.tex
\begin{table}[htbp]
  \centering
  \caption{\textbf{FPMS component ablations in KineSelect.}
  Base combines fitting-trajectory simulation loss and expanded ODE
  expression-tree complexity. The next two rows add derivative loss
  \(\mathcal{L}^{\mathrm{der}}\) or specialized kinetic-component count
  \(C^{\mathrm{spec}}\) to Base; Full FPMS includes both. Overall OOD
  \(R^2\) uses all 198 systems, including Stress, whereas Overall
  reaction F1 uses the 161 reaction-eligible systems. Main and Challenge
  use 95 and 66 systems for both metrics.
  Bold and underlined values denote the best and second-best results in each column.}
  \label{tab:supp_fpms_ablation}
  \begingroup
  \footnotesize
  \setlength{\tabcolsep}{2pt}
  \renewcommand{\arraystretch}{1.15}
  \newcolumntype{Y}{>{\centering\arraybackslash}X}
  \begin{tabularx}{\linewidth}{@{}l*{6}{Y}@{}}
    \toprule
    \multirow{2}{*}{\textbf{Selection}} & \multicolumn{3}{c}{\textbf{OOD \(R^2\)}\(\uparrow\)} & \multicolumn{3}{c}{\textbf{Reaction F1}\(\uparrow\)} \\
    \cmidrule(lr){2-4}\cmidrule(lr){5-7}
     & Overall & Main & Challenge & Overall & Main & Challenge \\
    \midrule
    Base & \underline{0.5141} & \textbf{0.6241} & 0.4040 & 0.4272 & 0.5315 & 0.2772 \\
    \(+\mathcal{L}^{\mathrm{der}}\) & \textbf{0.5158} & 0.6162 & \underline{0.4194} & \underline{0.4363} & \underline{0.5459} & 0.2784 \\
    \(+C^{\mathrm{spec}}\) & 0.5077 & \underline{0.6176} & 0.4060 & 0.4361 & 0.5424 & \underline{0.2832} \\
    Full FPMS & 0.5081 & 0.6011 & \textbf{0.4373} & \textbf{0.4475} & \textbf{0.5604} & \textbf{0.2848} \\
    \bottomrule
  \end{tabularx}
  \endgroup
\end{table}

%% file: Table/supp_noisy_trajectory.tex
  \centering
  \caption{\textbf{Trajectory fidelity under observation noise.}
  Each noise level reports OOD \(R^2\) and \(\mathrm{Acc}@.90\) (Acc.)
  over all 198 systems; higher is better for both.
  Bold and underlined values denote the best and second-best results
  for each metric and noise level, including ties.}
  \label{tab:supp_noisy_trajectory}
  \begingroup
  \footnotesize
  \setlength{\tabcolsep}{1.5pt}
  \renewcommand{\arraystretch}{1.15}
  \newcolumntype{Y}{>{\centering\arraybackslash}X}
  \begin{tabularx}{\linewidth}{@{}l*{10}{Y}@{}}
    \toprule
    \multirow{2}{*}{\textbf{Method}} & \multicolumn{2}{c}{\textbf{Noise-free}} & \multicolumn{2}{c}{\textbf{0.1\%}} & \multicolumn{2}{c}{\textbf{1\%}} & \multicolumn{2}{c}{\textbf{5\%}} & \multicolumn{2}{c}{\textbf{10\%}} \\
    \cmidrule(lr){2-3}\cmidrule(lr){4-5}\cmidrule(lr){6-7}\cmidrule(lr){8-9}\cmidrule(lr){10-11}
     & \(R^2\) & Acc. & \(R^2\) & Acc. & \(R^2\) & Acc. & \(R^2\) & Acc. & \(R^2\) & Acc. \\
    \midrule
    KineSelect & \textbf{0.508} & \textbf{0.374} & \textbf{0.532} & \textbf{0.399} & \textbf{0.485} & \textbf{0.354} & \textbf{0.421} & \textbf{0.278} & \textbf{0.383} & \textbf{0.217} \\
    SISR & \underline{0.426} & \underline{0.328} & \underline{0.418} & \underline{0.328} & \underline{0.407} & \underline{0.308} & \underline{0.410} & \underline{0.268} & \underline{0.363} & \underline{0.202} \\
    Reactive SINDy & 0.142 & 0.091 & 0.135 & 0.091 & 0.094 & 0.051 & 0.016 & 0.005 & 0.010 & 0.000 \\
    FFX & 0.367 & 0.222 & 0.318 & 0.232 & 0.226 & 0.152 & 0.175 & 0.106 & 0.162 & 0.066 \\
    SINDy & 0.033 & 0.030 & 0.034 & 0.030 & 0.047 & 0.030 & 0.038 & 0.030 & 0.037 & 0.030 \\
    SINDy-PI & 0.310 & 0.273 & 0.091 & 0.066 & 0.073 & 0.035 & 0.058 & 0.020 & 0.048 & 0.015 \\
    PySR & 0.233 & 0.177 & 0.227 & 0.172 & 0.186 & 0.111 & 0.126 & 0.071 & 0.094 & 0.040 \\
    ODEFormer & 0.136 & 0.061 & 0.121 & 0.045 & 0.109 & 0.056 & 0.096 & 0.045 & 0.094 & 0.035 \\
    \bottomrule
  \end{tabularx}
  \endgroup

%% file: Table/supp_noisy_reaction.tex
  \centering
  \caption{\textbf{Reaction F1 under observation noise.}
  Overall, Main, and Challenge report mean reaction F1 over 161, 95, and
  66 systems, respectively.
  Bold and underlined values denote the best and second-best results
  within each noise level and subset.}
  \label{tab:supp_noisy_reaction}
  \begingroup
  \footnotesize
  \setlength{\tabcolsep}{4pt}
  \renewcommand{\arraystretch}{1.15}
  \newcolumntype{Y}{>{\centering\arraybackslash}X}
  \begin{tabularx}{\linewidth}{@{}cl*{3}{Y}@{}}
    \toprule
    \textbf{Noise} & \textbf{Method} & \textbf{Overall}\(\uparrow\) & \textbf{Main}\(\uparrow\) & \textbf{Challenge}\(\uparrow\) \\
    \midrule
    \multirow{3}{*}{Noise-free} & KineSelect & \textbf{0.447} & \textbf{0.560} & \textbf{0.285} \\
     & SISR & \underline{0.391} & \underline{0.507} & \underline{0.225} \\
     & Reactive SINDy & 0.292 & 0.363 & 0.191 \\
    \addlinespace
    \multirow{3}{*}{0.1\%} & KineSelect & \textbf{0.453} & \textbf{0.566} & \textbf{0.291} \\
     & SISR & \underline{0.371} & \underline{0.467} & \underline{0.233} \\
     & Reactive SINDy & 0.278 & 0.335 & 0.195 \\
    \addlinespace
    \multirow{3}{*}{1\%} & KineSelect & \textbf{0.426} & \textbf{0.511} & \textbf{0.304} \\
     & SISR & \underline{0.378} & \underline{0.464} & \underline{0.253} \\
     & Reactive SINDy & 0.209 & 0.255 & 0.142 \\
    \addlinespace
    \multirow{3}{*}{5\%} & KineSelect & \underline{0.344} & \underline{0.392} & \textbf{0.277} \\
     & SISR & \textbf{0.350} & \textbf{0.432} & \underline{0.232} \\
     & Reactive SINDy & 0.143 & 0.169 & 0.106 \\
    \addlinespace
    \multirow{3}{*}{10\%} & KineSelect & \underline{0.306} & \underline{0.359} & \underline{0.229} \\
     & SISR & \textbf{0.317} & \textbf{0.377} & \textbf{0.231} \\
     & Reactive SINDy & 0.137 & 0.165 & 0.096 \\
    \bottomrule
  \end{tabularx}
  \endgroup

%% file: supplementary/noisy_multirun_results.tex
\begin{revision}
\subsubsection{Robustness across Noise Seeds}
\label{app:noisy_multirun_results}

\paragraph{Experimental setup and aggregation.}
We evaluate noise robustness at relative noise levels of 0.1\%, 1\%, 5\%, and 10\%, using different random seeds to generate observation noise as described in Appendix~\ref{app:id_ood_noise}.
The summaries include the original run with noise seed 2026.
All the configurations include seeds 2026, 20260928, and 20261216, giving three.
We average scores across systems within each run, then report the mean and standard deviation across noise seeds.
Trajectory metrics retain 198 systems, and reaction precision, recall, and F1 retain 161 Main + Challenge systems.

\paragraph{Performance across noise levels.}
KineSelect has the highest mean Overall OOD \(R^2\) and reaction F1 at 0.1\% and 1\% noise among the evaluated configurations (Tables~\ref{tab:supp_noisy_multirun_trajectory} and~\ref{tab:supp_noisy_multirun_reaction}).
With FPMS shared, KineSelect has higher mean OOD \(R^2\) than SISR + FPMS at all four noise levels, while their reaction F1 means become close at 5\% and 10\% noise.
At 5\% noise, KineSelect and SISR have similar mean OOD \(R^2\) (0.419 and 0.417), while SISR has higher reaction F1 (0.362 versus 0.354).

\input{Table/supp_noisy_multirun_trajectory}
\clearpage
\input{Table/supp_noisy_multirun_reaction}

\paragraph{Final selection under observation noise.}
SISR achieves higher mean OOD \(R^2\) and reaction F1 with \(\alpha\)-threshold selection than with FPMS at 5\% and 10\% noise.
FPMS combines derivative loss, simulation loss, and complexity, whereas \(\alpha\)-threshold selection applies a simulation-loss tolerance
followed by complexity minimization.
Noisy derivative estimates may therefore affect FPMS's final ranking in addition to the candidate fitting shared by both variants.
This difference offers one possible explanation for the stronger high-noise performance of \(\alpha\)-threshold selection.

\end{revision}

%% file: Table/supp_noisy_multirun_trajectory.tex
\begin{table}[htbp]
  \begin{revision}
  \centering
  \caption{\rev{\textbf{OOD trajectory fidelity across noise seeds.} Each noise level reports mean OOD \(R^2\) and \(\mathrm{Acc}@.90\) (Acc.) over 198 systems. SISR + FPMS selects from each SISR candidate archive with FPMS. Subscripts give standard deviation across noise seeds. All the configurations use three noise realizations. Bold and underlining mark the highest and second-highest means within each noise level.}}
  \label{tab:supp_noisy_multirun_trajectory}
  \begingroup
  \footnotesize
  \setlength{\tabcolsep}{1.2pt}
  \renewcommand{\arraystretch}{1.18}
  \newcolumntype{Y}{>{\centering\arraybackslash}X}
  \begin{tabularx}{\linewidth}{@{}l@{\hspace{2pt}}*{8}{Y}@{}}
    \toprule
    \multirow{2}{*}{\textbf{Method}} & \multicolumn{2}{c}{\textbf{0.1\%}} & \multicolumn{2}{c}{\textbf{1\%}} & \multicolumn{2}{c}{\textbf{5\%}} & \multicolumn{2}{c}{\textbf{10\%}} \\
    \cmidrule(lr){2-3}\cmidrule(lr){4-5}\cmidrule(lr){6-7}\cmidrule(lr){8-9}
     & \(R^2\) & Acc. & \(R^2\) & Acc. & \(R^2\) & Acc. & \(R^2\) & Acc. \\
    \midrule
    FFX & \(\text{0.326}_{\text{\(\pm\)0.009}}\) & \(\text{0.231}_{\text{\(\pm\)0.008}}\) & \(\text{0.280}_{\text{\(\pm\)0.046}}\) & \(\text{0.178}_{\text{\(\pm\)0.025}}\) & \(\text{0.264}_{\text{\(\pm\)0.077}}\) & \(\text{0.148}_{\text{\(\pm\)0.037}}\) & \(\text{0.226}_{\text{\(\pm\)0.055}}\) & \(\text{0.108}_{\text{\(\pm\)0.039}}\) \\
    PySR & \(\text{0.248}_{\text{\(\pm\)0.021}}\) & \(\text{0.180}_{\text{\(\pm\)0.011}}\) & \(\text{0.240}_{\text{\(\pm\)0.049}}\) & \(\text{0.162}_{\text{\(\pm\)0.044}}\) & \(\text{0.220}_{\text{\(\pm\)0.090}}\) & \(\text{0.125}_{\text{\(\pm\)0.053}}\) & \(\text{0.183}_{\text{\(\pm\)0.081}}\) & \(\text{0.098}_{\text{\(\pm\)0.054}}\) \\
    \midrule
    SINDy & \(\text{0.035}_{\text{\(\pm\)0.003}}\) & \(\text{0.030}_{\text{\(\pm\)0.000}}\) & \(\text{0.044}_{\text{\(\pm\)0.005}}\) & \(\text{0.034}_{\text{\(\pm\)0.006}}\) & \(\text{0.042}_{\text{\(\pm\)0.009}}\) & \(\text{0.030}_{\text{\(\pm\)0.000}}\) & \(\text{0.038}_{\text{\(\pm\)0.004}}\) & \(\text{0.032}_{\text{\(\pm\)0.008}}\) \\
    SINDy-PI & \(\text{0.107}_{\text{\(\pm\)0.015}}\) & \(\text{0.072}_{\text{\(\pm\)0.008}}\) & \(\text{0.084}_{\text{\(\pm\)0.010}}\) & \(\text{0.040}_{\text{\(\pm\)0.005}}\) & \(\text{0.061}_{\text{\(\pm\)0.006}}\) & \(\text{0.022}_{\text{\(\pm\)0.003}}\) & \(\text{0.049}_{\text{\(\pm\)0.004}}\) & \(\text{0.024}_{\text{\(\pm\)0.008}}\) \\
    Reactive SINDy & \(\text{0.136}_{\text{\(\pm\)0.015}}\) & \(\text{0.088}_{\text{\(\pm\)0.006}}\) & \(\text{0.094}_{\text{\(\pm\)0.002}}\) & \(\text{0.047}_{\text{\(\pm\)0.003}}\) & \(\text{0.015}_{\text{\(\pm\)0.001}}\) & \(\text{0.003}_{\text{\(\pm\)0.003}}\) & \(\text{0.007}_{\text{\(\pm\)0.002}}\) & \(\text{0.000}_{\text{\(\pm\)0.000}}\) \\
    \midrule
    ODEFormer & \(\text{0.123}_{\text{\(\pm\)0.002}}\) & \(\text{0.056}_{\text{\(\pm\)0.010}}\) & \(\text{0.116}_{\text{\(\pm\)0.006}}\) & \(\text{0.057}_{\text{\(\pm\)0.003}}\) & \(\text{0.106}_{\text{\(\pm\)0.009}}\) & \(\text{0.049}_{\text{\(\pm\)0.006}}\) & \(\text{0.096}_{\text{\(\pm\)0.003}}\) & \(\text{0.035}_{\text{\(\pm\)0.000}}\) \\
    \midrule
    SISR & \(\underline{\text{0.436}}_{\text{\(\pm\)0.016}}\) & \(\underline{\text{0.345}}_{\text{\(\pm\)0.018}}\) & \(\underline{\text{0.426}}_{\text{\(\pm\)0.018}}\) & \(\underline{\text{0.323}}_{\text{\(\pm\)0.015}}\) & \(\underline{\text{0.417}}_{\text{\(\pm\)0.006}}\) & \(\underline{\text{0.271}}_{\text{\(\pm\)0.020}}\) & \(\text{\bfseries 0.376}_{\text{\(\pm\)0.031}}\) & \(\text{\bfseries 0.226}_{\text{\(\pm\)0.021}}\) \\
    SISR + FPMS & \(\text{0.369}_{\text{\(\pm\)0.024}}\) & \(\text{0.301}_{\text{\(\pm\)0.006}}\) & \(\text{0.352}_{\text{\(\pm\)0.042}}\) & \(\text{0.276}_{\text{\(\pm\)0.015}}\) & \(\text{0.340}_{\text{\(\pm\)0.054}}\) & \(\text{0.236}_{\text{\(\pm\)0.020}}\) & \(\text{0.278}_{\text{\(\pm\)0.061}}\) & \(\text{0.195}_{\text{\(\pm\)0.037}}\) \\
    KineSelect & \(\text{\bfseries 0.532}_{\text{\(\pm\)0.001}}\) & \(\text{\bfseries 0.412}_{\text{\(\pm\)0.018}}\) & \(\text{\bfseries 0.491}_{\text{\(\pm\)0.008}}\) & \(\text{\bfseries 0.361}_{\text{\(\pm\)0.011}}\) & \(\text{\bfseries 0.419}_{\text{\(\pm\)0.004}}\) & \(\text{\bfseries 0.278}_{\text{\(\pm\)0.000}}\) & \(\underline{\text{0.363}}_{\text{\(\pm\)0.028}}\) & \(\underline{\text{0.207}}_{\text{\(\pm\)0.014}}\) \\
    \bottomrule
  \end{tabularx}
  \par
  \endgroup
  \end{revision}
\end{table}

%% file: Table/supp_noisy_multirun_reaction.tex
\begin{table}[htbp]
  \begin{revision}
  \centering
  \caption{\rev{\textbf{Reaction recovery from noisy trajectory data.} Overall, Main, and Challenge contain 161, 95, and 66 systems, respectively. Subscripts give standard deviation across noise seeds. All the configurations use three noise realizations. Bold and underlining mark the highest and second-highest means within each noise level and subset.}}
  \label{tab:supp_noisy_multirun_reaction}
  \begingroup
  \scriptsize
  \setlength{\tabcolsep}{0.4pt}
  \renewcommand{\arraystretch}{1.18}
  \newcolumntype{Y}{>{\centering\arraybackslash}X}
  \begin{tabularx}{\linewidth}{@{}c@{\hspace{3pt}}l@{\hspace{2pt}}*{9}{Y}@{}}
    \toprule
    \multirow{2}{*}{\textbf{Noise}} & \multirow{2}{*}{\textbf{Method}} & \multicolumn{3}{c}{\textbf{Overall}} & \multicolumn{3}{c}{\textbf{Main}} & \multicolumn{3}{c}{\textbf{Challenge}} \\
    \cmidrule(lr){3-5}\cmidrule(lr){6-8}\cmidrule(lr){9-11}
     & & Precision & Recall & F1 & Precision & Recall & F1 & Precision & Recall & F1 \\
    \midrule
    \multirow{4}{*}{0.1\%} & Reactive SINDy & \(\text{0.329}_{\text{\(\pm\)0.004}}\) & \(\text{0.337}_{\text{\(\pm\)0.011}}\) & \(\text{0.276}_{\text{\(\pm\)0.007}}\) & \(\text{0.394}_{\text{\(\pm\)0.004}}\) & \(\text{0.432}_{\text{\(\pm\)0.011}}\) & \(\text{0.338}_{\text{\(\pm\)0.006}}\) & \(\text{0.236}_{\text{\(\pm\)0.008}}\) & \(\text{0.201}_{\text{\(\pm\)0.011}}\) & \(\text{0.187}_{\text{\(\pm\)0.011}}\) \\
     & SISR & \(\text{0.415}_{\text{\(\pm\)0.014}}\) & \(\underline{\text{0.409}}_{\text{\(\pm\)0.019}}\) & \(\text{0.388}_{\text{\(\pm\)0.015}}\) & \(\text{0.517}_{\text{\(\pm\)0.030}}\) & \(\underline{\text{0.527}}_{\text{\(\pm\)0.030}}\) & \(\text{0.494}_{\text{\(\pm\)0.027}}\) & \(\text{0.270}_{\text{\(\pm\)0.011}}\) & \(\underline{\text{0.240}}_{\text{\(\pm\)0.012}}\) & \(\text{0.235}_{\text{\(\pm\)0.008}}\) \\
     & SISR + FPMS & \(\text{\bfseries 0.541}_{\text{\(\pm\)0.038}}\) & \(\text{0.387}_{\text{\(\pm\)0.008}}\) & \(\underline{\text{0.420}}_{\text{\(\pm\)0.014}}\) & \(\text{\bfseries 0.665}_{\text{\(\pm\)0.052}}\) & \(\text{0.495}_{\text{\(\pm\)0.013}}\) & \(\underline{\text{0.533}}_{\text{\(\pm\)0.022}}\) & \(\text{\bfseries 0.363}_{\text{\(\pm\)0.021}}\) & \(\text{0.231}_{\text{\(\pm\)0.005}}\) & \(\underline{\text{0.258}}_{\text{\(\pm\)0.003}}\) \\
     & KineSelect & \(\underline{\text{0.460}}_{\text{\(\pm\)0.001}}\) & \(\text{\bfseries 0.474}_{\text{\(\pm\)0.005}}\) & \(\text{\bfseries 0.454}_{\text{\(\pm\)0.001}}\) & \(\underline{\text{0.561}}_{\text{\(\pm\)0.002}}\) & \(\text{\bfseries 0.590}_{\text{\(\pm\)0.001}}\) & \(\text{\bfseries 0.563}_{\text{\(\pm\)0.004}}\) & \(\underline{\text{0.316}}_{\text{\(\pm\)0.004}}\) & \(\text{\bfseries 0.307}_{\text{\(\pm\)0.012}}\) & \(\text{\bfseries 0.296}_{\text{\(\pm\)0.007}}\) \\
    \midrule
    \multirow{4}{*}{1\%} & Reactive SINDy & \(\text{0.288}_{\text{\(\pm\)0.005}}\) & \(\text{0.240}_{\text{\(\pm\)0.005}}\) & \(\text{0.213}_{\text{\(\pm\)0.004}}\) & \(\text{0.350}_{\text{\(\pm\)0.015}}\) & \(\text{0.303}_{\text{\(\pm\)0.003}}\) & \(\text{0.262}_{\text{\(\pm\)0.009}}\) & \(\text{0.200}_{\text{\(\pm\)0.011}}\) & \(\text{0.149}_{\text{\(\pm\)0.016}}\) & \(\text{0.142}_{\text{\(\pm\)0.016}}\) \\
     & SISR & \(\text{0.417}_{\text{\(\pm\)0.016}}\) & \(\underline{\text{0.400}}_{\text{\(\pm\)0.007}}\) & \(\text{0.386}_{\text{\(\pm\)0.010}}\) & \(\text{0.512}_{\text{\(\pm\)0.030}}\) & \(\underline{\text{0.507}}_{\text{\(\pm\)0.018}}\) & \(\text{0.486}_{\text{\(\pm\)0.021}}\) & \(\text{0.281}_{\text{\(\pm\)0.006}}\) & \(\underline{\text{0.247}}_{\text{\(\pm\)0.009}}\) & \(\text{0.242}_{\text{\(\pm\)0.010}}\) \\
     & SISR + FPMS & \(\text{\bfseries 0.527}_{\text{\(\pm\)0.015}}\) & \(\text{0.360}_{\text{\(\pm\)0.017}}\) & \(\underline{\text{0.397}}_{\text{\(\pm\)0.012}}\) & \(\text{\bfseries 0.633}_{\text{\(\pm\)0.021}}\) & \(\text{0.456}_{\text{\(\pm\)0.027}}\) & \(\underline{\text{0.498}}_{\text{\(\pm\)0.019}}\) & \(\text{\bfseries 0.374}_{\text{\(\pm\)0.011}}\) & \(\text{0.221}_{\text{\(\pm\)0.012}}\) & \(\underline{\text{0.251}}_{\text{\(\pm\)0.009}}\) \\
     & KineSelect & \(\underline{\text{0.452}}_{\text{\(\pm\)0.007}}\) & \(\text{\bfseries 0.423}_{\text{\(\pm\)0.002}}\) & \(\text{\bfseries 0.423}_{\text{\(\pm\)0.004}}\) & \(\underline{\text{0.531}}_{\text{\(\pm\)0.010}}\) & \(\text{\bfseries 0.511}_{\text{\(\pm\)0.001}}\) & \(\text{\bfseries 0.507}_{\text{\(\pm\)0.006}}\) & \(\underline{\text{0.338}}_{\text{\(\pm\)0.001}}\) & \(\text{\bfseries 0.295}_{\text{\(\pm\)0.003}}\) & \(\text{\bfseries 0.303}_{\text{\(\pm\)0.002}}\) \\
    \midrule
    \multirow{4}{*}{5\%} & Reactive SINDy & \(\text{0.262}_{\text{\(\pm\)0.004}}\) & \(\text{0.131}_{\text{\(\pm\)0.002}}\) & \(\text{0.145}_{\text{\(\pm\)0.003}}\) & \(\text{0.311}_{\text{\(\pm\)0.004}}\) & \(\text{0.153}_{\text{\(\pm\)0.007}}\) & \(\text{0.174}_{\text{\(\pm\)0.006}}\) & \(\text{0.192}_{\text{\(\pm\)0.009}}\) & \(\text{0.099}_{\text{\(\pm\)0.004}}\) & \(\text{0.104}_{\text{\(\pm\)0.002}}\) \\
     & SISR & \(\underline{\text{0.396}}_{\text{\(\pm\)0.021}}\) & \(\text{\bfseries 0.369}_{\text{\(\pm\)0.010}}\) & \(\text{\bfseries 0.362}_{\text{\(\pm\)0.012}}\) & \(\underline{\text{0.468}}_{\text{\(\pm\)0.018}}\) & \(\text{\bfseries 0.465}_{\text{\(\pm\)0.012}}\) & \(\text{\bfseries 0.445}_{\text{\(\pm\)0.013}}\) & \(\text{0.294}_{\text{\(\pm\)0.027}}\) & \(\underline{\text{0.232}}_{\text{\(\pm\)0.008}}\) & \(\text{0.242}_{\text{\(\pm\)0.010}}\) \\
     & SISR + FPMS & \(\text{\bfseries 0.469}_{\text{\(\pm\)0.005}}\) & \(\text{0.313}_{\text{\(\pm\)0.019}}\) & \(\text{0.350}_{\text{\(\pm\)0.016}}\) & \(\text{\bfseries 0.541}_{\text{\(\pm\)0.014}}\) & \(\text{0.383}_{\text{\(\pm\)0.022}}\) & \(\underline{\text{0.423}}_{\text{\(\pm\)0.018}}\) & \(\text{\bfseries 0.365}_{\text{\(\pm\)0.017}}\) & \(\text{0.213}_{\text{\(\pm\)0.015}}\) & \(\underline{\text{0.246}}_{\text{\(\pm\)0.015}}\) \\
     & KineSelect & \(\text{0.394}_{\text{\(\pm\)0.019}}\) & \(\underline{\text{0.341}}_{\text{\(\pm\)0.009}}\) & \(\underline{\text{0.354}}_{\text{\(\pm\)0.013}}\) & \(\text{0.440}_{\text{\(\pm\)0.027}}\) & \(\underline{\text{0.398}}_{\text{\(\pm\)0.015}}\) & \(\text{0.406}_{\text{\(\pm\)0.020}}\) & \(\underline{\text{0.327}}_{\text{\(\pm\)0.006}}\) & \(\text{\bfseries 0.258}_{\text{\(\pm\)0.001}}\) & \(\text{\bfseries 0.278}_{\text{\(\pm\)0.002}}\) \\
    \midrule
    \multirow{4}{*}{10\%} & Reactive SINDy & \(\text{0.259}_{\text{\(\pm\)0.008}}\) & \(\text{0.117}_{\text{\(\pm\)0.004}}\) & \(\text{0.137}_{\text{\(\pm\)0.002}}\) & \(\text{0.317}_{\text{\(\pm\)0.014}}\) & \(\text{0.143}_{\text{\(\pm\)0.004}}\) & \(\text{0.167}_{\text{\(\pm\)0.003}}\) & \(\text{0.177}_{\text{\(\pm\)0.004}}\) & \(\text{0.080}_{\text{\(\pm\)0.004}}\) & \(\text{0.093}_{\text{\(\pm\)0.002}}\) \\
     & SISR & \(\underline{\text{0.355}}_{\text{\(\pm\)0.004}}\) & \(\text{\bfseries 0.325}_{\text{\(\pm\)0.003}}\) & \(\text{\bfseries 0.318}_{\text{\(\pm\)0.001}}\) & \(\underline{\text{0.407}}_{\text{\(\pm\)0.016}}\) & \(\text{\bfseries 0.391}_{\text{\(\pm\)0.010}}\) & \(\text{\bfseries 0.376}_{\text{\(\pm\)0.009}}\) & \(\underline{\text{0.281}}_{\text{\(\pm\)0.030}}\) & \(\underline{\text{0.229}}_{\text{\(\pm\)0.008}}\) & \(\underline{\text{0.235}}_{\text{\(\pm\)0.012}}\) \\
     & SISR + FPMS & \(\text{\bfseries 0.407}_{\text{\(\pm\)0.023}}\) & \(\text{0.276}_{\text{\(\pm\)0.029}}\) & \(\text{0.306}_{\text{\(\pm\)0.028}}\) & \(\text{\bfseries 0.466}_{\text{\(\pm\)0.046}}\) & \(\text{0.330}_{\text{\(\pm\)0.040}}\) & \(\underline{\text{0.361}}_{\text{\(\pm\)0.040}}\) & \(\text{\bfseries 0.322}_{\text{\(\pm\)0.012}}\) & \(\text{0.198}_{\text{\(\pm\)0.022}}\) & \(\text{0.227}_{\text{\(\pm\)0.022}}\) \\
     & KineSelect & \(\text{0.344}_{\text{\(\pm\)0.008}}\) & \(\underline{\text{0.301}}_{\text{\(\pm\)0.001}}\) & \(\underline{\text{0.309}}_{\text{\(\pm\)0.004}}\) & \(\text{0.388}_{\text{\(\pm\)0.008}}\) & \(\underline{\text{0.343}}_{\text{\(\pm\)0.015}}\) & \(\text{0.351}_{\text{\(\pm\)0.012}}\) & \(\text{0.280}_{\text{\(\pm\)0.030}}\) & \(\text{\bfseries 0.241}_{\text{\(\pm\)0.024}}\) & \(\text{\bfseries 0.249}_{\text{\(\pm\)0.028}}\) \\
    \bottomrule
  \end{tabularx}
  \par
  \endgroup
  \end{revision}
\end{table}

%% file: Table/supp_representative_cases.tex
\begingroup
\small
\definecolor{CfiveUnmatched}{HTML}{B34450}
\setlength{\tabcolsep}{4pt}
\renewcommand{\arraystretch}{1.15}
\setlength{\LTleft}{0pt}
\setlength{\LTright}{0pt}
\setlength{\LTcapwidth}{\linewidth}
\begin{longtable}{@{}p{\linewidth}@{}}
\caption{\textbf{Reference and recovered equations for the six representative cases.}
The three columns show Ground Truth, SISR, and KineSelect.
Ground-truth equations are black.
Predicted reaction-signature occurrences are shown in \textcolor{black}{black (matched)} or \textcolor{CfiveUnmatched}{muted red (unmatched)}.
Parameter indices are local to each model and retain shared parameters across its states.
OOD $R^2$ denotes the unclipped system-level score.
Negative values are retained for the individual case studies.
Precision, recall, and F1 are computed from reaction-signature matches.
TP/FP/FN are occurrence counts.
Case labels match Figure~\ref{fig:supp_representative_trajectories}.}
\label{tab:supp_representative_cases} \\
\endfirsthead
\caption[]{Reference and recovered equations for the six representative cases (continued).} \\
\endhead
\endfoot
\endlastfoot
\begin{minipage}[t]{\linewidth}
\centering
\textbf{(a) BIOMD0000000157 (Zatorsky2006)}\par\smallskip
\begin{tabular}{@{}*{3}{>{\centering\arraybackslash}p{\dimexpr(\linewidth-16pt)/3\relax}}@{}}
\toprule
\textbf{Ground Truth} & \textbf{SISR} & \textbf{KineSelect} \\
\midrule
\begin{minipage}[t]{\linewidth}\vspace{0pt}\centering
\(\left\{\begin{aligned}
\dot{x}_{0} &={} \textcolor{black}{k_{1}} \textcolor{black}{{}-\frac{k_{2}\,x_{0}\,x_{1}}{K_{\mathrm{M},2} + x_{0}}} \\[3pt]
\dot{x}_{1} &={} \textcolor{black}{k_{3}\,x_{2}} \textcolor{black}{{}-k_{4}\,x_{1}} \\[3pt]
\dot{x}_{2} &={} \textcolor{black}{k_{5}\,x_{0}} \textcolor{black}{{}-k_{6}\,x_{2}}
\end{aligned}\right.\)
\end{minipage} &
\begin{minipage}[t]{\linewidth}\vspace{0pt}\centering
\(\left\{\begin{aligned}
\dot{x}_{0} &={} \textcolor{CfiveUnmatched}{-x_{0}\,x_{1}\,\left(k_{1} + k_{3}\right)} \\
&\phantom{{}={}} \textcolor{CfiveUnmatched}{{}+k_{2}\,{x_{1}}^{2}} \textcolor{CfiveUnmatched}{{}+k_{4}\,x_{0}} \\[3pt]
\dot{x}_{1} &={} \textcolor{CfiveUnmatched}{x_{0}\,x_{1}\,\left(k_{1} - k_{3}\right)} \\
&\phantom{{}={}} \textcolor{CfiveUnmatched}{{}-2\,k_{2}\,{x_{1}}^{2}} \\[3pt]
\dot{x}_{2} &={} \textcolor{CfiveUnmatched}{x_{0}\,x_{1}\,\left(k_{1} + k_{3}\right)} \\
&\phantom{{}={}} \textcolor{CfiveUnmatched}{{}+k_{2}\,{x_{1}}^{2}} \textcolor{CfiveUnmatched}{{}-2\,k_{5}\,{x_{2}}^{2}}
\end{aligned}\right.\)
\end{minipage} &
\begin{minipage}[t]{\linewidth}\vspace{0pt}\centering
\(\left\{\begin{aligned}
\dot{x}_{0} &={} \textcolor{CfiveUnmatched}{k_{2}\,\frac{K_{\mathrm{H},2}}{K_{\mathrm{H},2} + x_{1}}} \textcolor{CfiveUnmatched}{{}+k_{4}\,x_{2}} \\
&\phantom{{}={}} \textcolor{black}{{}-\frac{k_{6}\,x_{0}\,x_{1}}{K_{\mathrm{M},6} + x_{0}}} \\[3pt]
\dot{x}_{1} &={} \textcolor{black}{k_{3}\,x_{2}} \textcolor{black}{{}-k_{5}\,x_{1}} \\[3pt]
\dot{x}_{2} &={} \textcolor{black}{k_{1}\,x_{0}} \textcolor{black}{{}-k_{3}\,x_{2}} \textcolor{CfiveUnmatched}{{}-k_{4}\,x_{2}}
\end{aligned}\right.\)
\end{minipage} \\
\end{tabular}\par\smallskip
\begin{tabularx}{\linewidth}{@{}l*{5}{>{\centering\arraybackslash}X}@{}}
\toprule
Method & OOD $R^2$ & Precision & Recall & F1 & TP/FP/FN \\
\midrule
SISR & -0.236 & 0.000 & 0.000 & 0.000 & 0/8/6 \\
KineSelect & 0.995 & 0.625 & 0.833 & 0.714 & 5/3/1 \\
\bottomrule
\end{tabularx}
\end{minipage} \\
\addlinespace[16pt]
\begin{minipage}[t]{\linewidth}
\centering
\textbf{(b) BIOMD0000000546 (Miao2010)}\par\smallskip
\begin{tabular}{@{}*{3}{>{\centering\arraybackslash}p{\dimexpr(\linewidth-16pt)/3\relax}}@{}}
\toprule
\textbf{Ground Truth} & \textbf{SISR} & \textbf{KineSelect} \\
\midrule
\begin{minipage}[t]{\linewidth}\vspace{0pt}\centering
\(\left\{\begin{aligned}
\dot{x}_{0} &={} \textcolor{black}{k_{1}\,x_{0}} \textcolor{black}{{}-k_{2}\,x_{0}\,x_{2}} \\[3pt]
\dot{x}_{1} &={} \textcolor{black}{-k_{3}\,x_{1}} \textcolor{black}{{}+k_{4}\,x_{0}\,x_{2}} \\[3pt]
\dot{x}_{2} &={} \textcolor{black}{k_{5}\,x_{1}} \textcolor{black}{{}-k_{6}\,x_{2}}
\end{aligned}\right.\)
\end{minipage} &
\begin{minipage}[t]{\linewidth}\vspace{0pt}\centering
\(\left\{\begin{aligned}
\dot{x}_{0} &={} \textcolor{black}{-k_{3}\,x_{0}\,x_{2}} \\[3pt]
\dot{x}_{1} &={} \textcolor{black}{k_{3}\,x_{0}\,x_{2}} \textcolor{CfiveUnmatched}{{}-k_{4}\,x_{1}\,x_{2}} \\[3pt]
\dot{x}_{2} &={} \textcolor{CfiveUnmatched}{2\,k_{1}\,x_{0}\,x_{2}} \textcolor{CfiveUnmatched}{{}+k_{2}\,x_{1}\,x_{2}} \\
&\phantom{{}={}} \textcolor{CfiveUnmatched}{{}-k_{5}\,{x_{2}}^{2}}
\end{aligned}\right.\)
\end{minipage} &
\begin{minipage}[t]{\linewidth}\vspace{0pt}\centering
\(\left\{\begin{aligned}
\dot{x}_{0} &={} \textcolor{black}{-k_{4}\,x_{0}\,x_{2}} \\[3pt]
\dot{x}_{1} &={} \textcolor{black}{-k_{3}\,x_{1}} \textcolor{black}{{}+k_{4}\,x_{0}\,x_{2}} \\[3pt]
\dot{x}_{2} &={} \textcolor{black}{k_{1}\,x_{1}} \textcolor{black}{{}-k_{2}\,x_{2}}
\end{aligned}\right.\)
\end{minipage} \\
\end{tabular}\par\smallskip
\begin{tabularx}{\linewidth}{@{}l*{5}{>{\centering\arraybackslash}X}@{}}
\toprule
Method & OOD $R^2$ & Precision & Recall & F1 & TP/FP/FN \\
\midrule
SISR & -1.507 & 0.333 & 0.333 & 0.333 & 2/4/4 \\
KineSelect & 1.000 & 1.000 & 0.833 & 0.909 & 5/0/1 \\
\bottomrule
\end{tabularx}
\end{minipage} \\
\addlinespace[16pt]
\begin{minipage}[t]{\linewidth}
\centering
\textbf{(c) BIOMD0000000417 (Ratushny2012)}\par\smallskip
\begin{tabular}{@{}*{3}{>{\centering\arraybackslash}p{\dimexpr(\linewidth-16pt)/3\relax}}@{}}
\toprule
\textbf{Ground Truth} & \textbf{SISR} & \textbf{KineSelect} \\
\midrule
\begin{minipage}[t]{\linewidth}\vspace{0pt}\centering
\(\begin{aligned}
\dot{x}_{0} &={} \textcolor{black}{k_{1}} \textcolor{black}{{}-k_{2}\,x_{0}}
\end{aligned}\)
\end{minipage} &
\begin{minipage}[t]{\linewidth}\vspace{0pt}\centering
\(\begin{aligned}
\dot{x}_{0} &={} \textcolor{black}{k_{1}} \textcolor{black}{{}-k_{2}\,x_{0}}
\end{aligned}\)
\end{minipage} &
\begin{minipage}[t]{\linewidth}\vspace{0pt}\centering
\(\begin{aligned}
\dot{x}_{0} &={} \textcolor{black}{k_{1}} \textcolor{black}{{}-k_{2}\,x_{0}}
\end{aligned}\)
\end{minipage} \\
\end{tabular}\par\smallskip
\begin{tabularx}{\linewidth}{@{}l*{5}{>{\centering\arraybackslash}X}@{}}
\toprule
Method & OOD $R^2$ & Precision & Recall & F1 & TP/FP/FN \\
\midrule
SISR & 1.000 & 1.000 & 1.000 & 1.000 & 2/0/0 \\
KineSelect & 1.000 & 1.000 & 1.000 & 1.000 & 2/0/0 \\
\bottomrule
\end{tabularx}
\end{minipage} \\
\addlinespace[16pt]
\begin{minipage}[t]{\linewidth}
\centering
\textbf{(d) BIOMD0000000936 (Ferrel2011)}\par\smallskip
\begin{tabular}{@{}*{3}{>{\centering\arraybackslash}p{\dimexpr(\linewidth-16pt)/3\relax}}@{}}
\toprule
\textbf{Ground Truth} & \textbf{SISR} & \textbf{KineSelect} \\
\midrule
\begin{minipage}[t]{\linewidth}\vspace{0pt}\centering
\(\begin{aligned}
\dot{x}_{0} &={} \textcolor{black}{k_{1}} \textcolor{black}{{}-\frac{k_{2}\,{x_{0}}^{9}}{{K_{\mathrm{H},2}}^{8} + {x_{0}}^{8}}}
\end{aligned}\)
\end{minipage} &
\begin{minipage}[t]{\linewidth}\vspace{0pt}\centering
\(\begin{aligned}
\dot{x}_{0} &={} \textcolor{CfiveUnmatched}{-k_{1}\,{x_{0}}^{2}} \textcolor{CfiveUnmatched}{{}+k_{2}\,x_{0}} \textcolor{black}{{}+k_{3}}
\end{aligned}\)
\end{minipage} &
\begin{minipage}[t]{\linewidth}\vspace{0pt}\centering
\(\begin{aligned}
\dot{x}_{0} &={} \textcolor{black}{k_{1}} \textcolor{CfiveUnmatched}{{}+k_{2}\,x_{0}} \textcolor{CfiveUnmatched}{{}-2\,k_{3}\,{x_{0}}^{2}}
\end{aligned}\)
\end{minipage} \\
\end{tabular}\par\smallskip
\begin{tabularx}{\linewidth}{@{}l*{5}{>{\centering\arraybackslash}X}@{}}
\toprule
Method & OOD $R^2$ & Precision & Recall & F1 & TP/FP/FN \\
\midrule
SISR & 0.993 & 0.333 & 0.500 & 0.400 & 1/2/1 \\
KineSelect & 0.993 & 0.333 & 0.500 & 0.400 & 1/2/1 \\
\bottomrule
\end{tabularx}
\end{minipage} \\
\addlinespace[16pt]
\begin{minipage}[t]{\linewidth}
\centering
\textbf{(e) BIOMD0000000299 (Leloup1999)}\par\smallskip
\begin{tabular}{@{}*{3}{>{\centering\arraybackslash}p{\dimexpr(\linewidth-16pt)/3\relax}}@{}}
\toprule
\textbf{Ground Truth} & \textbf{SISR} & \textbf{KineSelect} \\
\midrule
\begin{minipage}[t]{\linewidth}\vspace{0pt}\centering
\(\left\{\begin{aligned}
\dot{x}_{0} &={} \textcolor{black}{k_{1}\,\frac{{K_{\mathrm{H},1}}^{4}}{{K_{\mathrm{H},1}}^{4} + {x_{2}}^{4}}} \\
&\phantom{{}={}} \textcolor{black}{{}-\frac{k_{2}\,x_{0}}{K_{\mathrm{M},2} + x_{0}}} \\[3pt]
\dot{x}_{1} &={} \textcolor{black}{k_{3}\,x_{0}} \textcolor{black}{{}+k_{4}\,x_{2}} \textcolor{black}{{}-k_{5}\,x_{1}} \\
&\phantom{{}={}} \textcolor{black}{{}-\frac{k_{6}\,x_{1}}{K_{\mathrm{M},6} + x_{1}}} \\[3pt]
\dot{x}_{2} &={} \textcolor{black}{k_{7}\,x_{1}} \textcolor{black}{{}-k_{8}\,x_{2}}
\end{aligned}\right.\)
\end{minipage} &
\begin{minipage}[t]{\linewidth}\vspace{0pt}\centering
\(\left\{\begin{aligned}
\dot{x}_{0} &={} \textcolor{CfiveUnmatched}{k_{1}\,x_{0}} \textcolor{CfiveUnmatched}{{}-k_{5}\,x_{0}\,x_{1}} \\
&\phantom{{}={}} \textcolor{CfiveUnmatched}{{}+k_{6}\,{x_{0}}^{2}} \\[3pt]
\dot{x}_{1} &={} \textcolor{CfiveUnmatched}{2\,k_{3}\,x_{0}\,x_{1}} \textcolor{black}{{}-k_{4}\,x_{1}} \\
&\phantom{{}={}} \textcolor{CfiveUnmatched}{{}+k_{6}\,{x_{0}}^{2}} \\[3pt]
\dot{x}_{2} &={} \textcolor{black}{-k_{2}\,x_{2}} \textcolor{black}{{}+2\,k_{4}\,x_{1}}
\end{aligned}\right.\)
\end{minipage} &
\begin{minipage}[t]{\linewidth}\vspace{0pt}\centering
\(\left\{\begin{aligned}
\dot{x}_{0} &={} \textcolor{CfiveUnmatched}{k_{3}\,\frac{x_{0}}{K_{\mathrm{H},3} + x_{0}}} \\
&\phantom{{}={}} \textcolor{black}{{}+k_{4}\,\frac{{K_{\mathrm{H},4}}^{2}}{{K_{\mathrm{H},4}}^{2} + {x_{2}}^{2}}} \\
&\phantom{=} \textcolor{CfiveUnmatched}{{}-\frac{k_{8}\,x_{0}}{K_{\mathrm{M},8} + x_{0}}\,\frac{x_{2}}{K_{\mathrm{H},8} + x_{2}}} \\
&\phantom{{}={}} \textcolor{CfiveUnmatched}{{}-k_{9}\,x_{0}\,x_{1}} \\[3pt]
\dot{x}_{1} &={} \textcolor{black}{k_{2}\,x_{0}} \textcolor{black}{{}-\frac{k_{6}\,x_{1}}{K_{\mathrm{M},6} + x_{1}}} \\
&\phantom{{}={}} \textcolor{CfiveUnmatched}{{}-k_{9}\,x_{0}\,x_{1}} \\[3pt]
\dot{x}_{2} &={} \textcolor{black}{k_{1}\,x_{1}} \textcolor{black}{{}-k_{5}\,x_{2}}
\end{aligned}\right.\)
\end{minipage} \\
\end{tabular}\par\smallskip
\begin{tabularx}{\linewidth}{@{}l*{5}{>{\centering\arraybackslash}X}@{}}
\toprule
Method & OOD $R^2$ & Precision & Recall & F1 & TP/FP/FN \\
\midrule
SISR & -1.502 & 0.375 & 0.375 & 0.375 & 3/5/5 \\
KineSelect & -0.407 & 0.556 & 0.625 & 0.588 & 5/4/3 \\
\bottomrule
\end{tabularx}
\end{minipage} \\
\addlinespace[16pt]
\begin{minipage}[t]{\linewidth}
\centering
\textbf{(f) BIOMD0000000573 (Aguilera2014)}\par\smallskip
\begin{tabular}{@{}*{3}{>{\centering\arraybackslash}p{\dimexpr(\linewidth-16pt)/3\relax}}@{}}
\toprule
\textbf{Ground Truth} & \textbf{SISR} & \textbf{KineSelect} \\
\midrule
\begin{minipage}[t]{\linewidth}\vspace{0pt}\centering
\(\left\{\begin{aligned}
\dot{x}_{0} &={} \textcolor{black}{k_{1}} \textcolor{black}{{}-k_{2}\,x_{0}} \\
&\phantom{{}={}} \textcolor{black}{{}+k_{3}\,\frac{x_{0}}{K_{\mathrm{H},3} + x_{0}}} \\
&\phantom{{}={}} \textcolor{black}{{}-k_{4}\,x_{0}\,x_{1}} \\[3pt]
\dot{x}_{1} &={} \textcolor{black}{k_{5}} \textcolor{black}{{}-k_{6}\,x_{1}} \textcolor{black}{{}-k_{7}\,x_{0}\,x_{1}}
\end{aligned}\right.\)
\end{minipage} &
\begin{minipage}[t]{\linewidth}\vspace{0pt}\centering
\(\left\{\begin{aligned}
\dot{x}_{0} &={} \textcolor{CfiveUnmatched}{k_{1}\,{x_{0}}^{2}} \textcolor{black}{{}-k_{3}\,x_{0}} \\[3pt]
\dot{x}_{1} &={} \textcolor{black}{-k_{2}\,x_{1}} \textcolor{black}{{}+k_{4}}
\end{aligned}\right.\)
\end{minipage} &
\begin{minipage}[t]{\linewidth}\vspace{0pt}\centering
\(\left\{\begin{aligned}
\dot{x}_{0} &={} \textcolor{CfiveUnmatched}{-\frac{2\,k_{4}\,{x_{0}}^{2}}{K_{\mathrm{M},4} + x_{0}}} \\[3pt]
\dot{x}_{1} &={} \textcolor{black}{k_{1}} \textcolor{CfiveUnmatched}{{}+k_{2}\,\frac{x_{1}}{K_{\mathrm{H},2} + x_{1}}} \\
&\phantom{{}={}} \textcolor{CfiveUnmatched}{{}-\frac{k_{3}\,x_{1}}{K_{\mathrm{M},3} + x_{1}}}
\end{aligned}\right.\)
\end{minipage} \\
\end{tabular}\par\smallskip
\begin{tabularx}{\linewidth}{@{}l*{5}{>{\centering\arraybackslash}X}@{}}
\toprule
Method & OOD $R^2$ & Precision & Recall & F1 & TP/FP/FN \\
\midrule
SISR & 0.495 & 0.750 & 0.429 & 0.545 & 3/1/4 \\
KineSelect & -0.152 & 0.250 & 0.143 & 0.182 & 1/3/6 \\
\bottomrule
\end{tabularx}
\end{minipage} \\
\end{longtable}
\endgroup

%% file: supplementary/clean_multirun_results.tex
\begin{revision}
\subsection{Robustness across Search Seeds}
\label{app:clean_multirun_results}

\paragraph{Experimental setup and aggregation.}
We report four clean runs per stochastic configuration: the historical result (seed 42) and three additional search runs using seeds 53, 25, and 81.
For the deterministic methods FFX, SINDy, SINDy-PI, and Reactive SINDy, we retain the historical result as a single run.
We first aggregate systems within each run, then report the mean and standard deviation across runs.

Trajectory metrics retain all 198 systems, with 95 Main, 66 Challenge, and 37 Stress systems.
Reaction precision, recall, and F1 each use system-level averages over the 161 Main + Challenge systems.

\paragraph{Clean results across methods.}
KineSelect has the highest mean Overall OOD \(R^2\) among the configurations in Table~\ref{tab:supp_clean_multirun_trajectory} and the highest mean Overall reaction F1 in Table~\ref{tab:supp_clean_multirun_reaction}.
Its OOD \(R^2\) is \(0.531_{\pm 0.025}\), compared with \(0.440_{\pm 0.015}\) for SISR.
The corresponding reaction F1 scores are \(0.454_{\pm 0.007}\) and \(0.387_{\pm 0.008}\).

\input{Table/supp_clean_multirun_trajectory}
\clearpage
\input{Table/supp_clean_multirun_reaction}

\paragraph{Comparison with shared FPMS.}
SISR uses \(\alpha\)-threshold selection with \(\alpha=1.05\), whereas KineSelect uses FPMS.
The SISR + FPMS configuration applies FPMS to the same saved SISR candidate archive within each run.
FPMS increases SISR's mean Overall reaction F1 from 0.387 to 0.430 and precision from 0.407 to 0.530 (Table~\ref{tab:supp_clean_multirun_reaction}).
Its mean Overall OOD \(R^2\) decreases from 0.440 to 0.386 (Table~\ref{tab:supp_clean_selector_trajectory}).

With FPMS in both methods, KineSelect retains higher mean Overall OOD \(R^2\) (0.531 versus 0.386) and reaction F1 (0.454 versus 0.430).
The F1 difference is larger on Challenge (0.302 versus 0.258) than on Main (0.559 versus 0.550).
KineSelect has higher recall and lower precision than SISR + FPMS on both subsets (Table~\ref{tab:supp_clean_multirun_reaction}).
KineSelect maintains higher mean OOD \(R^2\) and reaction F1 than SISR + FPMS under the same final selection rule.

\input{Table/supp_clean_selector_trajectory}
\end{revision}

%% file: Table/supp_clean_multirun_trajectory.tex
\begin{table}[htbp]
  \begin{revision}
  \centering
  \caption{\rev{\textbf{Clean trajectory fidelity across runs.} SISR + FPMS replaces \(\alpha\)-threshold selection with FPMS while reusing the same SISR candidate archive within each run. Acc. denotes \(\mathrm{Acc}@.90\). Subscripts give standard deviation for stochastic methods. For deterministic methods (\(\dagger\)), we display \(\pm0.000\) by convention. Bold and underlining mark the highest and second-highest means.}}
  \label{tab:supp_clean_multirun_trajectory}
  \begingroup
  \scriptsize
  \setlength{\tabcolsep}{0.4pt}
  \renewcommand{\arraystretch}{1.18}
  \newcolumntype{Y}{>{\centering\arraybackslash}X}
  \begin{tabularx}{\linewidth}{@{}l@{\hspace{2pt}}*{10}{Y}@{}}
    \toprule
    \multirow{3}{*}{\textbf{Method}} & \multicolumn{4}{c}{\textbf{Overall}} & \multicolumn{2}{c}{\textbf{Main}} & \multicolumn{2}{c}{\textbf{Challenge}} & \multicolumn{2}{c}{\textbf{Stress}} \\
    \cmidrule(lr){2-5}\cmidrule(lr){6-7}\cmidrule(lr){8-9}\cmidrule(lr){10-11}
    & \multicolumn{2}{c}{\(R^2\)} & \multicolumn{2}{c}{Acc.} & \multicolumn{2}{c}{\(R^2\)} & \multicolumn{2}{c}{\(R^2\)} & \multicolumn{2}{c}{\(R^2\)} \\
    & ID & OOD & ID & OOD & ID & OOD & ID & OOD & ID & OOD \\
    \midrule
    EPLEX (default) & \(\text{0.027}_{\text{\(\pm\)0.012}}\) & \(\text{0.014}_{\text{\(\pm\)0.006}}\) & \(\text{0.018}_{\text{\(\pm\)0.011}}\) & \(\text{0.008}_{\text{\(\pm\)0.005}}\) & \(\text{0.033}_{\text{\(\pm\)0.015}}\) & \(\text{0.015}_{\text{\(\pm\)0.010}}\) & \(\text{0.024}_{\text{\(\pm\)0.007}}\) & \(\text{0.012}_{\text{\(\pm\)0.015}}\) & \(\text{0.020}_{\text{\(\pm\)0.026}}\) & \(\text{0.012}_{\text{\(\pm\)0.024}}\) \\
    EPLEX (poly) & \(\text{0.109}_{\text{\(\pm\)0.022}}\) & \(\text{0.054}_{\text{\(\pm\)0.008}}\) & \(\text{0.074}_{\text{\(\pm\)0.030}}\) & \(\text{0.034}_{\text{\(\pm\)0.008}}\) & \(\text{0.134}_{\text{\(\pm\)0.035}}\) & \(\text{0.055}_{\text{\(\pm\)0.007}}\) & \(\text{0.090}_{\text{\(\pm\)0.020}}\) & \(\text{0.069}_{\text{\(\pm\)0.010}}\) & \(\text{0.078}_{\text{\(\pm\)0.044}}\) & \(\text{0.026}_{\text{\(\pm\)0.022}}\) \\
    FE-AFP & \(\text{0.120}_{\text{\(\pm\)0.013}}\) & \(\text{0.081}_{\text{\(\pm\)0.012}}\) & \(\text{0.096}_{\text{\(\pm\)0.015}}\) & \(\text{0.052}_{\text{\(\pm\)0.018}}\) & \(\text{0.142}_{\text{\(\pm\)0.030}}\) & \(\text{0.083}_{\text{\(\pm\)0.027}}\) & \(\text{0.120}_{\text{\(\pm\)0.017}}\) & \(\text{0.098}_{\text{\(\pm\)0.019}}\) & \(\text{0.066}_{\text{\(\pm\)0.027}}\) & \(\text{0.045}_{\text{\(\pm\)0.023}}\) \\
    FFX\({}^{\dagger}\) & \(\text{0.658}_{\text{\(\pm\)0.000}}\) & \(\text{0.367}_{\text{\(\pm\)0.000}}\) & \(\text{0.525}_{\text{\(\pm\)0.000}}\) & \(\text{0.222}_{\text{\(\pm\)0.000}}\) & \(\text{0.755}_{\text{\(\pm\)0.000}}\) & \(\text{0.402}_{\text{\(\pm\)0.000}}\) & \(\underline{\text{0.592}}_{\text{\(\pm\)0.000}}\) & \(\underline{\text{0.364}}_{\text{\(\pm\)0.000}}\) & \(\text{0.529}_{\text{\(\pm\)0.000}}\) & \(\text{0.280}_{\text{\(\pm\)0.000}}\) \\
    PySR & \(\text{0.632}_{\text{\(\pm\)0.194}}\) & \(\text{0.291}_{\text{\(\pm\)0.041}}\) & \(\text{0.552}_{\text{\(\pm\)0.166}}\) & \(\text{0.221}_{\text{\(\pm\)0.032}}\) & \(\text{0.747}_{\text{\(\pm\)0.175}}\) & \(\text{0.380}_{\text{\(\pm\)0.033}}\) & \(\text{0.551}_{\text{\(\pm\)0.214}}\) & \(\text{0.250}_{\text{\(\pm\)0.059}}\) & \(\text{0.482}_{\text{\(\pm\)0.222}}\) & \(\text{0.136}_{\text{\(\pm\)0.053}}\) \\
    \midrule
    SINDy\({}^{\dagger}\) & \(\text{0.038}_{\text{\(\pm\)0.000}}\) & \(\text{0.033}_{\text{\(\pm\)0.000}}\) & \(\text{0.035}_{\text{\(\pm\)0.000}}\) & \(\text{0.030}_{\text{\(\pm\)0.000}}\) & \(\text{0.035}_{\text{\(\pm\)0.000}}\) & \(\text{0.018}_{\text{\(\pm\)0.000}}\) & \(\text{0.063}_{\text{\(\pm\)0.000}}\) & \(\text{0.075}_{\text{\(\pm\)0.000}}\) & \(\text{0.000}_{\text{\(\pm\)0.000}}\) & \(\text{0.000}_{\text{\(\pm\)0.000}}\) \\
    SINDy-PI\({}^{\dagger}\) & \(\text{0.391}_{\text{\(\pm\)0.000}}\) & \(\text{0.310}_{\text{\(\pm\)0.000}}\) & \(\text{0.379}_{\text{\(\pm\)0.000}}\) & \(\text{0.273}_{\text{\(\pm\)0.000}}\) & \(\text{0.491}_{\text{\(\pm\)0.000}}\) & \(\text{0.393}_{\text{\(\pm\)0.000}}\) & \(\text{0.333}_{\text{\(\pm\)0.000}}\) & \(\text{0.261}_{\text{\(\pm\)0.000}}\) & \(\text{0.241}_{\text{\(\pm\)0.000}}\) & \(\text{0.182}_{\text{\(\pm\)0.000}}\) \\
    Reactive SINDy\({}^{\dagger}\) & \(\text{0.166}_{\text{\(\pm\)0.000}}\) & \(\text{0.142}_{\text{\(\pm\)0.000}}\) & \(\text{0.101}_{\text{\(\pm\)0.000}}\) & \(\text{0.091}_{\text{\(\pm\)0.000}}\) & \(\text{0.246}_{\text{\(\pm\)0.000}}\) & \(\text{0.220}_{\text{\(\pm\)0.000}}\) & \(\text{0.094}_{\text{\(\pm\)0.000}}\) & \(\text{0.068}_{\text{\(\pm\)0.000}}\) & \(\text{0.089}_{\text{\(\pm\)0.000}}\) & \(\text{0.074}_{\text{\(\pm\)0.000}}\) \\
    \midrule
    ODEFormer & \(\text{0.239}_{\text{\(\pm\)0.017}}\) & \(\text{0.146}_{\text{\(\pm\)0.007}}\) & \(\text{0.125}_{\text{\(\pm\)0.011}}\) & \(\text{0.069}_{\text{\(\pm\)0.006}}\) & \(\text{0.288}_{\text{\(\pm\)0.018}}\) & \(\text{0.163}_{\text{\(\pm\)0.010}}\) & \(\text{0.205}_{\text{\(\pm\)0.028}}\) & \(\text{0.137}_{\text{\(\pm\)0.013}}\) & \(\text{0.175}_{\text{\(\pm\)0.015}}\) & \(\text{0.119}_{\text{\(\pm\)0.016}}\) \\
    LLM-SR & \(\text{0.344}_{\text{\(\pm\)0.023}}\) & \(\text{0.224}_{\text{\(\pm\)0.010}}\) & \(\text{0.322}_{\text{\(\pm\)0.024}}\) & \(\text{0.182}_{\text{\(\pm\)0.007}}\) & \(\text{0.419}_{\text{\(\pm\)0.014}}\) & \(\text{0.283}_{\text{\(\pm\)0.012}}\) & \(\text{0.300}_{\text{\(\pm\)0.060}}\) & \(\text{0.184}_{\text{\(\pm\)0.043}}\) & \(\text{0.230}_{\text{\(\pm\)0.037}}\) & \(\text{0.143}_{\text{\(\pm\)0.037}}\) \\
    \midrule
    SISR & \(\underline{\text{0.681}}_{\text{\(\pm\)0.014}}\) & \(\underline{\text{0.440}}_{\text{\(\pm\)0.015}}\) & \(\underline{\text{0.571}}_{\text{\(\pm\)0.007}}\) & \(\underline{\text{0.337}}_{\text{\(\pm\)0.014}}\) & \(\underline{\text{0.770}}_{\text{\(\pm\)0.023}}\) & \(\underline{\text{0.535}}_{\text{\(\pm\)0.029}}\) & \(\text{0.587}_{\text{\(\pm\)0.006}}\) & \(\text{0.342}_{\text{\(\pm\)0.004}}\) & \(\underline{\text{0.620}}_{\text{\(\pm\)0.024}}\) & \(\underline{\text{0.372}}_{\text{\(\pm\)0.019}}\) \\
    SISR + FPMS & \(\text{0.500}_{\text{\(\pm\)0.083}}\) & \(\text{0.386}_{\text{\(\pm\)0.037}}\) & \(\text{0.443}_{\text{\(\pm\)0.065}}\) & \(\text{0.306}_{\text{\(\pm\)0.019}}\) & \(\text{0.623}_{\text{\(\pm\)0.060}}\) & \(\text{0.499}_{\text{\(\pm\)0.027}}\) & \(\text{0.391}_{\text{\(\pm\)0.115}}\) & \(\text{0.280}_{\text{\(\pm\)0.047}}\) & \(\text{0.381}_{\text{\(\pm\)0.093}}\) & \(\text{0.284}_{\text{\(\pm\)0.050}}\) \\
    KineSelect & \(\text{\bfseries 0.765}_{\text{\(\pm\)0.012}}\) & \(\text{\bfseries 0.531}_{\text{\(\pm\)0.025}}\) & \(\text{\bfseries 0.663}_{\text{\(\pm\)0.012}}\) & \(\text{\bfseries 0.414}_{\text{\(\pm\)0.029}}\) & \(\text{\bfseries 0.842}_{\text{\(\pm\)0.021}}\) & \(\text{\bfseries 0.621}_{\text{\(\pm\)0.026}}\) & \(\text{\bfseries 0.720}_{\text{\(\pm\)0.009}}\) & \(\text{\bfseries 0.468}_{\text{\(\pm\)0.025}}\) & \(\text{\bfseries 0.649}_{\text{\(\pm\)0.013}}\) & \(\text{\bfseries 0.411}_{\text{\(\pm\)0.025}}\) \\
    \bottomrule
  \end{tabularx}
  \par
  \endgroup
  \end{revision}
\end{table}

%% file: Table/supp_clean_multirun_reaction.tex
\begin{table}[htbp]
  \begin{revision}
  \centering
  \caption{\rev{\textbf{Reaction recovery from clean trajectory data.} Overall, Main, and Challenge contain 161, 95, and 66 systems, respectively. Subscripts give standard deviation for stochastic methods. For deterministic methods (\(\dagger\)), we display \(\pm0.000\) by convention. Bold and underlining mark the highest and second-highest means.}}
  \label{tab:supp_clean_multirun_reaction}
  \begingroup
  \scriptsize
  \setlength{\tabcolsep}{0.4pt}
  \renewcommand{\arraystretch}{1.18}
  \newcolumntype{Y}{>{\centering\arraybackslash}X}
  \begin{tabularx}{\linewidth}{@{}l@{\hspace{2pt}}*{9}{Y}@{}}
    \toprule
    \multirow{2}{*}{\textbf{Method}} & \multicolumn{3}{c}{\textbf{Overall}} & \multicolumn{3}{c}{\textbf{Main}} & \multicolumn{3}{c}{\textbf{Challenge}} \\
    \cmidrule(lr){2-4}\cmidrule(lr){5-7}\cmidrule(lr){8-10}
    & Precision & Recall & F1 & Precision & Recall & F1 & Precision & Recall & F1 \\
    \midrule
    Reactive SINDy\({}^{\dagger}\) & \(\text{0.320}_{\text{\(\pm\)0.000}}\) & \(\text{0.370}_{\text{\(\pm\)0.000}}\) & \(\text{0.292}_{\text{\(\pm\)0.000}}\) & \(\text{0.383}_{\text{\(\pm\)0.000}}\) & \(\text{0.477}_{\text{\(\pm\)0.000}}\) & \(\text{0.363}_{\text{\(\pm\)0.000}}\) & \(\text{0.228}_{\text{\(\pm\)0.000}}\) & \(\text{0.215}_{\text{\(\pm\)0.000}}\) & \(\text{0.191}_{\text{\(\pm\)0.000}}\) \\
    \midrule
    SISR & \(\text{0.407}_{\text{\(\pm\)0.010}}\) & \(\underline{\text{0.417}}_{\text{\(\pm\)0.006}}\) & \(\text{0.387}_{\text{\(\pm\)0.008}}\) & \(\text{0.504}_{\text{\(\pm\)0.017}}\) & \(\underline{\text{0.540}}_{\text{\(\pm\)0.005}}\) & \(\text{0.494}_{\text{\(\pm\)0.010}}\) & \(\text{0.269}_{\text{\(\pm\)0.027}}\) & \(\text{0.239}_{\text{\(\pm\)0.010}}\) & \(\text{0.234}_{\text{\(\pm\)0.012}}\) \\
    SISR + FPMS & \(\text{\bfseries 0.530}_{\text{\(\pm\)0.012}}\) & \(\text{0.408}_{\text{\(\pm\)0.006}}\) & \(\underline{\text{0.430}}_{\text{\(\pm\)0.006}}\) & \(\text{\bfseries 0.654}_{\text{\(\pm\)0.011}}\) & \(\text{0.524}_{\text{\(\pm\)0.008}}\) & \(\underline{\text{0.550}}_{\text{\(\pm\)0.007}}\) & \(\text{\bfseries 0.351}_{\text{\(\pm\)0.022}}\) & \(\underline{\text{0.241}}_{\text{\(\pm\)0.014}}\) & \(\underline{\text{0.258}}_{\text{\(\pm\)0.016}}\) \\
    KineSelect & \(\underline{\text{0.448}}_{\text{\(\pm\)0.010}}\) & \(\text{\bfseries 0.489}_{\text{\(\pm\)0.007}}\) & \(\text{\bfseries 0.454}_{\text{\(\pm\)0.007}}\) & \(\underline{\text{0.546}}_{\text{\(\pm\)0.005}}\) & \(\text{\bfseries 0.604}_{\text{\(\pm\)0.008}}\) & \(\text{\bfseries 0.559}_{\text{\(\pm\)0.005}}\) & \(\underline{\text{0.307}}_{\text{\(\pm\)0.022}}\) & \(\text{\bfseries 0.323}_{\text{\(\pm\)0.016}}\) & \(\text{\bfseries 0.302}_{\text{\(\pm\)0.017}}\) \\
    \bottomrule
  \end{tabularx}
  \par
  \endgroup
  \end{revision}
\end{table}

%% file: Table/supp_clean_selector_trajectory.tex
\begin{table}[htbp]
  \begin{revision}
  \centering
  \caption{\rev{\textbf{OOD trajectory fidelity with shared FPMS.} Overall, Main, Challenge, and Stress contain 198, 95, 66, and 37 systems, respectively. Acc. denotes \(\mathrm{Acc}@.90\). Subscripts give standard deviation for stochastic methods. Bold and underlining mark the highest and second-highest means.}}
  \label{tab:supp_clean_selector_trajectory}
  \begingroup
  \footnotesize
  \setlength{\tabcolsep}{1.2pt}
  \renewcommand{\arraystretch}{1.18}
  \newcolumntype{Y}{>{\centering\arraybackslash}X}
  \begin{tabularx}{\linewidth}{@{}l@{\hspace{2pt}}*{8}{Y}@{}}
    \toprule
    \multirow{2}{*}{\textbf{Method}} & \multicolumn{2}{c}{\textbf{Overall}} & \multicolumn{2}{c}{\textbf{Main}} & \multicolumn{2}{c}{\textbf{Challenge}} & \multicolumn{2}{c}{\textbf{Stress}} \\
    \cmidrule(lr){2-3}\cmidrule(lr){4-5}\cmidrule(lr){6-7}\cmidrule(lr){8-9}
    & \(R^2\) & Acc. & \(R^2\) & Acc. & \(R^2\) & Acc. & \(R^2\) & Acc. \\
    \midrule
    SISR & \(\underline{\text{0.440}}_{\text{\(\pm\)0.015}}\) & \(\underline{\text{0.337}}_{\text{\(\pm\)0.014}}\) & \(\underline{\text{0.535}}_{\text{\(\pm\)0.029}}\) & \(\underline{\text{0.445}}_{\text{\(\pm\)0.034}}\) & \(\underline{\text{0.342}}_{\text{\(\pm\)0.004}}\) & \(\underline{\text{0.239}}_{\text{\(\pm\)0.008}}\) & \(\underline{\text{0.372}}_{\text{\(\pm\)0.019}}\) & \(\underline{\text{0.236}}_{\text{\(\pm\)0.034}}\) \\
    SISR + FPMS & \(\text{0.386}_{\text{\(\pm\)0.037}}\) & \(\text{0.306}_{\text{\(\pm\)0.019}}\) & \(\text{0.499}_{\text{\(\pm\)0.027}}\) & \(\text{0.421}_{\text{\(\pm\)0.017}}\) & \(\text{0.280}_{\text{\(\pm\)0.047}}\) & \(\text{0.205}_{\text{\(\pm\)0.026}}\) & \(\text{0.284}_{\text{\(\pm\)0.050}}\) & \(\text{0.189}_{\text{\(\pm\)0.022}}\) \\
    KineSelect & \(\text{\bfseries 0.531}_{\text{\(\pm\)0.025}}\) & \(\text{\bfseries 0.414}_{\text{\(\pm\)0.029}}\) & \(\text{\bfseries 0.621}_{\text{\(\pm\)0.026}}\) & \(\text{\bfseries 0.553}_{\text{\(\pm\)0.028}}\) & \(\text{\bfseries 0.468}_{\text{\(\pm\)0.025}}\) & \(\text{\bfseries 0.295}_{\text{\(\pm\)0.038}}\) & \(\text{\bfseries 0.411}_{\text{\(\pm\)0.025}}\) & \(\text{\bfseries 0.270}_{\text{\(\pm\)0.022}}\) \\
    \bottomrule
  \end{tabularx}
  \par
  \endgroup
  \end{revision}
\end{table}

%% file: supplementary/multi_ood_results.tex
\begin{revision}
\subsection{OOD Evaluation across Multiple Initial Conditions}
\label{app:multi_ood_results}

\paragraph{Frozen models and initial conditions.}
We evaluate frozen SISR, SISR + FPMS, and KineSelect models from clean random seeds 53, 25, and 81 on 20 new initial conditions per system.
Each model uses only the default training trajectory.
We sample each initial state coordinate uniformly between its minimum and maximum along that default trajectory.
For each system, we randomly sampled initial conditions from the coordinate-wise min–max range of its default training trajectory and retained 20 with complete, finite reference trajectories on the original time grid. The resulting OOD-trajectories set is shared across methods and random seeds.
Reference integration uses LSODA (rtol \(10^{-8}\), atol \(10^{-10}\)) with a 300-second timeout.

\paragraph{Scoring and aggregation.}
We calculate \(R^2\) for each trajectory, truncate negative scores to zero, and average first over the 20 initial conditions within each system, then equally over systems.
We report the mean and standard deviation of the resulting benchmark scores across the three random seeds.
For Acc@.90, we compute the fraction of trajectories with $R^2 \geq 0.90$, averaging first over initial conditions within each system and then equally across systems.

\paragraph{Results across initial conditions.}
KineSelect retains the highest mean \(R^2\) and \(\mathrm{Acc}@.90\) in both OOD protocols and each multi-IC benchmark subset (Tables~\ref{tab:supp_multi_ood_comparison} and~\ref{tab:supp_multi_ood_subsets}).
KineSelect's mean OOD \(R^2\) remains nearly unchanged between the single-IC and multi-IC evaluations (0.538 in both), while \(\mathrm{Acc}@.90\) increases from 0.428 to 0.435.
These results support its overall advantage in trajectory fidelity when evaluation covers 20 new initial conditions per system.

\input{Table/supp_multi_ood_comparison}
\input{Table/supp_multi_ood_subsets}

\paragraph{Paired comparisons.}
The mean multi-IC OOD \(R^2\) difference between KineSelect and SISR is 0.078, with a 95\% bootstrap interval of [0.036, 0.123].
With FPMS shared, the difference between KineSelect and SISR + FPMS is 0.171 [0.122, 0.221].
We obtain these intervals from 2,000 paired bootstrap resamples of whole systems.
These intervals describe system-level variation, whereas the table subscripts describe variation across random seeds.
\end{revision}

%% file: Table/supp_multi_ood_comparison.tex
\begin{table}[!htbp]
  \begin{revision}
  \centering
  \caption{\rev{\textbf{Single- and multiple-initial-condition OOD evaluation after clean fitting.} Both protocols evaluate the same frozen models over 198 systems. Single IC uses the original perturbed trajectory, and Multi-IC uses 20 new initial conditions per system. Acc. denotes \(\mathrm{Acc}@.90\). Subscripts give standard deviation across random seeds. Bold and underlining mark the highest and second-highest means.}}
  \label{tab:supp_multi_ood_comparison}
  \begingroup
  \footnotesize
  \setlength{\tabcolsep}{2pt}
  \renewcommand{\arraystretch}{1.18}
  \newcolumntype{Y}{>{\centering\arraybackslash}X}
  \begin{tabularx}{\linewidth}{@{}l@{\hspace{5pt}}*{4}{Y}@{}}
    \toprule
    \multirow{2}{*}{\textbf{Method}} & \multicolumn{2}{c}{\textbf{Single IC}} & \multicolumn{2}{c}{\textbf{Multi-IC (20)}} \\
    \cmidrule(lr){2-3}\cmidrule(lr){4-5}
    & \(R^2\) & Acc. & \(R^2\) & Acc. \\
    \midrule
    SISR & \(\underline{\text{0.445}}_{\text{\(\pm\)0.014}}\) & \(\underline{\text{0.340}}_{\text{\(\pm\)0.015}}\) & \(\underline{\text{0.461}}_{\text{\(\pm\)0.007}}\) & \(\underline{\text{0.361}}_{\text{\(\pm\)0.005}}\) \\
    SISR + FPMS & \(\text{0.368}_{\text{\(\pm\)0.009}}\) & \(\text{0.298}_{\text{\(\pm\)0.013}}\) & \(\text{0.367}_{\text{\(\pm\)0.008}}\) & \(\text{0.313}_{\text{\(\pm\)0.010}}\) \\
    KineSelect & \(\text{\bfseries 0.538}_{\text{\(\pm\)0.024}}\) & \(\text{\bfseries 0.428}_{\text{\(\pm\)0.013}}\) & \(\text{\bfseries 0.538}_{\text{\(\pm\)0.011}}\) & \(\text{\bfseries 0.435}_{\text{\(\pm\)0.008}}\) \\
    \bottomrule
  \end{tabularx}
  \par
  \endgroup
  \end{revision}
\end{table}

%% file: Table/supp_multi_ood_subsets.tex
\begin{table}[!htbp]
  \begin{revision}
  \centering
  \caption{\rev{\textbf{Multi-IC OOD trajectory fidelity by benchmark subset.} Main, Challenge, and Stress contain 95, 66, and 37 systems, respectively, with 20 initial conditions per system. Acc. denotes \(\mathrm{Acc}@.90\). Subscripts give standard deviation across random seeds. Bold and underlining mark the highest and second-highest means.}}
  \label{tab:supp_multi_ood_subsets}
  \begingroup
  \footnotesize
  \setlength{\tabcolsep}{2pt}
  \renewcommand{\arraystretch}{1.18}
  \newcolumntype{Y}{>{\centering\arraybackslash}X}
  \begin{tabularx}{\linewidth}{@{}l@{\hspace{5pt}}*{6}{Y}@{}}
    \toprule
    \multirow{2}{*}{\textbf{Method}} & \multicolumn{2}{c}{\textbf{Main}} & \multicolumn{2}{c}{\textbf{Challenge}} & \multicolumn{2}{c}{\textbf{Stress}} \\
    \cmidrule(lr){2-3}\cmidrule(lr){4-5}\cmidrule(lr){6-7}
    & \(R^2\) & Acc. & \(R^2\) & Acc. & \(R^2\) & Acc. \\
    \midrule
    SISR & \(\underline{\text{0.604}}_{\text{\(\pm\)0.018}}\) & \(\underline{\text{0.512}}_{\text{\(\pm\)0.012}}\) & \(\underline{\text{0.329}}_{\text{\(\pm\)0.011}}\) & \(\underline{\text{0.234}}_{\text{\(\pm\)0.003}}\) & \(\underline{\text{0.328}}_{\text{\(\pm\)0.015}}\) & \(\underline{\text{0.203}}_{\text{\(\pm\)0.011}}\) \\
    SISR + FPMS & \(\text{0.514}_{\text{\(\pm\)0.011}}\) & \(\text{0.455}_{\text{\(\pm\)0.020}}\) & \(\text{0.239}_{\text{\(\pm\)0.015}}\) & \(\text{0.194}_{\text{\(\pm\)0.011}}\) & \(\text{0.219}_{\text{\(\pm\)0.032}}\) & \(\text{0.159}_{\text{\(\pm\)0.014}}\) \\
    KineSelect & \(\text{\bfseries 0.658}_{\text{\(\pm\)0.001}}\) & \(\text{\bfseries 0.573}_{\text{\(\pm\)0.002}}\) & \(\text{\bfseries 0.454}_{\text{\(\pm\)0.016}}\) & \(\text{\bfseries 0.337}_{\text{\(\pm\)0.012}}\) & \(\text{\bfseries 0.382}_{\text{\(\pm\)0.028}}\) & \(\text{\bfseries 0.254}_{\text{\(\pm\)0.025}}\) \\
    \bottomrule
  \end{tabularx}
  \par
  \endgroup
  \end{revision}
\end{table}

%% file: supplementary/fpms_ablation_across_seeds.tex
\begin{revision}
\subsection{FPMS Component Ablations across Search Seeds}
\label{app:fpms_ablation_across_seeds}

For each system and search seed, we apply the four FPMS settings in Table~\ref{tab:supp_fpms_ablation} to the same set of models fitted to clean data.
Both tables report means across four random seeds. We used the original search implementation for seed 42 and the updated implementation for seeds 53, 25, and 81.
As in the original ablation, Full FPMS achieved higher reaction F1 than Base for all four seeds (Table~\ref{tab:supp_fpms_ablation_across_seeds}).

\input{Table/supp_fpms_ablation_across_seeds}

On Challenge, adding \(C^{\mathrm{spec}}\) raised mean OOD \(R^2\) in both initial-condition tests and lowered mean reaction F1 for MM and Hill kinetics (Table~\ref{tab:supp_fpms_challenge_cspec}).

\input{Table/supp_fpms_challenge_cspec}
\end{revision}

%% file: Table/supp_fpms_ablation_across_seeds.tex
\begin{table}[!htbp]
  \begin{revision}
  \centering
  \caption{\rev{\textbf{FPMS component ablations on clean data.} Base uses simulation loss and expression-tree complexity. Full FPMS adds derivative loss \(\mathcal{L}^{\mathrm{der}}\) and the number of non-MA kinetic terms \(C^{\mathrm{spec}}\).}}
  \label{tab:supp_fpms_ablation_across_seeds}
  \begingroup
  \footnotesize
  \setlength{\tabcolsep}{0.8pt}
  \renewcommand{\arraystretch}{1.18}
  \begin{tabularx}{\linewidth}{@{}l*{6}{>{\centering\arraybackslash}X}@{}}
    \toprule
    \multirow{2}{*}{\textbf{Selection}} & \multicolumn{3}{c}{\textbf{OOD \(R^2\)}} & \multicolumn{3}{c}{\textbf{Reaction F1}} \\
    \cmidrule(lr){2-4}\cmidrule(lr){5-7}
     & Overall & Main & Challenge & Overall & Main & Challenge \\
    \midrule
    Base & \(\text{0.5320}_{\text{\(\pm\)0.0222}}\) & \(\text{0.6195}_{\text{\(\pm\)0.0154}}\) & \(\text{0.4511}_{\text{\(\pm\)0.0341}}\) & \(\text{0.4318}_{\text{\(\pm\)0.0054}}\) & \(\text{0.5319}_{\text{\(\pm\)0.0095}}\) & \(\text{0.2876}_{\text{\(\pm\)0.0103}}\) \\
    Base \(+\mathcal{L}^{\mathrm{der}}\) & \(\text{0.5299}_{\text{\(\pm\)0.0208}}\) & \(\text{0.6197}_{\text{\(\pm\)0.0266}}\) & \(\text{0.4479}_{\text{\(\pm\)0.0209}}\) & \(\text{0.4437}_{\text{\(\pm\)0.0052}}\) & \(\text{0.5481}_{\text{\(\pm\)0.0034}}\) & \(\text{0.2934}_{\text{\(\pm\)0.0116}}\) \\
    Base \(+C^{\mathrm{spec}}\) & \(\text{\bfseries 0.5384}_{\text{\(\pm\)0.0249}}\) & \(\text{\bfseries 0.6276}_{\text{\(\pm\)0.0188}}\) & \(\text{0.4631}_{\text{\(\pm\)0.0397}}\) & \(\text{0.4415}_{\text{\(\pm\)0.0045}}\) & \(\text{0.5451}_{\text{\(\pm\)0.0050}}\) & \(\text{0.2924}_{\text{\(\pm\)0.0062}}\) \\
    Full FPMS & \(\text{0.5308}_{\text{\(\pm\)0.0245}}\) & \(\text{0.6208}_{\text{\(\pm\)0.0258}}\) & \(\text{\bfseries 0.4684}_{\text{\(\pm\)0.0248}}\) & \(\text{\bfseries 0.4536}_{\text{\(\pm\)0.0065}}\) & \(\text{\bfseries 0.5590}_{\text{\(\pm\)0.0052}}\) & \(\text{\bfseries 0.3019}_{\text{\(\pm\)0.0169}}\) \\
    \bottomrule
  \end{tabularx}
  \par
  \endgroup
  \end{revision}
\end{table}

%% file: Table/supp_fpms_challenge_cspec.tex
\begin{table}[!htbp]
  \begin{revision}
  \centering
  \caption{\rev{\textbf{Effect of \(C^{\mathrm{spec}}\) on the Challenge subset.}}}
  \label{tab:supp_fpms_challenge_cspec}
  \begingroup
  \footnotesize
  \setlength{\tabcolsep}{0.8pt}
  \renewcommand{\arraystretch}{1.18}
  \begin{tabularx}{\linewidth}{@{}l*{4}{>{\centering\arraybackslash}X}@{}}
    \toprule
    \multirow{2}{*}{\(C^{\mathrm{spec}}\)} & \multicolumn{2}{c}{\textbf{Reaction F1}} & \multicolumn{2}{c}{\textbf{OOD \(R^2\)}} \\
    \cmidrule(lr){2-3}\cmidrule(lr){4-5}
     & MM & Hill & Single IC & Multi-IC (20) \\
    \midrule
    Off & \(\text{\bfseries 0.1174}_{\text{\(\pm\)0.0060}}\) & \(\text{\bfseries 0.0533}_{\text{\(\pm\)0.0097}}\) & \(\text{0.4479}_{\text{\(\pm\)0.0209}}\) & \(\text{0.4333}_{\text{\(\pm\)0.0160}}\) \\
    On & \(\text{0.1113}_{\text{\(\pm\)0.0079}}\) & \(\text{0.0477}_{\text{\(\pm\)0.0104}}\) & \(\text{\bfseries 0.4684}_{\text{\(\pm\)0.0248}}\) & \(\text{\bfseries 0.4435}_{\text{\(\pm\)0.0246}}\) \\
    \bottomrule
  \end{tabularx}
  \par
  \endgroup
  \end{revision}
\end{table}

%% file: supplementary/reproducibility_details.tex
\clearpage
\section{Reproducibility and Implementation Details}
\label{app:reproducibility}

We apply each method to the trajectory-to-ODE task through a shared data interface and method-specific fitting routines.
This appendix specifies the inputs, search spaces, fitting budgets, and final-model selection used in the experiments.

\subsection{Common Experimental Pipeline}
\label{app:common_pipeline}

\paragraph{Trajectory inputs and derivative targets.}
For each system, we fit the default trajectory using its recorded observation times and state-variable order.
We follow the preprocessing routines of the adopted implementations; Tables~\ref{tab:supp_baseline_search} and \ref{tab:supp_baseline_other} list the method-specific settings.
Derivative-based methods estimate state-derivative targets from the fitting observations, whereas ODEFormer receives the trajectory directly.
In the noise experiments, these methods re-estimate derivatives from the corrupted fitting trajectory.

\paragraph{Coupled-system adaptation.}
For scalar regressors, we fit each state equation using the full state vector and assemble the selected state equations into a coupled ODE.
LLM-SR also receives time as an input.
Reaction-based methods fit a mechanism whose stoichiometry couples the state equations.

\paragraph{Candidate selection and scoring.}
Internal parameter selection and candidate ranking use the fitting trajectory, with method-specific criteria described below.
SISR and KineSelect compute their internal simulation losses against the fitting observations.
For final scoring, each method's prediction routine integrates the recovered coupled ODE in the benchmark's original state coordinates and time units.
SISR and KineSelect use LSODA~\citep{petzold1983automatic} with relative tolerance \(10^{-6}\) and absolute tolerance \(10^{-9}\).
LLM-SR uses the prediction settings in Table~\ref{tab:supp_baseline_other}.
We compute trajectory metrics from the saved predictions.
ID reconstruction and OOD extrapolation use the initial conditions and time grids in Appendix~\ref{app:id_ood_noise}.
We sum expression complexity across state equations (Section~\ref{sec:evaluation}) and score reaction outputs from Reactive SINDy, SISR, and KineSelect using Appendix~\ref{app:reaction_matching}.
Failure handling and score aggregation follow Appendix~\ref{app:metric_contract}.

\subsection{Baseline Implementations and Configurations}
\label{app:method_configurations}

EPLEX, FE-AFP, FFX, and SINDy use baseline implementations adapted from the ODEFormer codebase~\citep{d2023odeformer}.
The remaining methods use the packages or Python implementations described below.
We follow the adopted implementations' default fitting procedures, including their built-in parameter selection where enabled.
Tables~\ref{tab:supp_baseline_search} and \ref{tab:supp_baseline_other} list the search spaces and budgets for the eleven baseline configurations.
The descriptions below explain their implementation and task adaptations.

\paragraph{Generic symbolic-search methods.}
We run EPLEX~\citep{la2016epsilon} using Ellyn.
The default configuration includes arithmetic, exponential, logarithmic, square-root, square, and cube operators.
The poly configuration uses arithmetic, square, and cube operators, including division.
The two configurations use their respective population sizes, generation limits, and time limits in Table~\ref{tab:supp_baseline_search}.

FE-AFP~\citep{schmidt2010age} uses Ellyn with fitness estimation enabled.
Each estimator approximates candidate-program fitness using a small subset of the fitting data.
We score estimators by how well they preserve the full-data fitness ranking of a trainer set of candidate programs.
Table~\ref{tab:supp_baseline_search} reports the estimator and trainer population sizes, observations per estimator, and trainer-selection interval.

We retain the expression-construction settings of FFX~\citep{mcconaghy2011ffx} and its built-in parameter selection.
The fitting routine ranks returned ODE candidates by trajectory \(R^2\).

PySR~\citep{cranmer2023interpretable} uses the installed Python package with its Julia search backend.
We fit derivative targets with squared error, the configured arithmetic operators, and 50 search iterations.
The fitting routine combines component expressions into ODE candidates and ranks them by simulated trajectory \(R^2\).

\newpage
\paragraph{Sparse system-identification methods.}
We run SINDy~\citep{brunton2016discovering} with a polynomial library and sequentially thresholded least squares (STLSQ).
The search covers polynomial degrees one through five, together with the threshold, regularization, iteration, and differentiation settings in Table~\ref{tab:supp_baseline_other}.

We use a Python implementation of SINDy-PI~\citep{kaheman2020sindy}.
It searches implicit relations with left-hand sides of the form \(\dot{x}_j g(\mathbf{x})\), then converts each selected relation to an explicit differential equation.
The state library has degree two, and the denominator library has degree one.
We search the listed sparsity thresholds.

Reactive SINDy~\citep{hoffmann2019reactive} uses a local Python implementation of nonnegative sparse regression over vector-valued reaction basis functions.
We enumerate mass-action reactions up to order two, including fission and double-conversion candidates, and remove duplicates.
The fitter uses L-BFGS-B with nonnegative rate constraints and the regularization grid in Table~\ref{tab:supp_baseline_other}.
Absolute and relative rate thresholds remove small fitted terms.
Both SINDy-PI and Reactive SINDy normalize library columns and fit derivatives at interior time points.

\paragraph{Neural and language-model methods.}
We load the authors' pretrained ODEFormer~\citep{d2023odeformer} checkpoint without fine-tuning.
We use its symbolic-transformer regressor with candidate beam size 50, sampling temperature \(0.1\), and input rescaling.
The regressor ranks candidate systems by trajectory \(R^2\).
We return the highest-ranked system.

LLM-SR~\citep{shojaee2024llm} runs a separate search for each state equation, using Qwen3-Coder-Flash~\citep{yang2025qwen3} as the language model.
Each search receives time and the full state vector and fits program parameters against the corresponding derivative target.
We allocate 100 generated samples per state variable, giving a nominal budget of \(100d\) samples for a \(d\)-dimensional system.
Each program has at most eight fitted parameters and a 30-second evaluation limit.
We assemble the selected programs into the final ODE.

\paragraph{SISR.}
We evaluate SISR~\citep{palma2026stoichiometrically} over mechanisms containing 2--12 reactions, with a separate subpopulation for each reaction count.
We fit their rate constants by nonnegative least squares.
The clean configuration uses 2,000 individuals and 20 generations per subpopulation.
The original study selects a mechanism by inspecting the complexity--error Pareto front~\citep{palma2026stoichiometrically}.
We implement this loss--complexity principle using deterministic \(\alpha\)-threshold selection with \(\alpha=1.05\) and \(\epsilon=10^{-12}\) (Appendix~\ref{app:alpha_threshold}).
The fixed-archive comparison in Section~\ref{app:selection_noise_settings} evaluates the effect of the final selector.

\clearpage
\input{Table/supp_method_configurations}
\clearpage

\subsection{KineSelect Configuration}
\label{app:kineselect_configuration}

\paragraph{Genetic search.}
We use the reaction representation and variation operators in Appendix~\ref{app:reaction_search}.
Each reaction-count subpopulation has a configured population of 2,000 and runs for at most 20 generations.
Tournament selection, elitism, and mutation/crossover use the values in Table~\ref{tab:supp_kineselect_configuration}.
Early stopping monitors the derivative-fitting loss and ends a subpopulation search after four generations without a relative improvement of at least \(10^{-4}\).

\paragraph{Kinetic candidates and fitting.}
For each species, we construct \(K\) candidates from the 0.05 and 0.50 quantiles of its positive fitting observations.
Thus, systems share the candidate-generation rule while their numerical \(K\) values depend on the observed species trajectories.
Hill candidates use exponents \(1,2,4\).
We set the regularization coefficient to \(\lambda=10^{-6}\) in Eq.~\eqref{eq:kfi_joint_fit}.

\paragraph{Archive and final selection.}
We simulate the 32 candidates with the lowest derivative loss from each subpopulation to form the archive described in Appendix~\ref{app:reaction_search}.
FPMS selects the final model using fidelity criteria \(\mathcal{L}_i^{\mathrm{der}}\) and \(\mathcal{L}_i^{\mathrm{sim}}\), and parsimony criteria \(C_i^{\mathrm{tree}}\) and \(C_i^{\mathrm{spec}}\), as defined in Section~\ref{subsec:fpms}.
Appendix~\ref{app:fpms_contract} gives normalization and tie-breaking, and Appendix~\ref{app:reaction_search} specifies the derivative-based fallback.

\input{Table/supp_kineselect_configuration}

\clearpage
\subsection{Settings for Selection Studies and Noise Experiments}
\label{app:selection_noise_settings}

\paragraph{KFI and FPMS ablations.}
We evaluate four configurations of KineSelect by enabling or disabling KFI and FPMS.
Without KFI, all reaction laws are fixed to mass action, and nonnegative least squares fits their rate constants.
Without FPMS, final models use \(\alpha\)-threshold selection with \(\alpha=1.05\).
Within each KFI setting, the two selectors reuse the same saved candidate archive and fitted parameters.
For the mass-action-only configuration, we reintegrate the selected models with their saved rate constants using LSODA~\citep{petzold1983automatic}, relative tolerance \(10^{-6}\), and absolute tolerance \(10^{-9}\).

\paragraph{FPMS component ablations.}
We apply the four settings in Table~\ref{tab:supp_fpms_ablation_settings} to KineSelect.
The base uses simulation loss \(\mathcal{L}_i^{\mathrm{sim}}\) and expanded ODE expression-tree complexity \(C_i^{\mathrm{tree}}\).
We add derivative loss \(\mathcal{L}_i^{\mathrm{der}}\), specialized kinetic-component count \(C_i^{\mathrm{spec}}\), or both.
Each setting reuses the same candidates and fitted parameters, with normalization and tie-breaking from Appendix~\ref{app:fpms_contract}.

\input{Table/supp_fpms_ablation_settings}

\paragraph{Observation-noise experiments.}
We rerun FFX, SINDy, PySR, SINDy-PI, ODEFormer, Reactive SINDy, SISR, and KineSelect on fitting trajectories corrupted at the noise levels in Appendix~\ref{app:id_ood_noise}.
\rev{In the original runs, all methods receive the same corrupted observations for each system and noise level, using noise seed 2026.}
The recovered systems are evaluated against the clean ID reconstruction and OOD extrapolation trajectories.
For SISR, the noisy runs use elitism \(0.4\) in the two-reaction subpopulation and \(0.1\) in the remaining subpopulations; the clean configuration uses \(0.1\) throughout.

\subsection{Reproduction Resources}
\label{app:compute_resources}

The GitHub repository \sysbiogithub{\texttt{hengrao02/SysBio-SR-KineSelect}} contains the benchmark,
the KineSelect implementation, and the code for running and evaluating the experiments.
The repository also documents the software dependencies and experimental environment.
The benchmark is available on Hugging Face as \sysbiohuggingface{\texttt{HengRao/SysBio-SR}}.

%% file: Table/supp_method_configurations.tex
\begin{table}[!htp]
\centering
\caption{\textbf{Generic symbolic-search configurations.}
Entries give fixed settings or indicate built-in parameter selection.
All time limits are in seconds.
The two EPLEX configurations differ in both operators and budget.}
\label{tab:supp_baseline_search}
\begingroup
\setlength{\tabcolsep}{4pt}
\renewcommand{\arraystretch}{1.06}
\begin{tabularx}{\linewidth}{@{}>{\raggedright\arraybackslash}p{0.22\linewidth}>{\raggedright\arraybackslash}p{0.40\linewidth}>{\raggedright\arraybackslash}X@{}}
\toprule
Method & Hyperparameter & Value(s) \\
\midrule
EPLEX (default) & Population / generations & 200 / 100 \\
 & Time limit per state equation (s) & 180 \\
 & Operators & \(+,-,\times,\div,\allowbreak
   \exp,\log,\allowbreak
   \sqrt{\cdot},(\cdot)^2,(\cdot)^3\) \\
\addlinespace
EPLEX (poly) & Population / generations & 500 / 500 \\
 & Time limit per state equation (s) & 7,200 \\
 & Operators & \(+,-,\times,\div,(\cdot)^2,(\cdot)^3\) \\
\addlinespace
EPLEX (both) & Maximum / initial expression length & 64 / 20 \\
 & Difference order / smoothing & 2 / none \\
 & Built-in parameter selection & Disabled \\
 & Candidate ranking & Trajectory \(R^2\) \\
\midrule
FE-AFP & Population / generations & 1,000 / 250 \\
 & Search time limit per system & 3,600 \\
 & Operators & \(+,-,\times,\div,(\cdot)^2,(\cdot)^3\) \\
 & Maximum / initial expression length & 64 / 20 \\
 & Fitness estimation / rank-based scoring & Enabled / enabled \\
 & Fitness-estimator population & 100 estimators \\
 & Observations per estimator & 10 \\
 & Trainer population & 10 programs \\
 & Trainer-selection interval & 10 generations \\
 & Difference order / smoothing & 2 / none \\
 & Built-in parameter selection & Disabled \\
 & Candidate ranking & Trajectory \(R^2\) \\
\midrule
FFX & Difference order / smoothing defaults & 2 / none \\
 & Built-in parameter selection & Enabled \\
 & Candidate ranking & Trajectory \(R^2\) \\
\midrule
PySR & Search iterations & 50 \\
 & Binary operators & \(+,-,\times,\div,\mathrm{pow}\) \\
 & Unary operator & Negation \\
 & Loss / search processes & Squared error / 1 \\
 & Difference order / smoothing defaults & 2 / none \\
 & Built-in parameter selection & Enabled \\
 & Candidate ranking & Trajectory \(R^2\) \\
\bottomrule
\end{tabularx}
\endgroup
\end{table}

\clearpage
\begin{table}[!htp]
\centering
\caption{\textbf{Sparse, neural, language-model, and SISR configurations.}
\emph{Searched} denotes values explored by the built-in fitting procedure.
SISR population and generation limits apply to each reaction-count
subpopulation. The SISR elitism entry describes the clean configuration.}
\label{tab:supp_baseline_other}
\begingroup
\setlength{\tabcolsep}{4pt}
\renewcommand{\arraystretch}{1.04}
\begin{tabularx}{\linewidth}{@{}>{\raggedright\arraybackslash}p{0.22\linewidth}>{\raggedright\arraybackslash}p{0.42\linewidth}>{\raggedright\arraybackslash}X@{}}
\toprule
Method & Hyperparameter & Value(s) \\
\midrule
SINDy & Polynomial degree (searched) & \(1,2,3,4,5\) \\
 & STLSQ threshold (searched) & \(0.05,0.1,0.15\) \\
 & STLSQ regularization (searched) & \(0.025,0.05,0.075\) \\
 & STLSQ iterations (searched) & \(20,100\) \\
 & Difference order (searched) & \(2,3,4\) \\
 & Smoothing window (searched) & None, 15 \\
 & State normalization & None \\
\addlinespace
SINDy-PI & State / denominator polynomial degree & 2 / 1 \\
 & Sparsity threshold (searched) & \(0.01,0.025,0.05,0.1,0.2\) \\
 & STLSQ iterations & 20 \\
 & Difference order / smoothing & 2 / none \\
 & Minimum denominator magnitude & \(10^{-8}\) \\
\addlinespace
Reactive SINDy & Reaction library / maximum order & Enumerated / 2 \\
 & Fission / double conversion & Enabled / enabled \\
 & Penalty scale (searched) & \(0.001,0.01,0.1,0.5,1\) \\
 & \(\ell_1\) fraction & 1 \\
 & Optimizer / maximum iterations & L-BFGS-B / 5,000 \\
 & Optimization tolerance & \(10^{-10}\) \\
 & Absolute / relative rate threshold & \(10^{-8}\) / \(10^{-4}\) \\
\addlinespace
SINDy-PI and Reactive SINDy & Sparsity-selection tolerance & 0.05 \\
\midrule
ODEFormer & Candidate beam size / temperature & 50 / 0.1 \\
 & Maximum input points & 10,000 \\
 & Rescaling / candidate ranking & Enabled / trajectory \(R^2\) \\
\midrule
LLM-SR & Language model & Qwen3-Coder-Flash \\
 & Temperature / samples per prompt & 0.7 / 4 \\
 & Generated samples per state variable & 100 \\
 & Samplers / evaluators / program islands & 1 / 1 / 10 \\
 & Functions per prompt / reset period (s) & 2 / 14,400 \\
 & Maximum fitted parameters / time input & 8 / enabled \\
 & Program-evaluation limit (s) & 30 \\
 & Prediction integrator / timeout (s) & LSODA / 15 \\
 & Prediction rtol / atol & \(10^{-6}\) / \(10^{-8}\) \\
\midrule
SISR & Reaction-count range & 2--12 \\
 & Population / generations per subpopulation & 2,000 / 20 \\
 & Elitism / mutation probability & 0.1 / 0.1 \\
 & Base search seed & 42 \\
 & Reaction order / stoichiometric bound & 2 / 2 \\
 & Final selection & \(\alpha\)-threshold \\
 & \(\alpha\) / numerical \(\epsilon\) & 1.05 / \(10^{-12}\) \\
 & Candidate integration rtol / atol & \(10^{-6}\) / \(10^{-9}\) \\
\bottomrule
\end{tabularx}
\endgroup
\end{table}

%% file: Table/supp_kineselect_configuration.tex
\begin{table}[!htp]
\centering
\caption{\textbf{KineSelect runtime configuration.}
Population size and simulation-candidate limits apply to each reaction-count
subpopulation. The \(K\)-candidate quantiles use each species' fitting
observations.}
\label{tab:supp_kineselect_configuration}
\begingroup
\setlength{\tabcolsep}{5pt}
\renewcommand{\arraystretch}{1.06}
\begin{tabularx}{\linewidth}{@{}>{\raggedright\arraybackslash}p{0.56\linewidth}>{\raggedright\arraybackslash}X@{}}
\toprule
Hyperparameter & Value \\
\midrule
\multicolumn{2}{@{}l}{\textit{Genetic search}} \\
Reaction-count range & 2--12 \\
Population / maximum generations & 2,000 / 20 \\
Elitism / tournament size & 0.1 / 20 \\
Mutation / crossover rate & 0.3 / 0.7 \\
Base search seed & 42 \\
Early-stopping patience / relative improvement & 4 generations / \(10^{-4}\) \\
Maximum reactant / product order & 2 / 2 \\
Maximum stoichiometry per species & 2 \\
\midrule
\multicolumn{2}{@{}l}{\textit{Kinetic selection}} \\
Reactant-dependent kinetics & Mass action, Michaelis--Menten \\
Regulatory terms & None, linear, Hill activation/inhibition \\
Species-specific \(K\) quantiles & 0.05, 0.50 \\
Hill exponents & \(1,2,4\) \\
\midrule
\multicolumn{2}{@{}l}{\textit{Archive and final selection}} \\
Simulation candidates per subpopulation & 32 \\
Final selector & FPMS \\
\bottomrule
\end{tabularx}
\endgroup
\end{table}

%% file: Table/supp_fpms_ablation_settings.tex
\begin{table}[!htp]
\centering
\caption{\textbf{FPMS component-ablation settings.}
All settings use \mbox{\(U_i=(1-Q_i)(1-P_i)\)}.
Tildes denote the transformations and normalization in
Appendix~\ref{app:fpms_contract}.}
\label{tab:supp_fpms_ablation_settings}
\begingroup
\setlength{\tabcolsep}{5pt}
\renewcommand{\arraystretch}{1.55}
\begin{tabularx}{\linewidth}{@{}>{\raggedright\arraybackslash}p{0.40\linewidth}>{\centering\arraybackslash}X>{\centering\arraybackslash}X@{}}
\toprule
Setting & Fidelity score \(Q_i\) & Parsimony score \(P_i\) \\
\midrule
Base
 & \(\widetilde{\mathcal{L}}_i^{\mathrm{sim}}\)
 & \(\widetilde{C}_i^{\mathrm{tree}}\) \\
Base + derivative loss
 & \(\dfrac{\widetilde{\mathcal{L}}_i^{\mathrm{der}}+
             \widetilde{\mathcal{L}}_i^{\mathrm{sim}}}{2}\)
 & \(\widetilde{C}_i^{\mathrm{tree}}\) \\
Base + specialized kinetic-component count
 & \(\widetilde{\mathcal{L}}_i^{\mathrm{sim}}\)
 & \(\dfrac{\widetilde{C}_i^{\mathrm{tree}}+
             \widetilde{C}_i^{\mathrm{spec}}}{2}\) \\
Full FPMS
 & \(\dfrac{\widetilde{\mathcal{L}}_i^{\mathrm{der}}+
             \widetilde{\mathcal{L}}_i^{\mathrm{sim}}}{2}\)
 & \(\dfrac{\widetilde{C}_i^{\mathrm{tree}}+
             \widetilde{C}_i^{\mathrm{spec}}}{2}\) \\
\bottomrule
\end{tabularx}
\endgroup
\end{table}